\documentclass[a4paper,number]{elsarticle}
\pdfoutput=1
\usepackage{graphicx}
\usepackage[utf8]{inputenc}
\usepackage{amssymb,amsmath,array,amsthm}
\usepackage{stmaryrd}
\usepackage{url}
\usepackage{algorithm,algorithmic}
\biboptions{sort}

\newcommand{\R}[1]{\ensuremath{\mathbb{R}^{#1}}}
\newcommand{\esp}[1]{\mathbb{E}\left(#1\right)}
\newcommand{\espb}[2]{\mathbb{E}_{#1}\left(#2\right)}

\journal{Neurocomputing}

\begin{document}

\sloppy
\begin{frontmatter}

\title{Optimizing an Organized Modularity Measure for Topographic Graph
  Clustering: a Deterministic Annealing Approach}

\author[fab]{Fabrice Rossi}
\ead{Fabrice.Rossi@telecom-paristech.fr}
\author[nat1,nat2]{Nathalie Villa-Vialaneix}\ead{nathalie.villa@math.univ-toulouse.fr}
\address[fab]{BILab, Télécom ParisTech, LTCI - UMR CNRS 5141, France}
\address[nat1]{Institut de Mathématiques de Toulouse, Université de Toulouse, France}
\address[nat2]{IUT STID (Carcassonne), Université de Perpignan Via Domitia, France}

\begin{abstract}
  This paper proposes an organized generalization of Newman and Girvan's
  modularity measure for graph clustering. Optimized via a deterministic
  annealing scheme, this measure produces topologically ordered graph
  clusterings that lead to faithful and readable graph representations based on
  clustering induced graphs. Topographic graph clustering provides an alternative
  to more classical solutions in which a standard graph clustering method is
  applied to build a simpler graph that is then represented with a graph
  layout algorithm. A comparative study on four real world graphs ranging from
  34 to 1 133 vertices shows the interest of the proposed approach with
  respect to   classical solutions and to self-organizing maps for graphs.
\end{abstract}

\end{frontmatter}

\section{Introduction}
Large and complex graphs are natural ways of describing real world systems
that involve interactions between objects:
persons and/or organizations in social networks, articles in citation
networks, web sites on the world wide web, proteins in regulatory networks,
etc. \cite{Newman2003GraphSurveySIAM}. However, the complexity of real world
graphs limits the possibilities of exploratory analysis: while many graph
drawing and graph visualization methods have been proposed
\cite{DiBattistaEtAl1999GraphDrawing,HermanEtAl2000Graph}, displaying a graph
with a few hundred vertices in a meaningful way remains difficult.

A way to tackle this scalability issue is to simplify a graph prior its
drawing. This can be done by finding clusters of vertices via a graph clustering
method \cite{Schaeffer:COSREV2007}: rather than representing the original
graph, the visualization is restricted to the clusters themselves. More
precisely, the graph induced by the clustering is used: each cluster forms a
vertex, while edges between clusters are induced by edges between the vertices they
contain. As illustrated in the survey paper \cite{HermanEtAl2000Graph},
numerous implementations of this simple idea can be done, namely all the
pairwise combinations between graph clustering methods and graph visualization
methods (see also, e.g., \cite{DBLP:journals/algorithmica/EadesFLN06} for recent
work on planar clustered graphs and \cite{BourquiEtAl2007} for weighted
graphs). 

Rather than relying on a generic graph clustering algorithm, we propose in
this paper to build a clustering of a graph that is adapted to a visualization
of the graph induced by the clustering. We follow the general principal of
topographic mapping initiated by Prof. Kohonen with the Self Organizing Map
(SOM, \cite{KohonenSOM1995}): a SOM builds a clustering of a dataset in
homogeneous and separated clusters that are in addition arranged on a
geometrical structure chosen \emph{a priori}; points assigned to close
clusters in the prior structure are close in the original space.

Our method is built on an \emph{organized} generalization of the popular
\emph{modularity} measure \cite{NewmanGirvanModularity2004} for graph
clustering. Following the SOM rationale, the organized modularity uses a prior
structure to build topologically aware clusters: nodes assigned to close
clusters in the prior structure are more likely to be connected in the
original graph than nodes assigned to distant clusters. As for the SOM
\cite{Vesanto1999SomVisu,VesantoPhD2002}, using a two dimensional regular grid
as a prior structure enables straightforward visualization of the graph
induced by the clustering. 

While the modularity has interesting properties over other graph cut measures
\cite{Newman2006Eigenvectors,NewmanCommunity2004EurPhys,NewmanPNAS2006,LehmannHansen2007DAMOD},
especially in the visualization context \cite{Noack2009},
it has two drawbacks shared with its organized generalization. Firstly,
maximizing the (organized) modularity is a combinatorial problem and there is
no simple prototype based alternating optimization scheme as available for the
SOM and its non Euclidean variants
\cite{HammerEtAlRelationalWSOM2007,BouletEtAl2008Neurocomputing,VillaRossiKSOMWSOM2007,VillaEtAlMashs2008}. Following
\cite{LehmannHansen2007DAMOD} for the modularity, we rely on a deterministic
annealing approach \cite{RoseDeterministicAnnealing1999} to solve this
problem. Secondly, maximizing the (organized) modularity measure tends to miss
small clusters, i.e., to aggregate them in large clusters
\cite{FortunatoBarthelemy2007}. To limit this effect, we also propose to use the
intermediate
states of the deterministic annealing algorithm to build finer clusterings than the
one with maximal organized modularity. Combined with the prior structure,
those clusterings give refined views of the graph induced by the
clustering. In summary our contribution is twofold: we introduce an organized
modularity measure and we use deterministic annealing to build fair and
simplified visualizations of a graph, that are based on a clustering of the
vertices.

The rest of this paper is organized as follows: Section
\ref{topographicClustering} details the clustering approach to graph
visualization, recalls the modularity definition and introduces its
topographic generalization (the \emph{organized} modularity). Section
\ref{deterministic} presents the deterministic annealing scheme used to
optimize the organized modularity. Section \ref{section:graph:visu} describes
the proposed graph visualization methodology and shows in particular how to
leverage the intermediate results of the optimization algorithm to produce
fuzzy layouts that limit the resolution effect induced by modularity
maximization. Section \ref{karate} analyses in detail the behavior of the
proposed method on a simple graph, Zachary's Karate club social network
\cite{Zachary77Karate}. Finally, Section \ref{section:large:graphs}
demonstrates the practical interests of the method on three larger real world
graphs. Technical derivations are gathered in an appendix.

\section{Topographic graph clustering}\label{topographicClustering}

\subsection{Graph clustering for visualization}\label{graphclustvisu}
Cognitive scalability can be brought to graph visualization methods
\cite{HermanEtAl2000Graph} via graph clustering \cite{Schaeffer:COSREV2007}:
the rationale is to draw the smaller and hopefully simpler graph induced by
the clustering instead of the original graph. Let us define more precisely this
idea.

We consider given a non oriented graph $G$ with $N$ vertices (or nodes),
$V(G)=\{1,\ldots,N\}$ and $A$ weighted edges ($A(G)$ is the set of edges). The
(symmetric) weight matrix is denoted by $W$, and $W_{ij}$ is the non negative
weight between vertex $i$ and vertex $j$ ($W_{ij}=0$ when $(i,j)\not\in
A(G)$). A clustering of $V(G)$ into $C$ clusters $(C_k)_{1\leq k\leq C}$
induces a new non oriented graph $G_C$ defined as follows: each cluster is
associated to a vertex in $G_C$, i.e., $V(G_C)=\{1,\ldots,C\}$ and a pair
$(i,j)$ belongs to the edge set $A(G_C)$ if and only if there is $(k,l)\in
A(G)$ such that $k\in C_i$ and $l\in C_j$. $G_C$ is weighted and the
corresponding weight matrix is denoted $W^C$. The weights are defined by
$W^C_{ij}=\sum_{k\in C_i,l\in C_j}W_{kl}$ (for $i\neq j$).

The present paper focuses on the following graph visualization strategy: a
clustering $C$ of the original graph $G$ is constructed and a graph
visualization method is applied to $G_C$. In general, the visual
representation is completely unaware of the way $G_C$ has been produced: this
is a simpler setting than the one of
e.g. \cite{DBLP:journals/algorithmica/EadesFLN06,BourquiEtAl2007} in which the
proposed algorithms aim at representing a so-called \emph{clustered graph}. In
this latter context, the ultimate goal is to visualize the \emph{complete}
original graph in a way that respects the (hierarchical) clustering
structure. The clustering is used in this case both as a way to circumvent
algorithmic difficulties (e.g., the cost of some force directed methods) and
as a structural organization principle. In the present paper, the clustering
method makes ``a meaningful coarse graining of the graph'' (to quote
\cite{NewmanGirvanModularity2004}) and produces a new graph which will be the
main target of the visualization method.

The solution chosen in this paper is also related to but quite different from
the \emph{clustered layout} strategy \cite{HermanEtAl2000Graph}. In this
framework, one aims at producing  a visualization of a graph in which vertices
that belong to a given cluster are close, while distinct clusters remain
separated. Clusters are generally a by product of those algorithms rather than
given beforehand (see, e.g., \cite{Noack2007JGAA} for a recent example of such
a method).  

In practice, we target visualization methods in which each vertex of the new
graph $G_C$ is represented by a glyph (e.g., a disk) while each edge is drawn
as a straight segment between the corresponding vertices' glyphs (we do not
consider more sophisticated drawing, for instance with bends). Rendering hints
can be used to convey information about the original graph to the user: the
surface occupied by a glyph can be proportional to the size of the
corresponding cluster, while the thickness of a segment can encode the weight
associated to the corresponding edge. 

\subsection{Quality measures}\label{subsection:quality}
In order to be useful, a graph visualization produced via the chosen
clustering framework must be both \emph{readable} and \emph{faithful}. The
readability issue is common to all graph visualization algorithms and has been
a matter of research and debate in the graph visualization community with a
particular focus on aesthetic criteria, such as bend minimization, symmetry,
etc. \cite{DiBattistaEtAl1999GraphDrawing,HermanEtAl2000Graph,Purchase2002}.
Experimental studies show that a minimal number of edge crossings is one of
the most important quality criteria for readability (see
e.g. \cite{WareEtal2002}). Please note however that there is no consensus on
what other readability criteria should be used to evaluate the quality of a
graph representation beyond direct task based user studies. In particular,
neighborhood rank preservation measures used to assess the quality of
nonlinear projection methods \cite{LeeVerleysen2009} are likely to be
irrelevant in the graph context: as showed in \cite{FabrikantEtAl2004CGIS},
displaying the edges of a graph has a major influence on the way proximities
between nodes are inferred by users from the visualization.

The faithfulness issue is more specific to the cluster based visualization: as
the final representation of the graph hides most of its structure, inference
on the drawing can be misleading. More precisely, there are two potential
problems induced respectively by the vertices and the edges in the $G_C$
graph. The first difficulty is that the internal connectivity structure of
clusters is hidden by the glyph based representation: while color and/or shape
hints could be used to give an idea of the density of connections between the
vertices that have been put in a given cluster, the actual connection pattern
is lost. A similar problem happens for connection patterns between vertices
of distinct clusters: as the associated edges are summarized by a single edge
between the induced vertices, there is no way to infer how individual vertices
are connected in the original graph.

Therefore, building a faithful clustering of a graph amounts to balancing two
criteria that are somewhat contradictory: on the one hand, the density of
each cluster should be maximized, but on the other hand the connectivity
pattern between two vertices of distinct clusters should be independent of the
vertices. Intuitively, a dense cluster with a low outside connectivity is
likely to have a rather complex outside connectivity pattern: if all vertices
in a cluster were connected to e.g., a single outside vertex, then it might
make sense to assign this vertex to the cluster. In addition, enforcing a
simple between cluster connectivity has no reason to reduce edge crossing,
while within cluster density maximization should on the contrary reduce the
number of between cluster connection and therefore improve the readability of
the graph.

It seems therefore natural to focus on within cluster density maximization as
this solution should lead to a readable graph with a controlled impact on the
faithfulness of the representation. Among the clustering quality criteria that
favor within cluster density (see \cite{Schaeffer:COSREV2007}), we focus on the
\emph{modularity} introduced by Newman and Girvan in
\cite{NewmanGirvanModularity2004}. Let us denote $k_i=\sum_{j}W_{ij}$ the
degree of vertex $i$ and $m=\frac{1}{2}\sum_{i,j}W_{ij}$ the total weight of
the graph. Then the modularity of the clustering
$(C_k)_{1\leq k\leq C}$ of $G$ is
\begin{equation}
  \label{eq:Modularity}
Q\left((C_k)_{1\leq k\leq C}\right)=\frac{1}{2m}\sum_{k=1}^C\sum_{i,j \in C_k}\left(W_{ij}-P_{ij}\right),
\end{equation}
where $P$ is a $N\times N$ symmetric matrix given by
$P_{ij}=\frac{k_ik_j}{2m}$. 

The rationale of the measure is to compare the weight of a link between two
vertices in a cluster, $W_{ij}$, to a simple random model, $P_{ij}$, in which
the weights are proportional to the degrees of the vertices and independent of
the clusters. A good clustering tends to cluster vertices that are more
connected that one expects based solely on the degrees of the vertices: this
corresponds to $W_{ij}>P_{ij}$ and to higher values of $Q(M)$. Maximizing the
modularity tends therefore to produce dense clusters, but only when they are
meaningful as measured by a deviation from the simple random model.

The modularity is a rather successful quality measure for graph clustering,
especially compared to other graph cut like measures
\cite{Newman2006Eigenvectors,NewmanCommunity2004EurPhys,NewmanPNAS2006,LehmannHansen2007DAMOD}. In
addition, \cite{Noack2009} has recently shown that there is a strong link
between clusterings induced by modularity maximization and those obtained with
some versions of the clustered layout strategy described previously. The
modularity seems therefore to be a good candidate for the topographic
extension studied in this paper. However, this measure suffers from a
resolution problem \cite{FortunatoBarthelemy2007}: maximizing the modularity
may fail to identify small but meaningful clusters of nodes. We will tackle
this problem by leveraging both the chosen optimization algorithm and the prior
structure introduced below (see Section \ref{subsection:fuzzylayout} for
details). It should be noted in addition that the main concepts and techniques
used in this paper are quite independent from the actual quality measure. In
particular most graph cut based measures \cite{Schaeffer:COSREV2007} could be
handled in a similar way as we proceed with modularity.

\subsection{Organized modularity}\label{organized modularity}
The main limitation of graph clustering for visualization is that the two
phases of the methodology are generally completely independent: the clustering
method is not explicitly designed to help the subsequent visualization
method. We propose in this paper to address this limitation via a specific
quality measure for the clustering phase. 

Following the SOM rationale, we derive an \emph{organized} version of the
modularity. We assume given a prior structure (in \R{2}) which is represented
by a symmetric $C$ by $C$ matrix $S$ of prior similarities between
clusters such that $S_{kk}=1$ for all $k=1,\ldots,C$. For instance $S_{kl}=\exp(-\sigma\|\mathbf{x}_k-\mathbf{x}_l\|^2)$
where $\mathbf{x}_k$ is the prior position of cluster $C_k$ in \R{2}. Then the
\emph{organized modularity} of the clustering $(C_k)_{1\leq k\leq C}$ of $G$ is
\begin{equation}
  \label{eq:OrganizedModularity}
O\left((C_k)_{1\leq k\leq C}\right)=\frac{1}{2m}\sum_{i,j}S_{c(i)c(j)}\left(W_{ij}-P_{ij}\right),
\end{equation}
where $c(i)$ is the index of the cluster to which the vertex $i$ is assigned. 

In the standard modularity, the term $W_{ij}-P_{ij}$ is taken into account
only when $i$ and $j$ belong to the same cluster. In the organized version,
this term is \emph{always} taken in account, but with a weight $S_{c(i)c(j)}$
equal to the prior similarity between $C_{c(i)}$  and $C_{c(j)}$. This favors
proximity in the prior structure for connected clusters. If 
there are indeed significant connections between vertices in two clusters
$C_k$ and $C_l$ (i.e., $W_{ij}-P_{ij}>0$), then the value of $O$ will be
higher if $S_{kl}$ is high than if it is low. This is similar to the SOM
principle in which a prototype has to be close to observations assigned to its
unit but also, to a lesser extent, to observations assigned to neighboring
units in the prior structure. 

Maximizing $O\left((C_k)_{1\leq k\leq C}\right)$ gives a graph clustering that
can be used to coarsen the original graph prior a standard visualization
algorithm, as explained in Section \ref{graphclustvisu}. In addition, the
prior structure gives a natural layout for the clustered graph: nodes of
this new graph can be drawn at their corresponding positions on the grid.

\section{Organized modularity maximization}\label{deterministic}
\subsection{Modularity maximization}
The main difficulty with (organized) modularity maximization is that it is a
discrete optimization problem for which there is no simple alternate
minimization scheme (contrarily to the SOM or the $k$-means, for
instance). Deriving fast modularity maximization algorithms that produce
acceptable solutions has been an important research topic since the
introduction of this quality measure (this is generally the case for all graph
clusteringing measures). The best compromise between
computational load and quality seems to be currently achieved by some type of
heuristic algorithms that coarsen the graph in an iterative and multi-level
way \cite{BlondelEtAll2008,NoackRotta2008}. 

In the present paper, we use a quite different approach which is more adapted
to the organized modularity defined in the previous Section. Following
\cite{LehmannHansen2007DAMOD}, we propose to maximize the (organized)
modularity via a deterministic annealing (DA) approach
\cite{RoseDeterministicAnnealing1999}. As pointed out in
\cite{LehmannHansen2007DAMOD}, the main advantage of DA
over simulated annealing (as used by, e.g., \cite{ReichardtBornholdt2006} in the
graph clustering context) is the speed of the former: deterministic annealing
can use aggressive annealing schedules with a relatively small number of
iterations compared to simulated annealing. In addition, intermediate results
of DA can be used to limit the resolution effect induced by modularity
maximization, as will be shown in Section~\ref{subsection:fuzzylayout}. 

To ease the derivation of the DA
algorithm for modularity maximization, we use 
an assignment matrix notation for clusterings. An assignment matrix $M$ for a
clustering of $\{1,\ldots,N\}$ in $C$ clusters is a $N\times C$ matrix with
entries in $\{0,1\}$ such that $\sum_{k=1}^CM_{ik}=1$ for all $i$. In other
words, $M_{ik}$ is equal to 1 if and only if $i$ is assigned to cluster
$C_k$. We denote $\mathcal{M}$ the set of all valid assignment matrices for a
clustering of $\{1,\ldots,N\}$ in $C$ clusters ($N$ and $C$ will be given by
the context). Please note that we do not constrain an assignment matrix to
have non empty clusters.

The modularity and the organized modularity of an assignment
matrix $M$ are given respectively by:
\begin{eqnarray}
  \label{eq:Modularity:matrix}
Q(M)&=&\frac{1}{2m}\sum_{i,j}\sum_kM_{ik}M_{jk}\left(W_{ij}-P_{ij}\right) \\
\label{eq:OrganizedModularity:matrix}
O(M)&=&\frac{1}{2m}\sum_{i,j}\sum_{k,l}M_{ik}S_{kl}M_{jl}\left(W_{ij}-P_{ij}\right)
\end{eqnarray}
If $B$ is the $N\times N$ symmetric matrix defined by
\begin{equation}
  \label{eq:BMatrix}
B_{ij}=
\begin{cases}
0 & \text{if }i=j\\
\frac{1}{2m}\left(W_{ij}-P_{ij}\right) & \text{if } i\neq j.
\end{cases}
\end{equation}
then, maximizing $O(M)$ is equivalent to
maximizing  
\begin{equation}
  \label{eq:SimplifiedOMod}
F(M)=\sum_{i,j}\sum_{k,l}M_{ik}S_{kl}M_{jl}B_{ij}
\end{equation}
as shown in Section \ref{section:OandF}. Finally, it should be noted that by using the identity matrix for $S$, one
recovers the modularity: the algorithm derived for the maximization of $F(M)$
will therefore apply to both the standard modularity and the organized
version. 

\subsection{Deterministic annealing and mean field approximation}
Deterministic annealing tries to solve the complex combinatorial problem of
maximizing a function $F$ defined on a finite (but large) space $\mathcal{M}$
via an analysis of the Gibbs distribution obtained as the asymptotic regime of
a classical simulated annealing \cite{KirkpatrickEtalSA1983,Cerny1985} or via
the principal of maximum entropy \cite{RoseDeterministicAnnealing1999}. In our
case, the Gibbs distribution for temperature $\frac{1}{\beta}$ is
\begin{equation}
  \label{eq:GibbsDistribution}
P(M)=\frac{1}{Z_P}\exp(\beta F(M)),
\end{equation}
where the normalization constant $Z_P$ is given by $Z_P=\sum_{M\in
  \mathcal{M}}\exp(\beta F(M))$ (the sum ranges over the full space
$\mathcal{M}$). 

The main idea of DA is to compute the expectations of the assignment matrix at
a fixed temperature with respect to the Gibbs distribution, i.e., the
$\esp{M_{ik}}$, and then to decrease the temperature while tracking the
evolution of the expectations (in simulated annealing, the Gibbs distribution
is sampled rather than studied via expectation).

Unfortunately, $F$ is not linear with respect to $M$ and
computing $Z_P$ and $P$ is therefore difficult: one should resort on an
exhaustive exploration of $\mathcal{M}$ which is computationally
infeasible. Following previous work on 
similar topics, we approximate $P$ by a distribution that factorizes (see
e.g., \cite{HofmannBuhmann1997TPAMI,GraepelEtAl1998GSOM}). This corresponds to
approximating the interaction between say $M_{ik}$ and all the other variables
via a \emph{mean field} $E_{ik}$. Then we compute the expectation of the
assignment matrix with respect to the approximating distribution. 

More precisely, we consider the bi-linear cost function
$U(M,E)=\sum_{i}\sum_{k}M_{ik}E_{ik}$ where $E$ is the mean field, a $N$ by
$C$ matrix of partial assignment costs. For a fixed temperature
$\frac{1}{\beta}$, we look for a mean field $E$ that gives a distribution
$R(M,E)=\frac{1}{Z_R(E)}\exp(\beta U(M,E))$ close to $P(M)$, in the sense that
the Kullback-Leibler divergence $KL(R|P)$ between $R$ and $P$ is minimal, with
$KL(R|P)=\sum_{M}R(M,E)\ln\frac{R(M,E)}{P(M)}$ (this corresponds to a
variational approximation of $P$ \cite{JaakkolaTutorial2000}). 
 
At a minimum, the gradient of $KL(R|P)$ with respect to $E$  is
zero. This leads to the following classical mean field equations (see Appendix \ref{section:MeanFieldEquations} for details):
\begin{equation}\label{eqMeanField}
\frac{\partial \espb{R}{F(M)}}{\partial E_{jl}}=\sum_{k}\frac{\partial
  \espb{R}{M_{jk}}}{\partial E_{jl}}E_{jk},\ \forall j,l.
\end{equation}
They are obtained using the main consequence of the mean field approximation,
namely the independence between $M_{ik}$ and $M_{jl}$ for $i\neq j$ under the
distribution $R$, i.e., the fact that
$\espb{R}{M_{ik}M_{jl}}=\espb{R}{M_{ik}}\espb{R}{M_{jl}}$ for $i\neq j$. 

To solve the mean field equations, we use a EM-like approach. We consider the
$\espb{R}{M_{ik}}$ fixed and solve the equations for $E_{jl}$ (maximization
phase). Then we compute the new values of the $\espb{R}{M_{ik}}$ (expectation
phase). This latter phase leads to the very simple standard deterministic
annealing update rule:
\begin{equation}\label{eqEstep}
\espb{R}{M_{ik}}=\frac{\exp(\beta E_{ik})}{\sum_{l}\exp(\beta E_{il})}.
\end{equation}
Moreover, the independence property recalled above gives
\[
\espb{R}{F(M)}=\sum_{i\neq j}\sum_{k,l}\espb{R}{M_{ik}}S_{kl}\espb{R}{M_{jl}}B_{ij}.
\]
Then, some straightforward calculations (see Appendix
\ref{section:EMequations} for details) show that equation \eqref{eqMeanField}
is fulfilled if the mean field is given by
\begin{equation}\label{eqMstep}
E_{jk}=2\sum_{i\neq j}\sum_{l}\espb{R}{M_{il}}S_{kl}B_{ij},
\end{equation}
or, in matrix notations, $E=B\espb{R}{M}S$, using the symmetry of $B$ and
$S$. 

\subsection{Phase transition and final algorithm}
It is well known that deterministic annealing goes through several phase
transitions when the temperature is decreased
\cite{RoseDeterministicAnnealing1999,GraepelEtAl1997PhaseTransition,HofmannBuhmann1997TPAMI}. In
order to choose an adapted
annealing schedule, i.e., an increasing series of $(\beta_l)_{1\leq l\leq L}$
that is used to track the evolution of $\espb{R}{M}$, it is important to find
at least the first phase transition. 

At the limit of infinite temperature ($\beta=0$), it can be shown (see
Appendix \ref{subsec:fixedpoint}) that the following mean field
\begin{equation}
  \label{eq:MeanFieldZero}
E^0_{jk}=\frac{2}{C}\sum_{i\neq j}B_{ij}\sum_{l}S_{kl},  
\end{equation}
is a fixed point of the $EM$ like scheme (i.e., of equations \eqref{eqEstep} and
\eqref{eqMstep}). The corresponding assignment expectations are uniform, i.e.
\begin{equation}
  \label{eq:ExpectationZero}
\espb{R}{M_{ik}^0}=\frac{1}{C}.
\end{equation}
An analysis of the stability of the EM-like scheme (conducted in Appendix
\ref{subsec:fixedpoint}) shows that the fixed point $E^0$ is stable when the
temperature $\frac{1}{\beta}$ is higher than the critical temperature
$T_0=\frac{2\lambda_B\lambda_S}{C}$, where $\lambda_B$ and $\lambda_S$ are the
spectral radii, i.e., the largest eigenvalues in absolute value, respectively
of $B$ and $S$. The first phase transition 
will therefore happen when the temperature becomes lower than this limit. It
should be noted that the critical temperature is a very conservative
estimation of the temperature of the first phase transition, as it corresponds
to a worst case analysis of the stability of the fixed point. 

\begin{algorithm}
\caption{Deterministic annealing for organized modularity maximization}
\label{algo:da}
  \begin{algorithmic}[1]
    \STATE initialize $E$ to $E^0$ from equation \eqref{eq:MeanFieldZero}
    \STATE $T_0\leftarrow \frac{2\lambda_B\lambda_S}{C}$ 
    \COMMENT{critical temperature} 
    \STATE $T\leftarrow \alpha T_0$ 
    \FOR[annealing loop]{$l=1$ to $L$} 
      \STATE $E\leftarrow E + \varepsilon$ \COMMENT{noise injection} 
      \REPEAT[EM-like phase]
        \STATE compute $\espb{R}{M_{ik}}$ using equation \eqref{eqEstep} with
    $\beta=\frac{1}{T}$ 
        \STATE compute $E$ using equation \eqref{eqMstep}
       \UNTIL{convergence of $E$} 
       \STATE $T\leftarrow \gamma T$
    \ENDFOR
    \STATE threshold $\espb{R}{M}$ into an assignment matrix
  \end{algorithmic}
\end{algorithm}

To derive the final Algorithm \ref{algo:da}, we follow
\cite{RoseDeterministicAnnealing1999} 
rather than \cite{LehmannHansen2007DAMOD}. We use in particular the
perturbation idea of \cite{RoseDeterministicAnnealing1999}: each time the
temperature is lowered, some noise is injected in the mean field before
running the EM-like scheme. This method favors phase transitions and symmetric
breaks that are needed for a proper convergence. 

Deterministic annealing is very robust and quite insensitive to the parameters
that appear in Algorithm \ref{algo:da}:
\begin{itemize}
\item $L$ is the number of outer iterations, i.e., the number of temperatures
  considered during the annealing process (a typical value is $N$, the size of
  the graph);
\item $\alpha>1$ is the relative starting temperature above the critical
  temperature (typically $1.1$);
\item $\gamma$ is the damping factor of the temperature (typically chosen such
  that the final temperature is $0.1T_0$);
\item for the noise $\varepsilon$, we used a random multiplicative factor
  chosen uniformly in $[0.995,1.005]$ for each component of $E$;
\item the EM-like phase is considered to be stable when the mean squared difference
  between the components of two values of $E$ is below the square root of the
  machine precision (or after 500 iterations).
\end{itemize}
Finally, the computation cost of the algorithm remains reasonable: the costly
operation is the product $B\espb{R}{M}$. As explained in
\cite{Newman2006Eigenvectors}, one can leverage the structure of $B$ to obtain
a fast matrix multiplication. Indeed $B$ is made from $W$ the weight matrix
and $P$ the degree matrix. Computing $Wx$ costs $O(A)$ operations (where $A$ is
the number of edges of the graph), while $Px$ can be computed in $O(N)$
operations ($N$ is the number of vertices) exploiting the definition of
$P$. Therefore computing $B\espb{R}{M}$ costs $O(C(A+N))$. Then computing the
product of $B\espb{R}{M}$ by $S$ is a $O(NC^2)$ operation while applying
equation \eqref{eqEstep} costs $O(NC)$. The total cost of one iteration of
Algorithm \ref{algo:da} is therefore $O(CA+C^2N)$. Computing the critical
temperature can be done quite efficiently via the Lanczos method
\cite{GolubVanLoan1996} in $O(N(A+N))$ but in fact only a very rough estimate
of the spectral radius is needed. Therefore, a few iterations of a power method
should be enough to give a reasonable initial temperature. In practice, the
computational cost of the method makes it suitable for graphs with a few
thousands vertices as long as there are not too dense. 

\section{Graph visualization methodology}\label{section:graph:visu}
In principle, the proposed graph visualization methodology is rather
straightforward and strongly related to e.g., exploratory data analysis with
the SOM. We choose a regular grid as a prior structure and use Algorithm
\ref{algo:da} to find an optimal clustering with respect to the organized
modularity defined by equation \eqref{eq:OrganizedModularity:matrix}. Then, as
explained in Section \ref{graphclustvisu}, we display the clustering induced
graph: we do not need a graph layout algorithm in this phase as each cluster
has a dedicated position in the grid.

In practice, some parameters need to be carefully chosen to provide a
meaningful visualization. In addition, (organized) modularity maximization
faces a problem of limited resolution  and leads sometimes to oversimplified
clustering induced graph. We describe in this Section the proposed solutions to
those issues.

\subsection{Parameter tuning}\label{section:parameter:tuning}
The internal parameters of Algorithm \ref{algo:da} described in Section
\ref{algo:da} do not request any particular tuning and the guidelines provided
should in general give satisfactory results. On the contrary, the quality of
the visual representation strongly depends on the prior structure, both in
terms of the size of grid itself and in terms of the $S$ matrix. Those
parameters should be optimize as automatically as possible. The general
principle consists first in defining a finite set of acceptable prior structures
and then in selecting the best one via some quality criterion.

Our practical goal is to end up with readable graphs: we consider therefore
small prior structures, for instance square grids with an absolute maximum of
10 nodes per dimension (i.e., up to 100 clusters). The entries of the matrix
$S$ are calculated via a SOM like equation
  \begin{equation}
    \label{eq:grid:dist}
S_{ij}=H(\lambda\|x_i-x_j\|),
  \end{equation}
where $x_i$ is the position of cluster $i$ in the prior structure and $H$ is
either $H(t)=\exp(-t^2)$ (exponential decrease) or $H(t)=\max(0,1-t)$  (linear
decrease). The scaling parameter $\lambda$ is used to tune the neighboring
influence and can be chosen so as to include from zero neighbor to a complete
influence of all clusters on each other. 

Unfortunately, as recalled in Section \ref{subsection:quality}, there is no
universally accepted quality criterion for graph visualization. Moreover, the
case of clustered visualization naturally corresponds to a trade-off between
internal and external connectivity uniformity. To handle this trade-off we
propose to rely on dual objectives optimization principles. More precisely, we
use the standard modularity measure to assess clustering quality and the
number of edge crossing to assess visual quality. Rather than combining the
measures to select one value of the parameters of the prior structure, we
consider Pareto optimal prior structures: each selected structure is better than all others for at least one of the two criteria. 

In practice, this means that the parameter tuning process is only
semi-automatic: it selects some interesting visualizations from the explored
set of parameters. Those visualizations can then be presented to the user for
selection. As explained in Section \ref{subsection:quality}, while the
modularity is a clustering quality criterion, it favors both dense clusters
and low connectivity between clusters. It is therefore reasonable to sort
Pareto optima in decreasing modularity order prior user analysis. However,
limiting the results to the best prior structure according to modularity will
generally lead to oversimplified graphs because of the limited resolution of
this measure \cite{FortunatoBarthelemy2007}: this justifies presenting to the
user Pareto optima with sub-optimal modularity but with more non empty
clusters and less edge crossings.

\subsection{Fuzzy layout}\label{subsection:fuzzylayout}
In addition, one can leverage the annealing process
to produce intermediate results which can be considered as compromise between
the main strategy and the clustered layout strategy
\cite{Noack2007JGAA}. Indeed, as will be shown in Section \ref{karate},
the expectation of the assignment matrix $\espb{R}{M}$ does not contain 0/1
values, even after some phase transitions. The \emph{fuzzy layout strategy}
introduced in the present section takes advantage of this fact. 

Let us denote $(x_k)_{1\leq  k\leq C}$ the positions associated to the $C$
clusters in the prior structure. The position of vertex $i$ in the prior
structure based layout associated to the assignment matrix $M$ is given by
\[
p_i=\sum_{k}M_{ik}x_k,
\]
and therefore, the expected position is
\begin{equation}
  \label{eq:expected:position}
\espb{R}{p_i}=\sum_{k}\espb{R}{M_{ik}}x_k.
\end{equation}
At the limit of zero temperature, $\espb{R}{M}$ contains only 0 and 1, and the
expected positions are exactly positions in the prior structure. However,
during annealing, vertices that are difficult to classify will have in-between
positions reflecting non peaked values of $\espb{R}{M_{ik}}$. 

As a by product of the annealing scheme, one can therefore provide an animated
rendering of the evolution of $\espb{R}{M}$ by displaying for different values
of the temperature the $N$ points $(\espb{R}{p_i})_{1\leq i\leq N}$ (the
display is somewhat similar in principle to the posterior mean projection used
in the Generative Topographic Mapping \cite{BishopEtAl1998GTM}). To avoid
cluttering the layout with overlapping vertices, we rely on an elementary
simplification scheme: a complete linkage hierarchical clustering is applied
to the $\espb{R}{p_i}$ and the dendrogram is cut at an appropriate level (for
instance 5\% of the minimal distance between the points of the grid). This
leads to a finer clustering than the final one, with associated positions for
the clusters. Many of those clusters are positioned near or on cluster
positions on the grid, especially when the annealing scheme nears
completion. This induces generally some overlapping between edges. To limit
this effect, we use the positions computed above as initial positions of a
force directed placement algorithm (such as the Fruchterman-Reingold algorithm 
\cite{FruchtermanReingoldGraph1991}). The algorithm is used only for a few
iterations that move slightly the clusters and reduce overlapping. 

In practice, the user first browses through the Pareto optimal points sorted
in decreasing modularity order to select some interesting visualizations. Then
he/she can request an animated rendering of the most interesting graphs to
assess whether some sub-structure can be found in the clusters. 

\section{Detailed analysis of a small graph} \label{karate}
This section provides a detailed analysis of the behavior of the proposed
method on a small graph. The method has been implemented in R \cite{RProject},
using the igraph package \cite{igraph} complemented by the network package
\cite{ButtsNetwork2008}. 

\begin{figure}[htbp]
  \centering
\includegraphics[width=\textwidth]{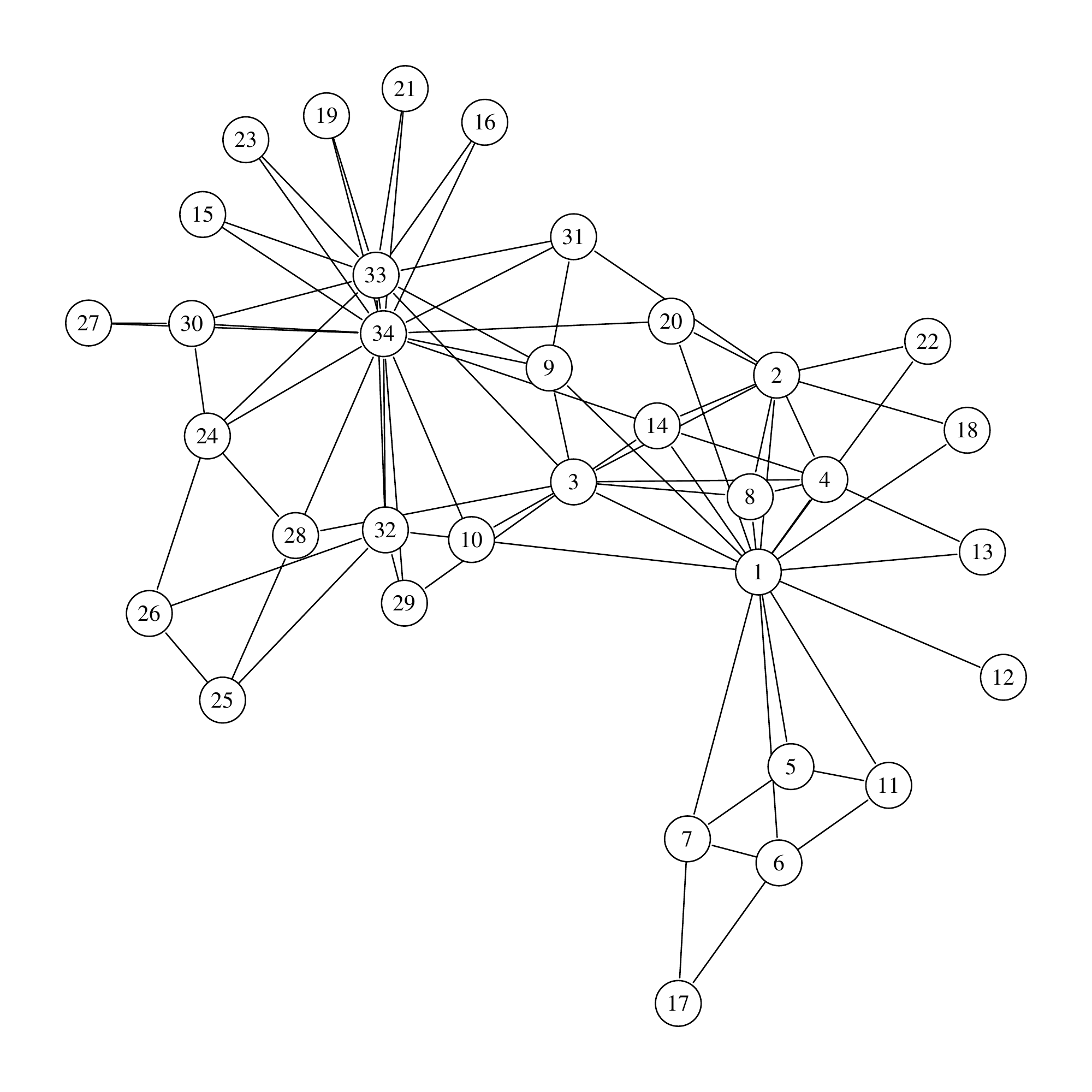}  
  \caption{Zachary's Karate club social network}
  \label{fig:Karate}
\end{figure}

We study Zachary's Karate club social network \cite{Zachary77Karate}. The
graph represents the friendship social network between the 34 members of a
Karate club at a US university in the 70s. The graph contains 78 unweighted
edges (its global density, the fraction of connected pairs of nodes, is thus
equal to 13.9 \%) and is represented on Figure \ref{fig:Karate} obtained with
the Fruchterman-Reingold force directed algorithm
\cite{FruchtermanReingoldGraph1991} as implemented in igraph
\cite{igraph}. The transitivity of the graph, that is a measure of the
probability that the adjacent vertices of a vertex are connected (see e.g.,
\cite{wasserman_faust_SNAMA1994}), is equal to 25.6 \%. The gap between the
global density and the transitivity is a good indication for a larger local
density and thus a relevant clustering.

\subsection{Modularity maximization}
\label{karate-modul}
\begin{figure}[htbp]
  \centering
\includegraphics[width=0.47\textwidth]{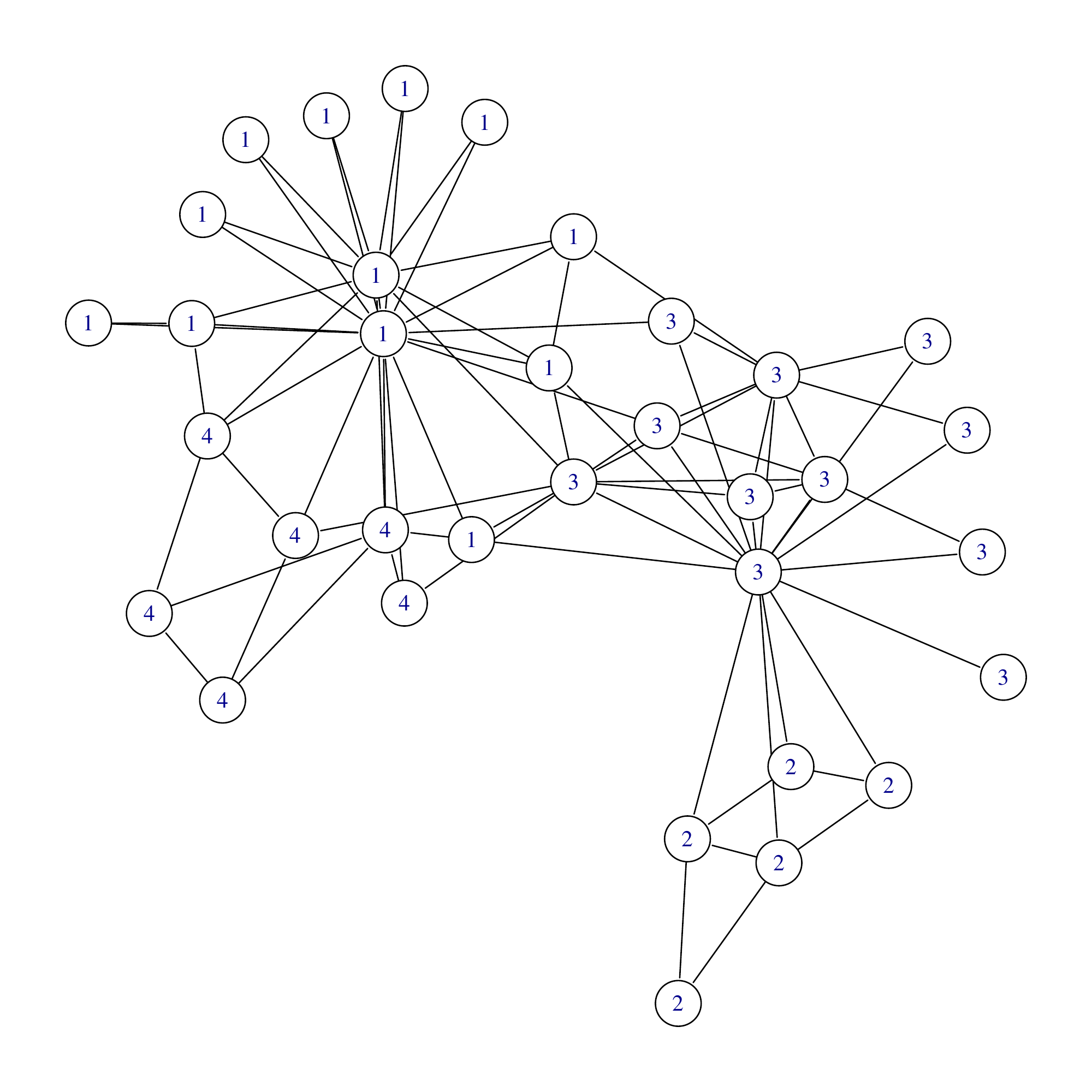}\hfill\includegraphics[width=0.47\textwidth]{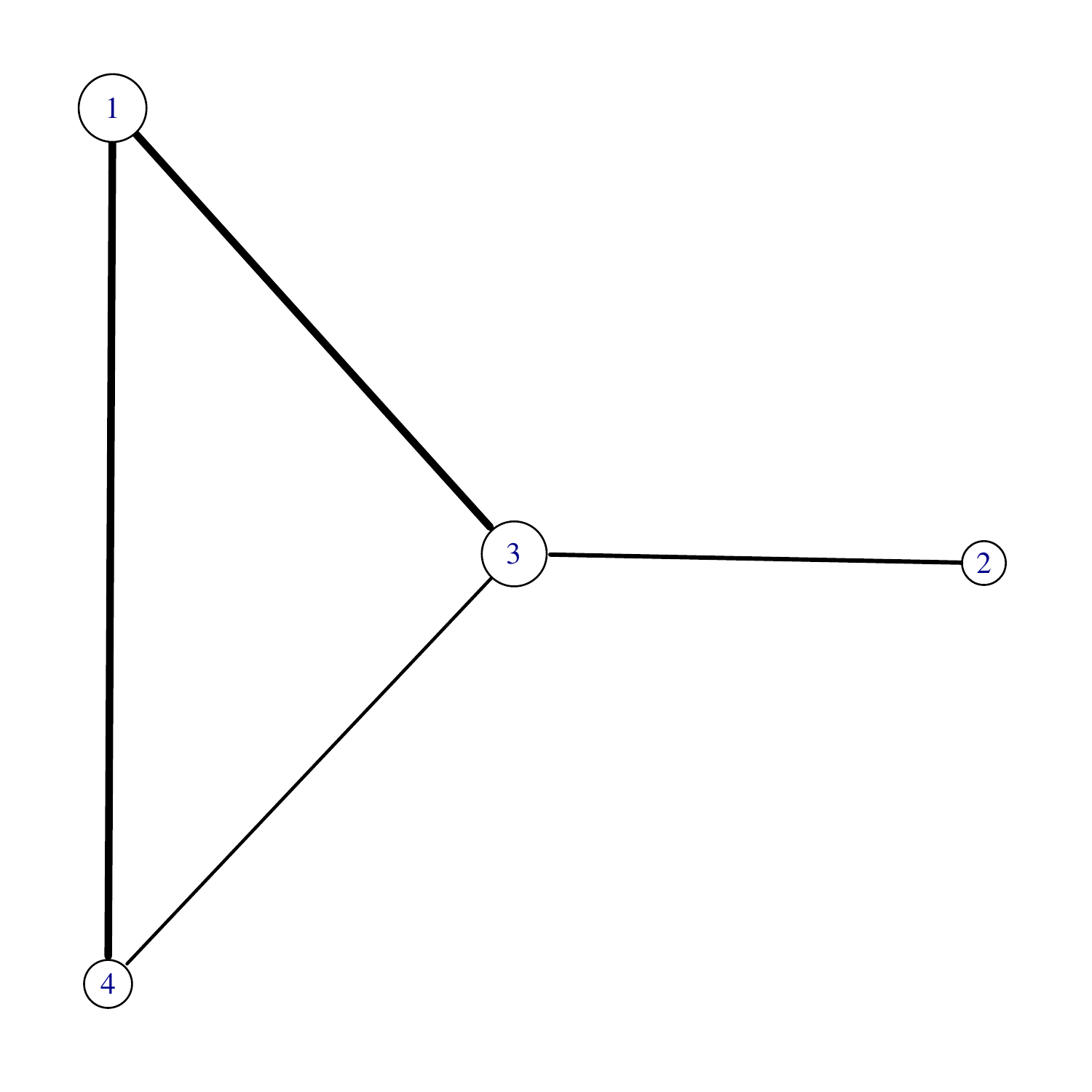}
\caption{Left: original layout numbered according to the clustering; right:
  clustering induced graph}  
\label{fig:Karate:induced:layout}
\end{figure}

It is well known from previous work on this graph that in terms of modularity,
the optimal number of clusters is four (see e.g., \cite{DuchArenas2005}). This
is a consequence of the structure of the modularity which tends to peak at a
certain graph specific number of clusters and then decreases (this is another
manifestation of the limited resolution of this measure
\cite{FortunatoBarthelemy2007}). For this graph, the social analysis conducted
by Zachary leads to the definition of two clusters which correspond to the
split between members of the club during the course of the study. Both
clusters are split into two sub-clusters by modularity maximization (one
ground truth cluster corresponds to clusters 1 and 4 in Figure
\ref{fig:Karate:induced:layout} and the other to clusters 2 and 3). We have
investigated the behavior of the deterministic annealing algorithm with $C$
ranging from $2$ to $8$. The parameters of the algorithm were the following
ones: $\alpha=1.1$, $\gamma$ is chosen so that the final temperature is
$T_0/10$ and $L=151$ (this relatively large value was used to obtained a
convergence for each EM-like phase; a faster annealing introduced
instabilities on this graph).

\begin{figure}[htbp]
  \centering
\includegraphics[width=0.9\textwidth]{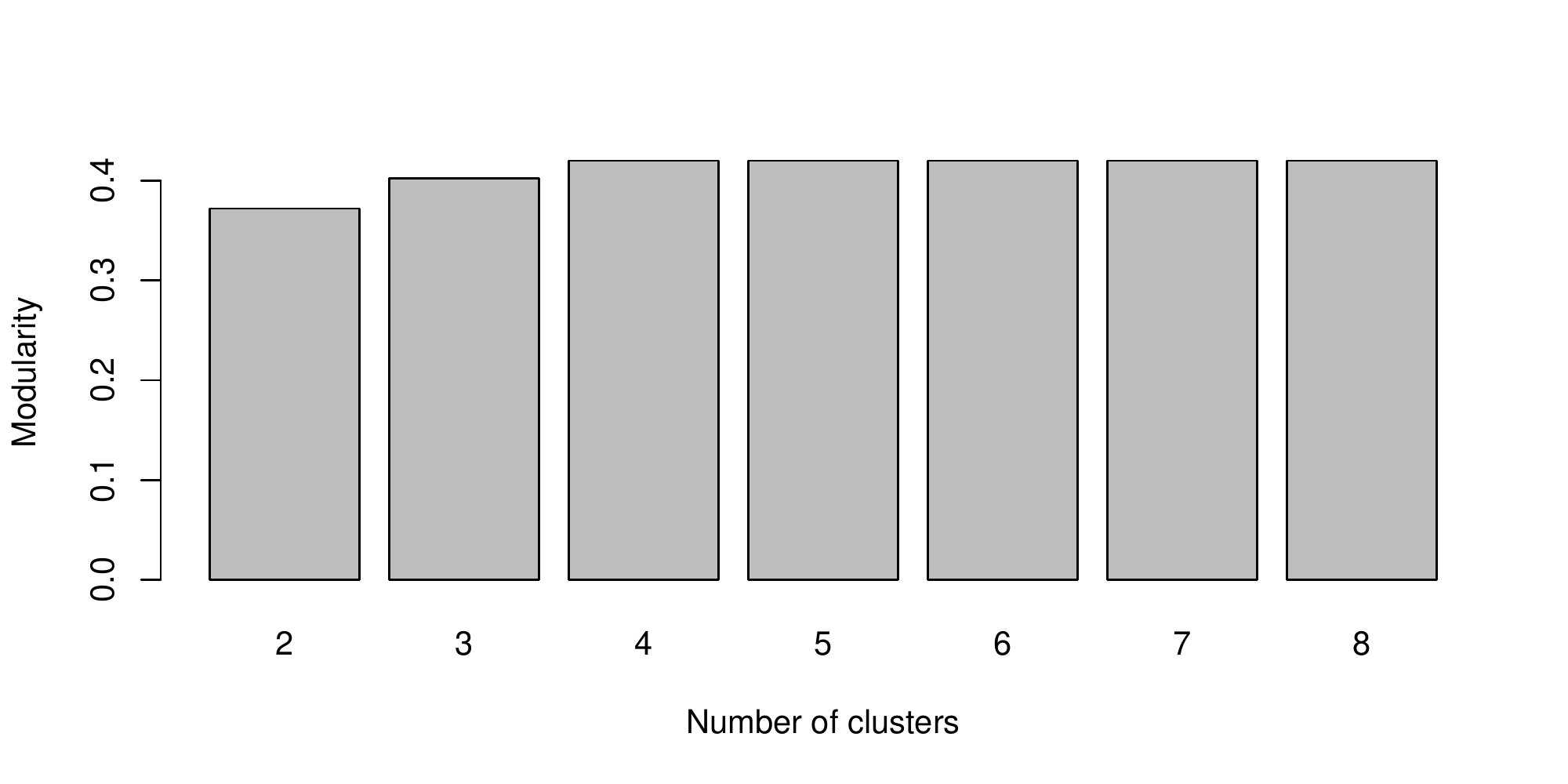}  
  \caption{Maximal modularity achieved by DA as a function of $C$}
  \label{fig:KarateModPerC}
\end{figure}

\begin{figure}[hbtp]
  \centering
\includegraphics[width=0.9\textwidth]{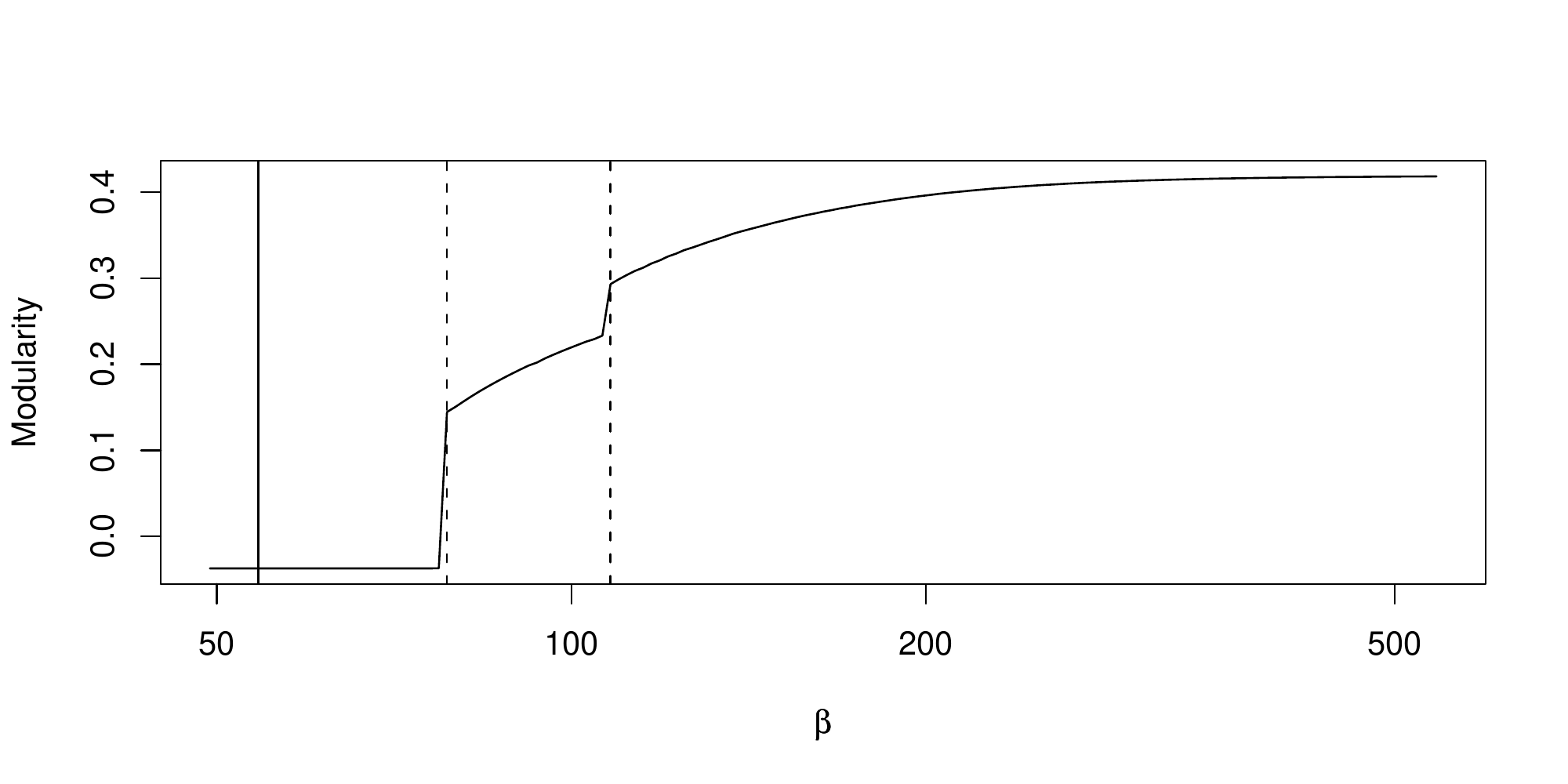}  
  \caption{Evolution of the modularity during annealing (the solid line
    corresponds to the theoretical temperature for the first phase
    transitions, dashed lines correspond to detected transitions)}
  \label{fig:AnnealingFourClusters}
\end{figure}

As shown on Figure \ref{fig:KarateModPerC}, the maximal modularity achieved by
deterministic annealing (without a prior structure) reaches its maximum at
$C=4$ clusters and then remains constant (the value of the maximum is
$0.4198$, as reported in e.g., \cite{DuchArenas2005}). This is an interesting
feature of DA: the algorithm is able to produce empty clusters if this
increases the objective function. In the present case, the algorithm produces
a maximum of four non empty clusters, even if we ask for eight. While this is
interesting in terms of maximizing the modularity, this tends to increase the
resolution problems faced by this quality measure
\cite{FortunatoBarthelemy2007}. The fuzzy layout strategy proposed in Section
\ref{subsection:fuzzylayout} is therefore expected to be very useful in this
context. 

\begin{figure}[hbtp]
  \centering
\includegraphics[width=0.9\textwidth]{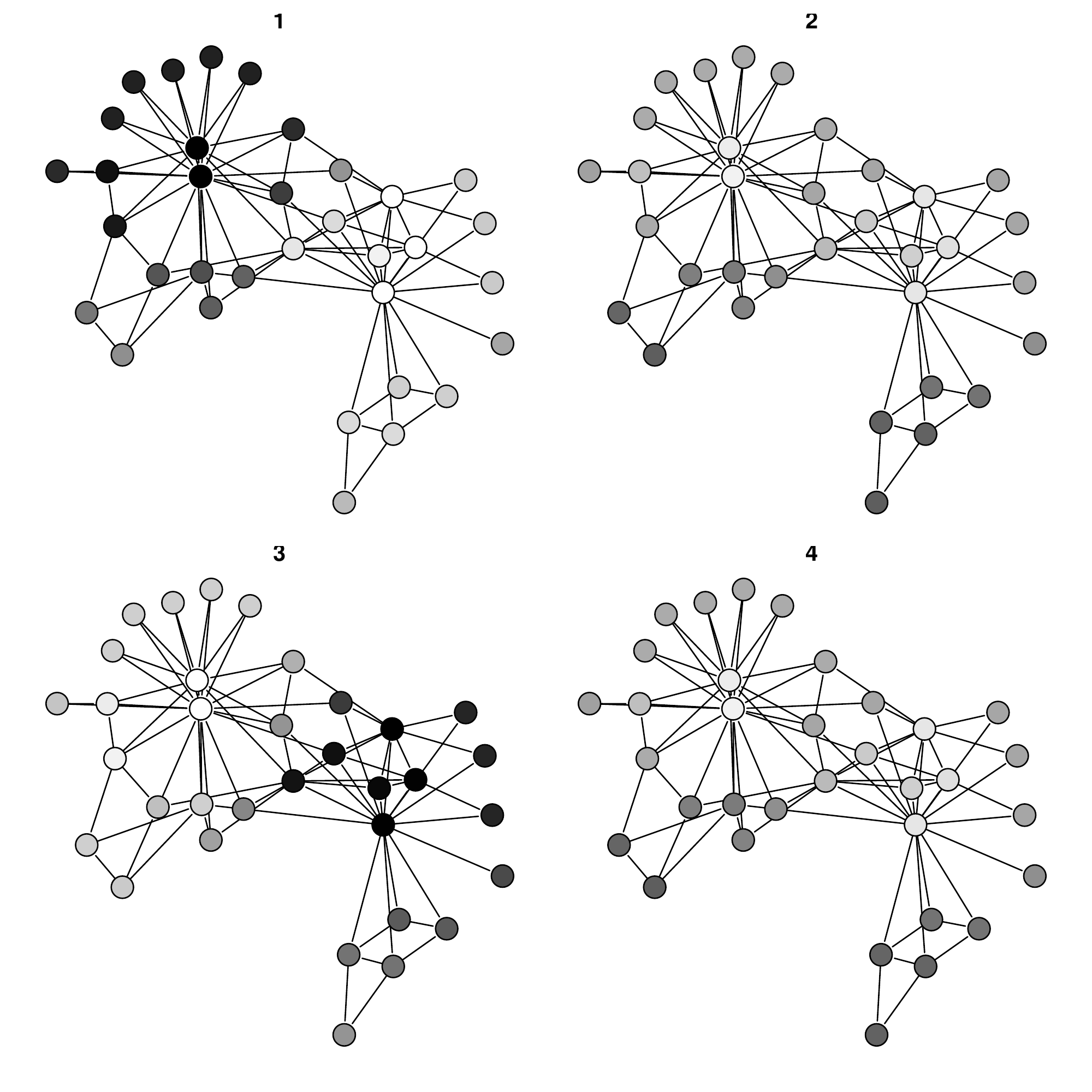}  
  \caption{Assignment probabilities after the first phase transition: black
  corresponds to probabilities close to one, white to probabilities close to
  zero}
  \label{fig:KarateEM:FPT}
\end{figure}

\begin{figure}[hbtp]
  \centering
\includegraphics[width=0.9\textwidth]{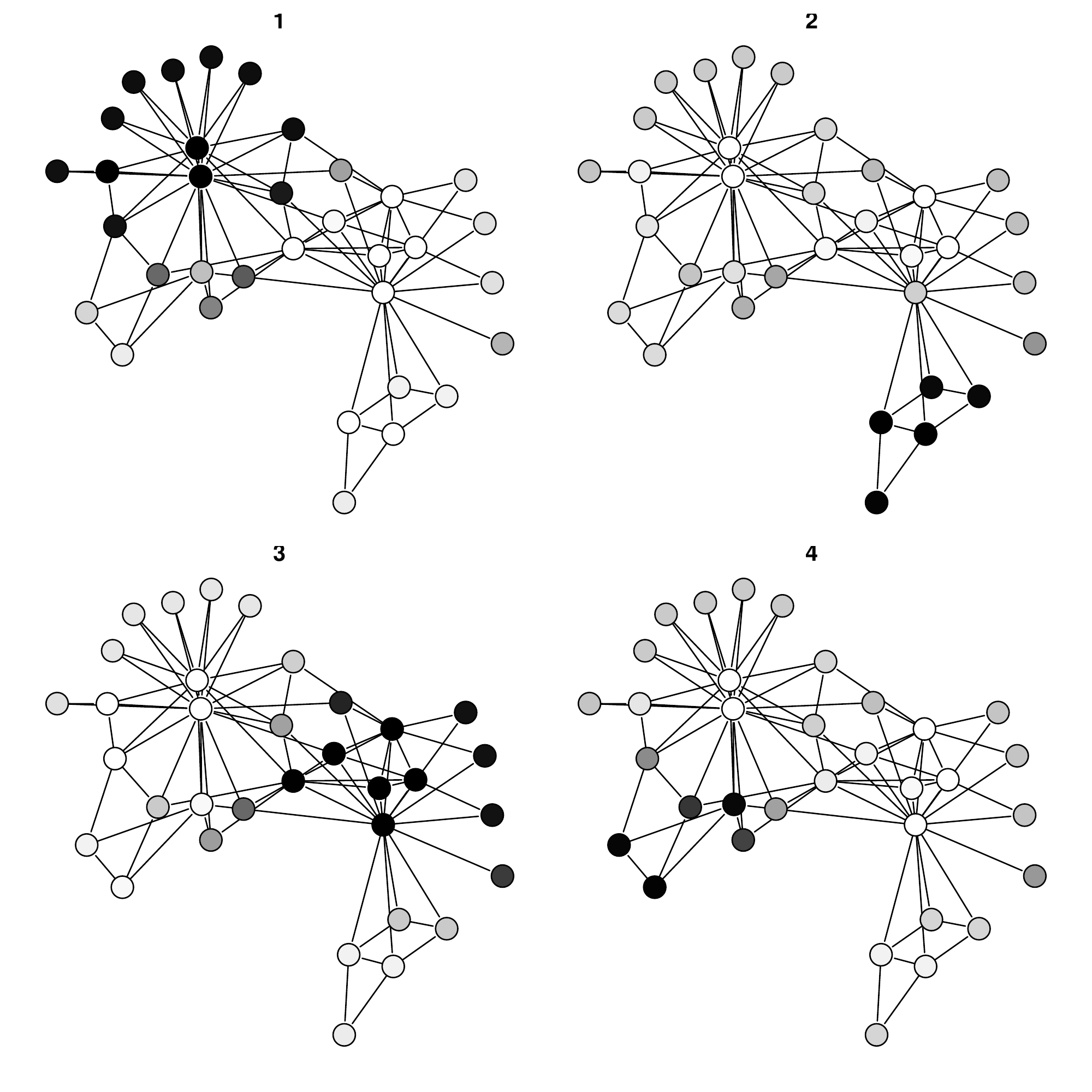}  
  \caption{Assignment probabilities after the second phase transition: black
  corresponds to probabilities close to one, white to probabilities close to
  zero}
  \label{fig:KarateEM:SPT}
\end{figure}

Figure
\ref{fig:AnnealingFourClusters} shows the evolution of the modularity during
annealing for the case $C=4$. More precisely, we compute the expectation of
the modularity with respect to the mean field distribution:
\begin{equation}
  \label{eq:ModularityExpectation}
\espb{R}{Q(M)}=\sum_{i\neq j}\sum_k\espb{R}{M_{ik}}\espb{R}{M_{jk}}B_{ij}+\frac{1}{2m}\sum_{i}\left(W_{ii}-P_{ii}\right)
\end{equation}
The critical temperature appears, as expected, as a very conservative estimate
of the temperature of the first phase transition. This first phase transition
corresponds to the detection of two well identified clusters (the ones
discovered by Zachary in \cite{Zachary77Karate}), while the second phase
transition introduces two additional clusters. The evolution of the
expectation of the assignment matrix $\espb{R}{M}$ is represented by Figures
\ref{fig:KarateEM:FPT} and \ref{fig:KarateEM:SPT}: each figure includes four
copies of the graph. In copy number $k$, the gray level of the node $i$
encodes the value of $\espb{R}{M_{ik}}$, i.e., of the probability for node $i$
to belong to cluster $k$ according to the approximating distribution $R$.
After the first phase transition, clusters 1 and 3 start to take (partial)
ownership of some of the vertices (the black ones on Figure
\ref{fig:KarateEM:FPT}), even if only 6 vertices out of 34 have a maximal
assignment probability higher than 0.75. More than 40\% of the vertices (14
out of 34) have rather fuzzy assignment probabilities (i.e., less than 0.5 for
the maximal value). 
After the second phase transition (shown on Figure \ref{fig:KarateEM:SPT}),
the algorithm has identified four distinct clusters accounting for 23 vertices
(which are all assignments to a cluster with more than 0.75 probability). Four
vertices still have a maximal assignment probability below 0.5. If the
annealing is brought to the limit of a very low temperature (below the
threshold of $T_0/10$ chosen here), the expectation of the assignment matrix
is almost a 0/1 matrix, but such effort is not needed: in the case of the
Karate club graph, all vertices have a maximal assignment probability higher
than 0.5 at $T=T_0/10$ and a winner-take-all approach leads to a well defined
clustering of the graph. 

Figure \ref{fig:Karate:induced:layout} compares the original layout numbered
via the clustering to the layout of the clustering induced graph (obtained by the
Fruchterman-Reingold force directed algorithm
\cite{FruchtermanReingoldGraph1991}). The induced graph emphasizes clearly the
relation between the clusters: cluster 2 is not directly connected to clusters
1 and 4, while the connection between 3 and 4 is less dense than the ones
between 1 and 3 or 4. The summary is therefore quite informative, even if, as
expected, most of the structure is lost. For instance, cluster 1 exhibits a
start shape which is of course lost in the glyph based representation. 

\subsection{Organized modularity maximization}
\begin{figure}[htbp]
  \centering
\includegraphics[width=0.9\textwidth]{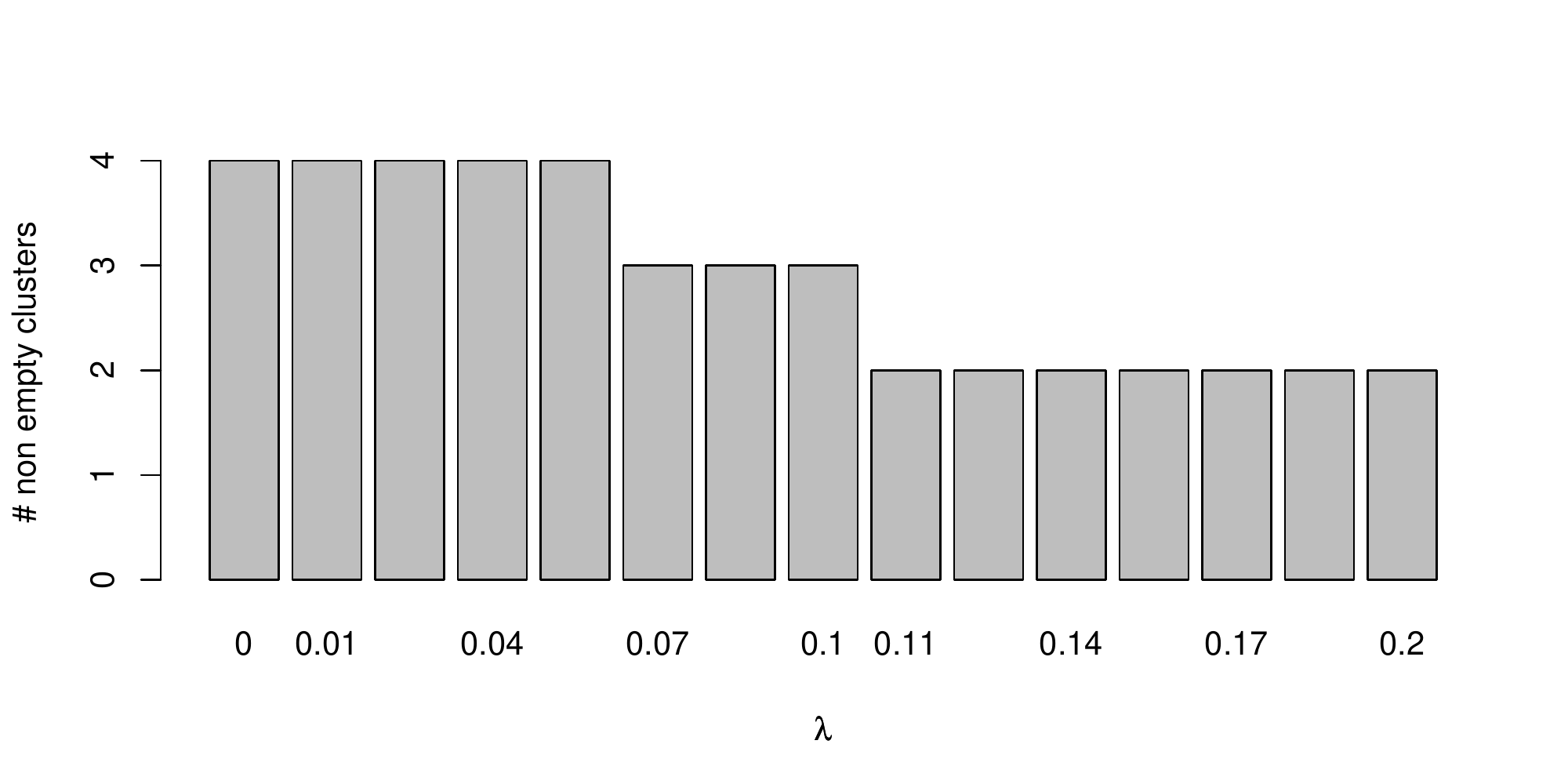}  
  \caption{Number of non empty clusters as a function of the influence parameter
$\lambda$}
  \label{fig:KarateSoftLambda}
\end{figure}

For the organized modularity, we use the simplest setting compatible with the
four clusters identified in the previous section: the prior structure consists
in a square in which each vertex is associated to a cluster. We use the
following influence matrix:
 \[
S=\left(
  \begin{array}{cccc}
1 & \lambda & \lambda & 0 \\
\lambda & 1 & 0 & \lambda  \\
\lambda & 0 & 1 & \lambda \\
0 &\lambda & \lambda & 0\\
  \end{array}
\right),
\]
where $\lambda$ specifies the amount of local influence (there is no diagonal
influence). Interestingly, the value of $\lambda$ has an impact on the number
of non empty clusters obtained by deterministic annealing. We use the same
parameters for the algorithm as in the previous section, while varying
$\lambda$ between $0$ and $0.2$. As shown on Figure
\ref{fig:KarateSoftLambda}, a large influence tends to reduce the number of
effective clusters produced by the algorithm. 
This can be explained by an analysis of the optimal four clusters clustering
obtained before. As shown on Figure \ref{fig:Karate:induced:layout}, there is
no direct connection between cluster 2 and clusters 1 and 4. As a consequence,
the corresponding entries $B_{ij}$ are negative: the only way to prevent the
organized modularity to be smaller than the standard one is to put cluster 2
far away from clusters 1 and 4 in the prior structure. This is not strictly
feasible as there are only two pairs of cluster positions with no direct
influence in $S$ (they corresponds to the diagonal of the square). As a
consequence, one might put cluster 1 far away from cluster 1 or from cluster
4, but not from both. Therefore, the modularity will decrease by e.g.,
\[
-\frac{1}{2m}\sum_{i\in C_2,j\in C_4}\lambda P_{ij},
\]
if cluster $2$ has no influence on cluster $1$. Then, when $\lambda$ is large
enough, the reduction of the four clusters organized modularity will be high
enough to bring its value below the one of the two or three clusters standard
modularity. Those latter configurations are easier to arrange on the prior
structure in a way that minimizes negative contributions: they will be
preferred by the algorithm over the four clusters solution. 

\begin{figure}[htbp]
  \centering
\includegraphics[width=0.47\textwidth]{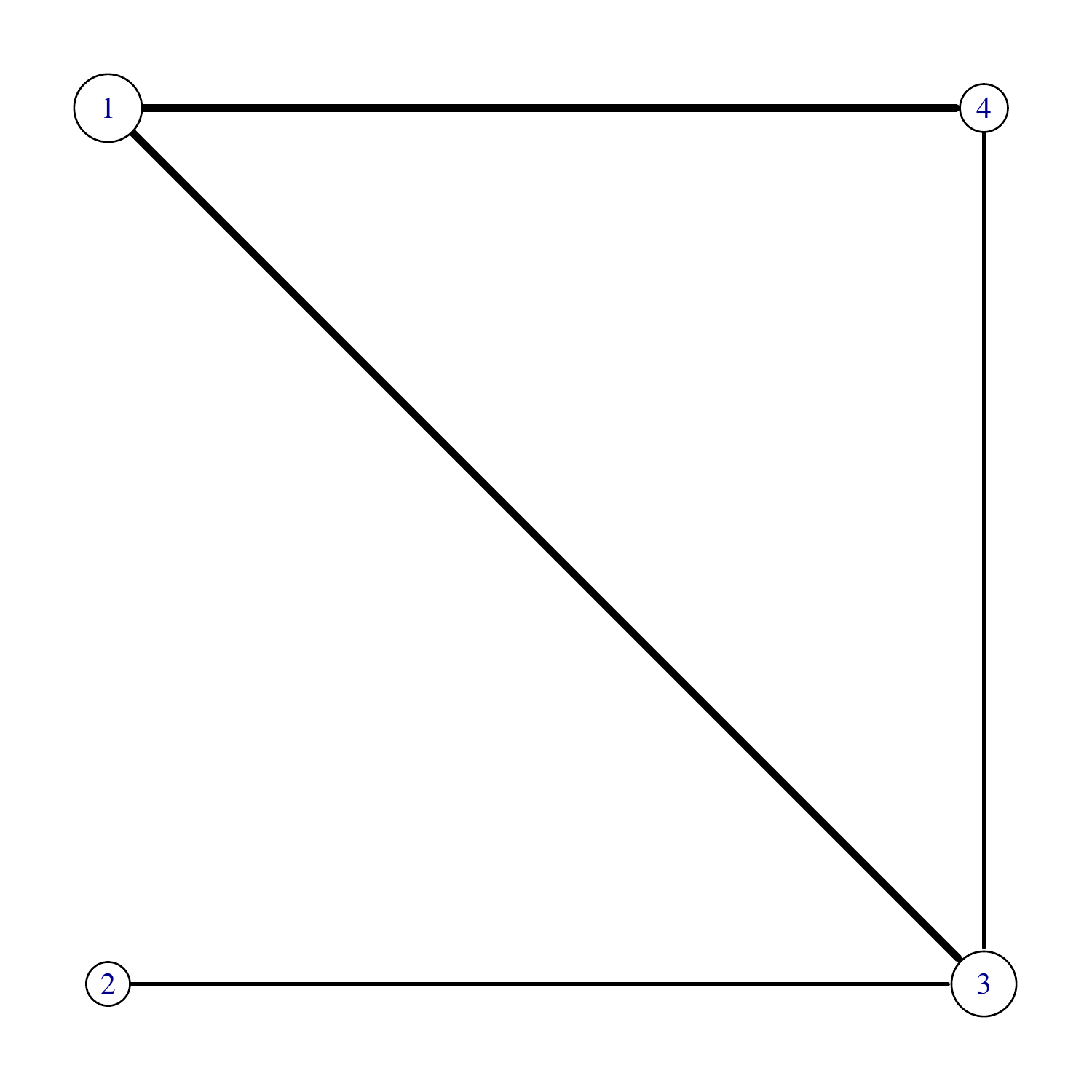}
\caption{clustering induced graph displayed on the prior structure}  
\label{fig:Karate:soft:induced:layout}
\end{figure}

In fact, the behavior of the algorithm is comparable to the one of a force
directed method: when there is no connection between two vertices, they only
repel each other and the algorithm tries to maximize their distances. As shown
on Figure \ref{fig:Karate:soft:induced:layout}, the representation of the
graph obtained via the prior structure is quite similar to the one provide by
Figure \ref{fig:Karate:induced:layout}. The underlying
clusters are completely identical and thus use the same numbering to ease
comparison between the figures (this is also the case when the value of
$\lambda$ induces a clustering with two or three clusters: the obtained
clusterings are the ones that maximize the standard modularity for $C=2$ or
$C=3$). 

\subsection{Fuzzy layout}
\label{karateFuzzy}
\begin{figure}[htbp]
  \centering
  \begin{tabular}{cc}
\includegraphics[width=0.47\textwidth]{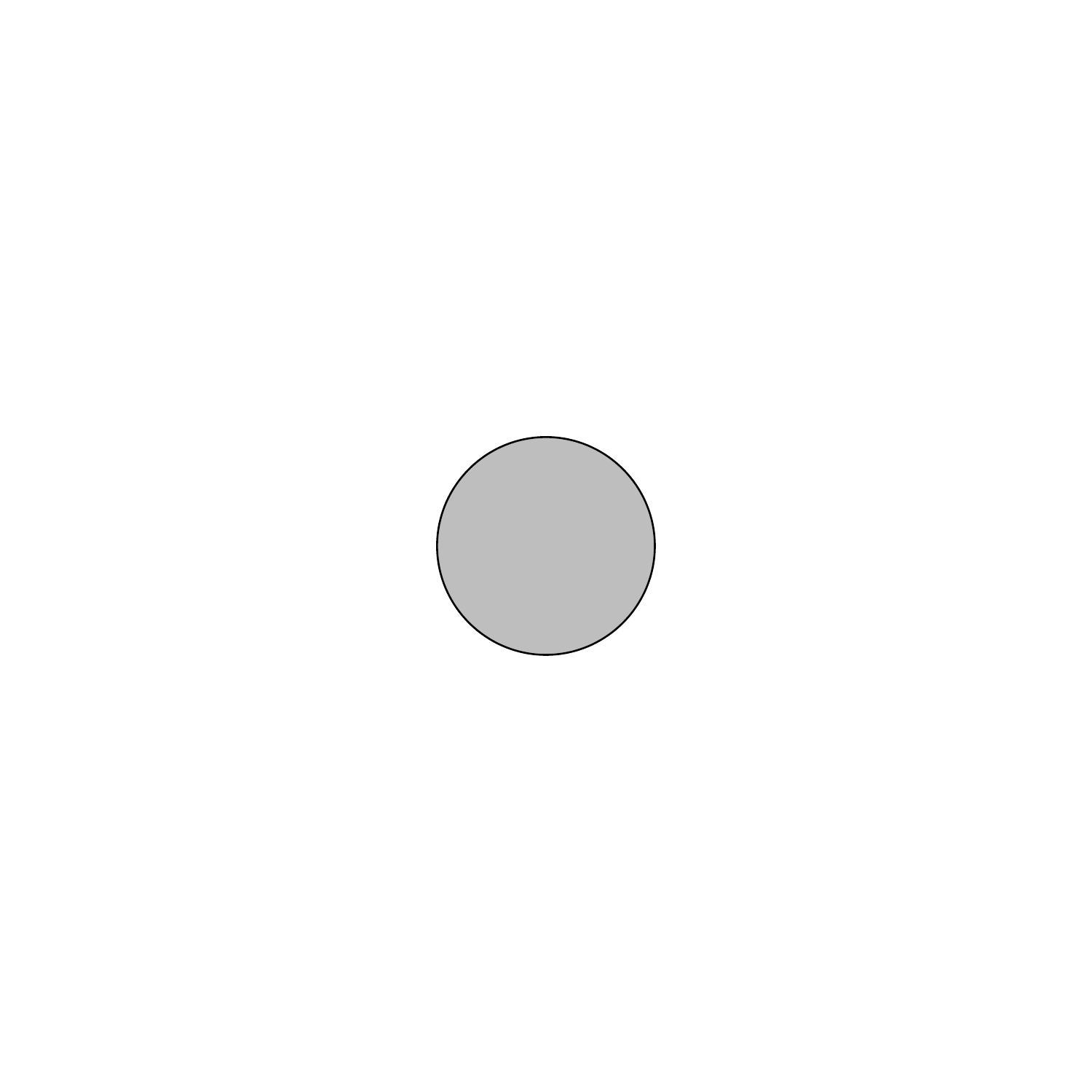}
&\includegraphics[width=0.47\textwidth]{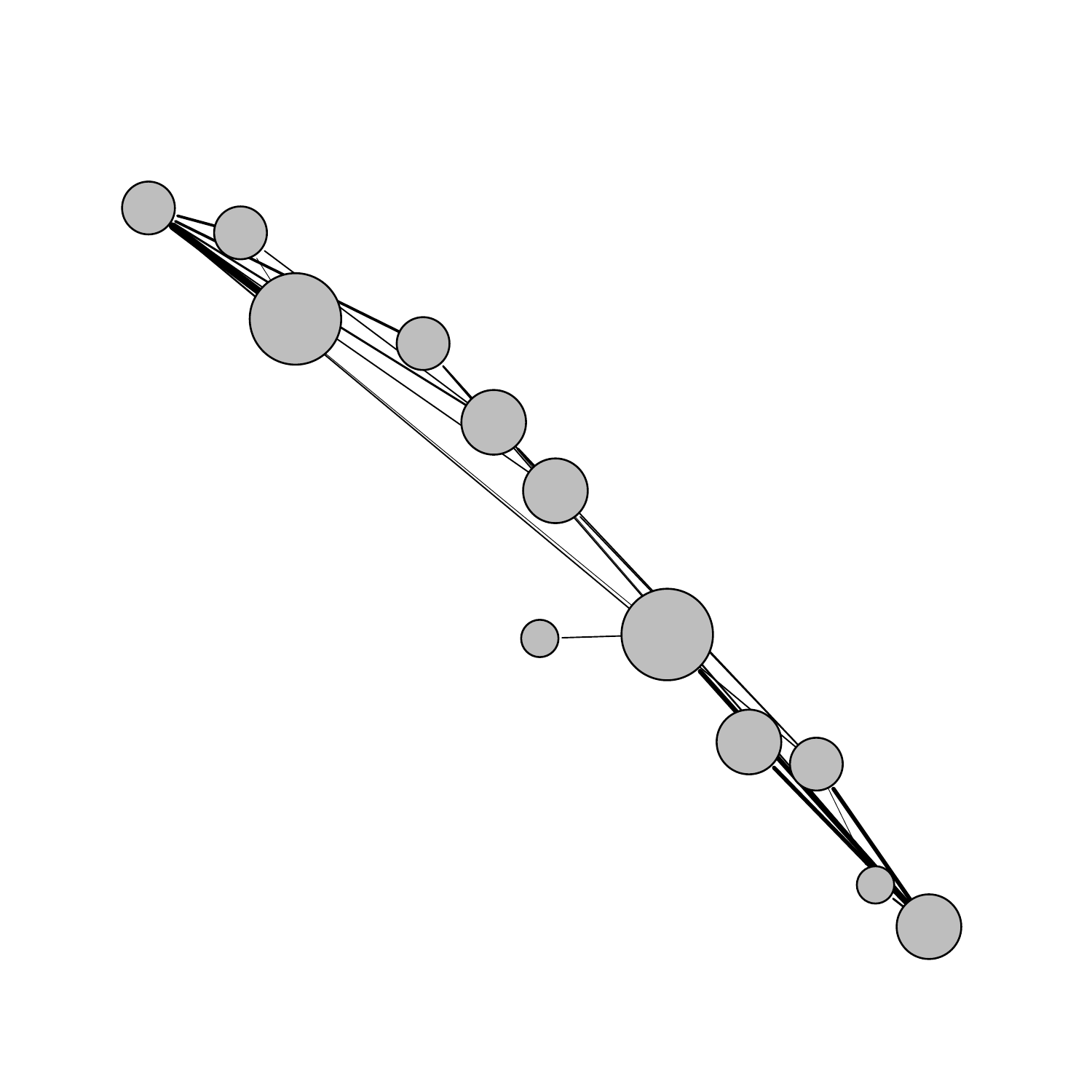}\\
initial layout & after first phase transition\\
\includegraphics[width=0.47\textwidth]{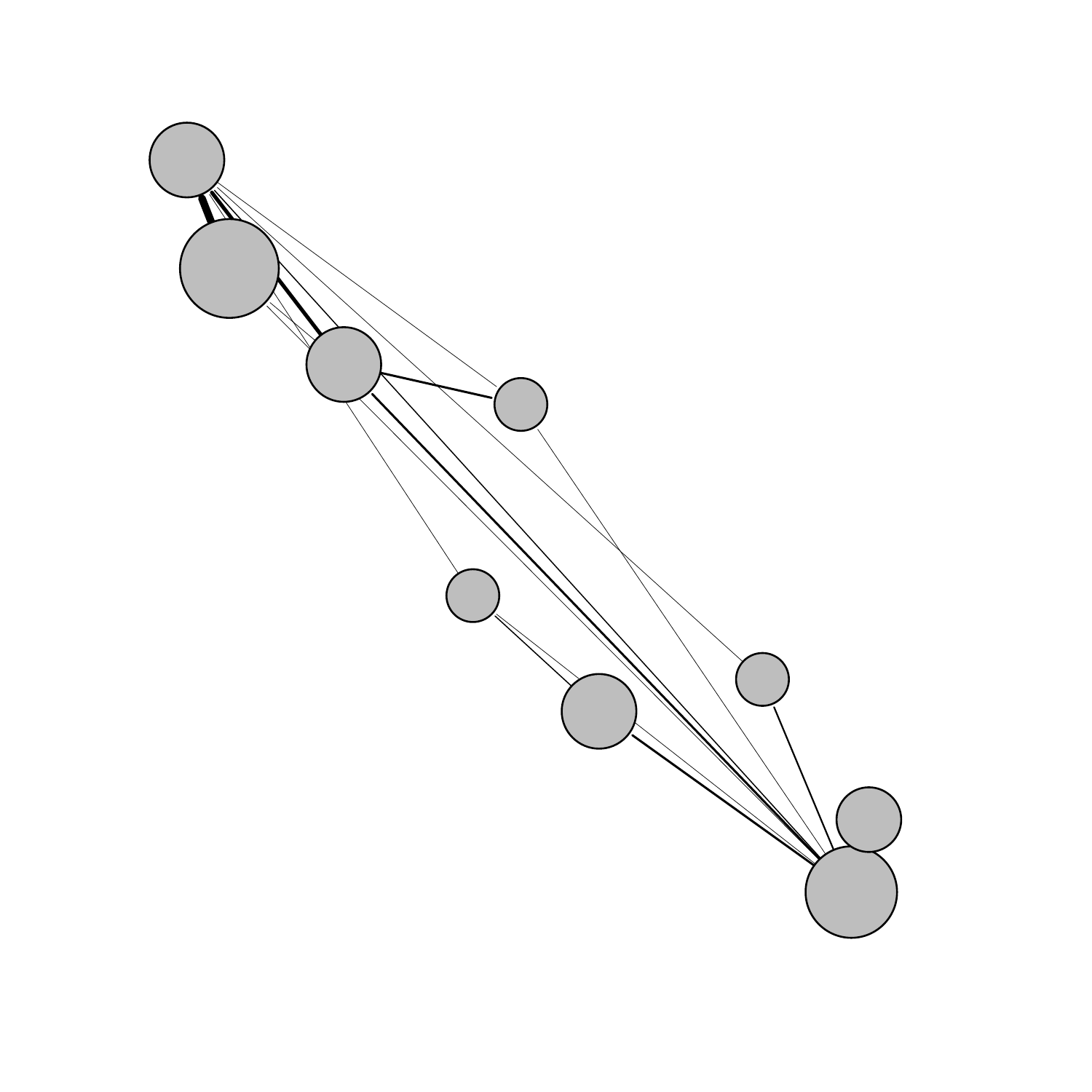}
&\includegraphics[width=0.47\textwidth]{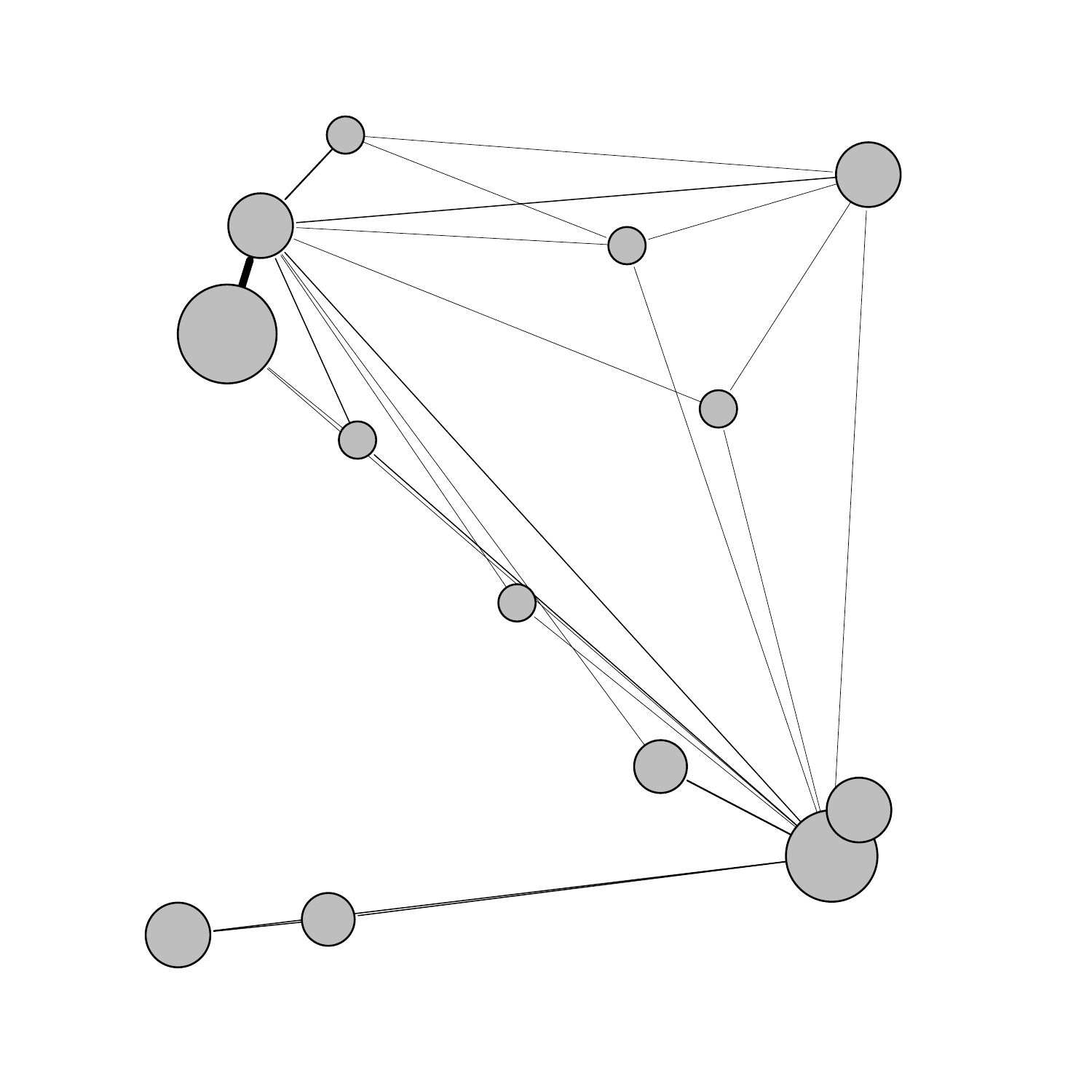}\\
just before second phase transition & just after second phase transition\\
\includegraphics[width=0.47\textwidth]{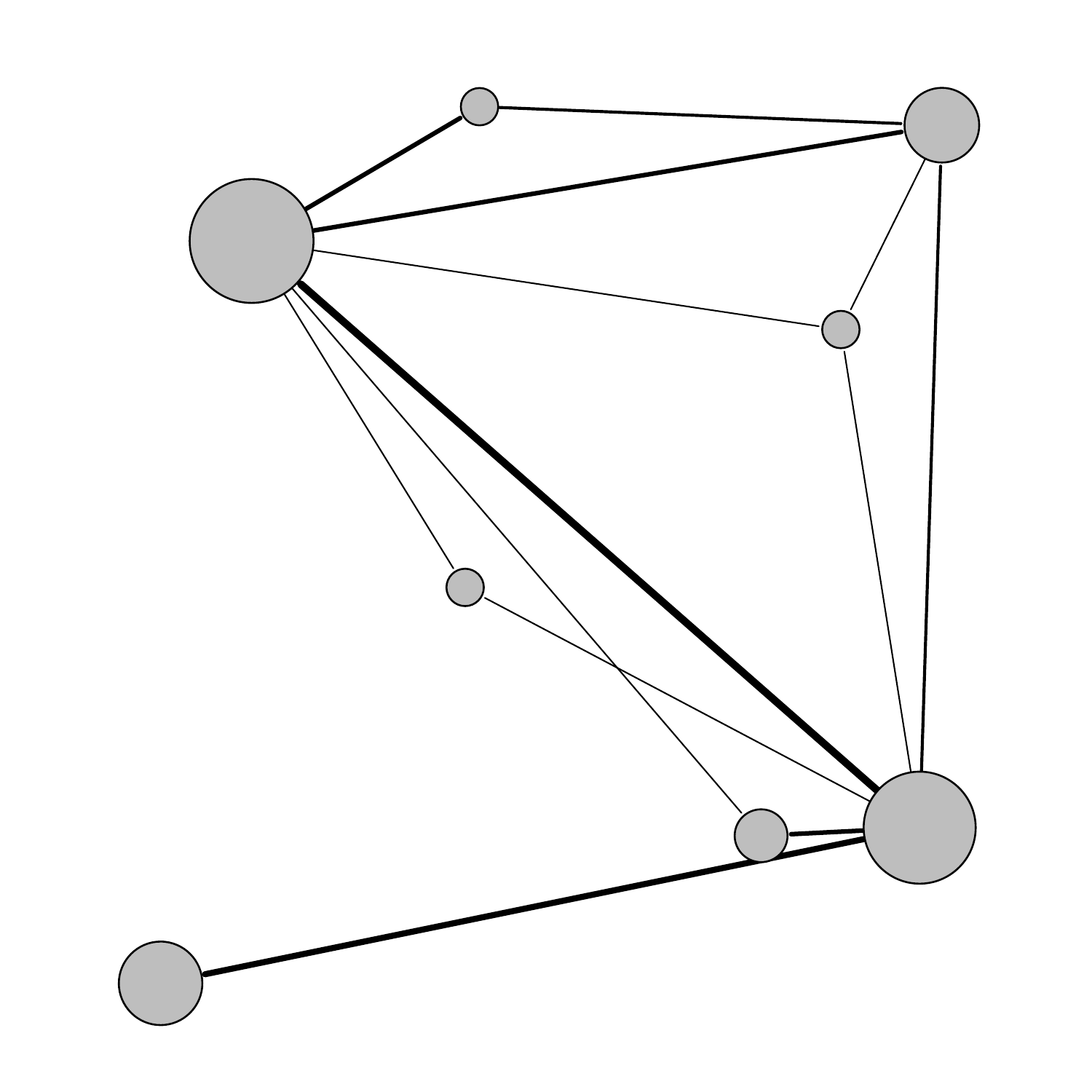}
&\includegraphics[width=0.47\textwidth]{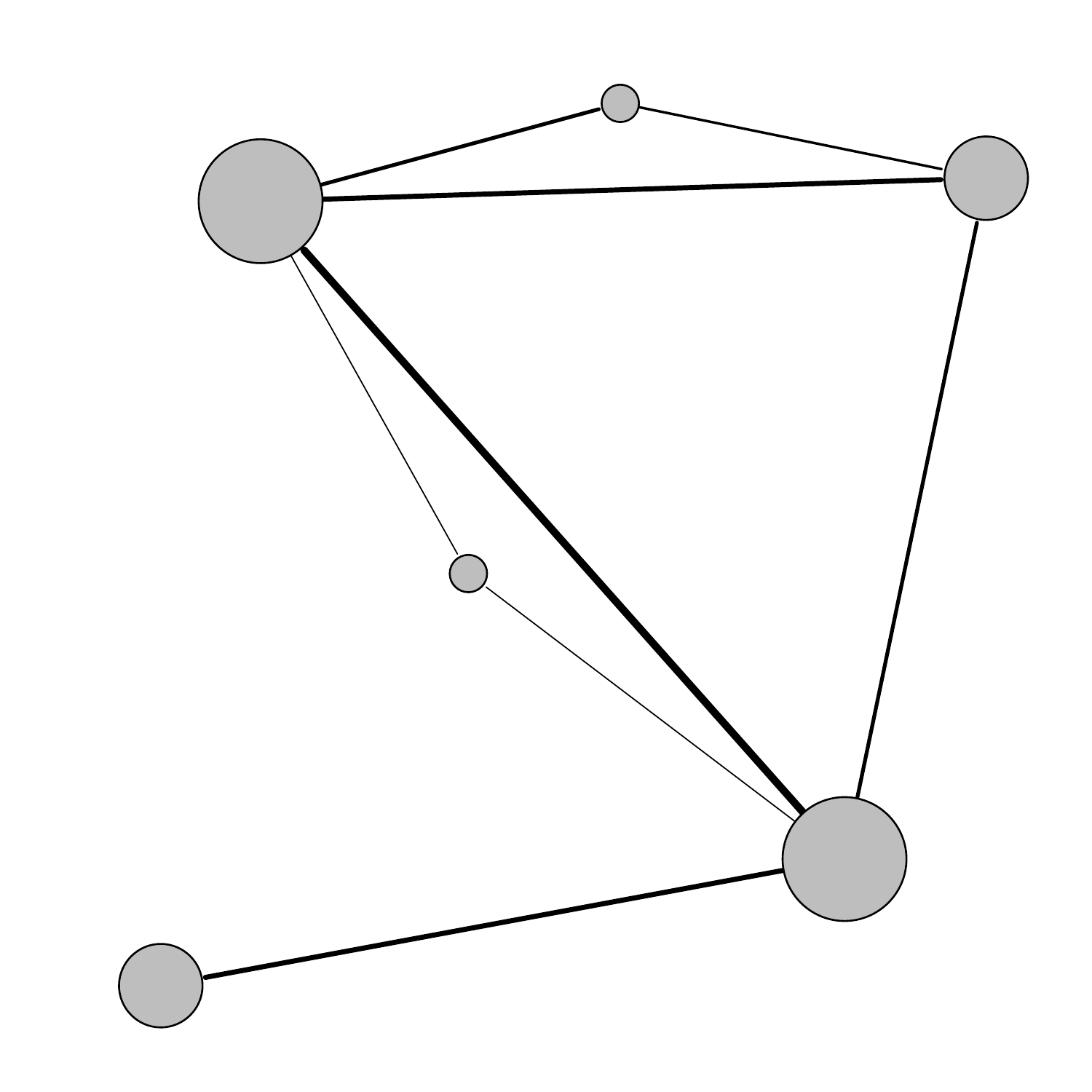}\\
during the final cooling phase & final configuration\\
  \end{tabular}
  \caption{Evolution of the fuzzy layout during annealing (the
    Fruchterman-Reingold algorithm was used to reduce edge overlapping)}
  \label{fig:FuzzyEvolutionKarate}
\end{figure}

It might seem from Figure \ref{fig:Karate:soft:induced:layout} and more
generally from the discussion above that the organized modularity offers no
particular interest compared to a two phases approach with a maximal modularity
clustering followed by a force directed layout. The behavior of the DA
algorithm on the Karate graph emphasizes the fact that the modularity does not
increase with the number of clusters: it tends to peak at an optimal graph
specific number. As explained before, there is a complex interaction between
the prior structure and the values of the modularity for different numbers of
non empty clusters. In some situations, the prior structure introduces a too
strong coupling between clusters and leads this way to a reduced number of non
empty clusters. This might lead to a better visualization of the graph, to an
over simplification or to nothing more than the two phases approach (Section
\ref{section:large:graphs} will illustrate this further).

The first way to address this problem is to test several prior structures and
to select optimal graphs with respect to different quality measures as
proposed in Section \ref{section:graph:visu}. A second (and complementary)
approach consists in leveraging the annealing process to limit the impact of
the peaking behavior of the modularity by means of the fuzzy layout strategy
described in Section \ref{subsection:fuzzylayout}. An example of such layouts
is given by Figure \ref{fig:FuzzyEvolutionKarate}. The representations give a
more complete picture of the graph than Figure \ref{fig:Karate:induced:layout}
(or Figure \ref{fig:Karate:soft:induced:layout}). It appears for instance
quite clearly that two vertices are not easy to assign to the final four
clusters (they are at the boundary of cluster 1 in Figure
\ref{fig:Karate:soft:induced:layout}). If we study carefully the original
representation on Figure \ref{fig:Karate} with the added knowledge provided by
Figures \ref{fig:Karate:induced:layout} and \ref{fig:FuzzyEvolutionKarate},
those vertices (number 10 and 24) appear quite clearly in between two
clusters. However, it would be almost impossible to obtain this information
directly from the original
representation. Figure \ref{fig:FuzzyEvolutionKarate} shows also that the
composition of cluster 3 (in Figure \ref{fig:Karate:soft:induced:layout},
bottom left in Figure \ref{fig:FuzzyEvolutionKarate}) is quite obvious
compared to others. This is easily confirmed on Figure
\ref{fig:Karate:induced:layout}. 

\subsection{Comparison with SOM variants}\label{subsectionKarateSOM}
As the proposed method is based on the SOM rationale, it seems natural to
compare it to SOM variants that can handle graph nodes, mainly the kernel SOM
with adapted kernels \cite{VillaRossiKSOMWSOM2007} and the spectral SOM
\cite{VillaEtAlMashs2008}. Details on those methods and on parameters
optimization are postponed to Section \ref{DescExperiments} in order to keep
the present Section focused on a detailed analysis of the Karate graph. 

Following the methodology described in Section \ref{DescExperiments}, we have
built Pareto optimal points with respect to the modularity of the clustering
and the number of edge crossings in the induced layout. The only global Pareto
optimal point among SOM variants has been obtained by the heat kernel (see
\cite{kondor_lafferty_ICML2002}). It has
zero edge crossing and a modularity of 0.4188. The obtained layout is
identical to one of Figure \ref{fig:Karate:soft:induced:layout}, up to a
rotation. However, the underlying clustering is slightly different, as shown
by the reduced modularity. In fact, node number 10 on Figure \ref{fig:Karate}
is assigned to the same cluster as node number 3 rather than to node number 34's
cluster in the optimal clustering (in terms of modularity). It turns out that
node number 10 should be assigned to node number 34's cluster according to
Zachary's analysis. Therefore, the best SOM variant fails to recover exactly
the ground truth clustering, while modularity optimization does recover a
finer clustering.

Other variants make similar or worse mistakes. For instance, the best result
obtained by the modularity kernel SOM has a modularity of 0.409 which
corresponds to two differences with the optimal modularity clustering. Node
number 10 is misclassified compared to Zachery's two clusters ground truth,
while node number 24 is assigned to the correct Zachery's cluster but to a
different cluster than in the highest modularity clustering. 

\begin{figure}[htbp]
  \centering
\includegraphics[width=0.6\textwidth]{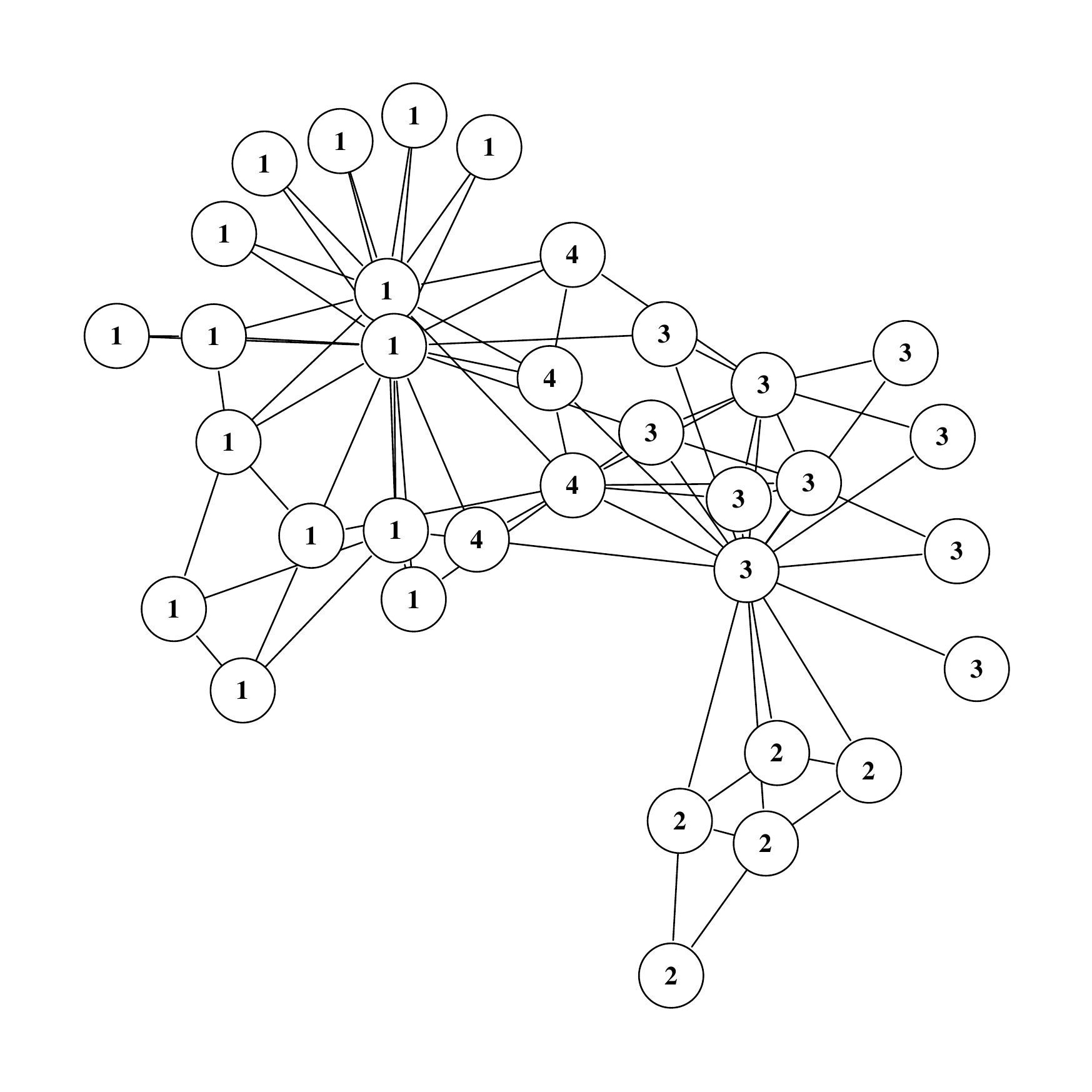}
\caption{Original layout numbered according to the clustering obtained by the
  Laplacian's inverse kernel SOM}  
\label{fig:Karate:induced:layout:Ginv}
\end{figure}

The best solution for the Laplacian's inverse kernel SOM has an even worse
modularity of 0.391. In this case, the differences between the proposed
clustering and the ground truth are more important. Figure
\ref{fig:Karate:induced:layout:Ginv} gives the clustering obtained by the
Laplacian's inverse kernel SOM on the full layout of the Karate graph. Node
number 3 is assigned to a wrong cluster according to the ground truth. While
the mismatch corresponds to only one node, the status of this node is much
more obvious than the one of node 10: it is connected to node number 1, the
central actor one of the clusters identified sociologically. In addition,
cluster number 4 seems to consist in boundary nodes rather than in a dense
subset of nodes. While this SOM variant manages to reach a fair approximation
of the ground truth clustering, it splits one of those clusters in a
misleading way: cluster 4 is not dense and its members are more connected to
outside nodes than between themselves. 

In summary, the SOM variants are unable to recover exactly the ground truth
clustering obtained by Zachary on this graph. In addition, the SOMs give
an example of a clustering with reduced modularity (compared to optimal ones)
with a possibly misleading cluster: the subgraph is not dense and nodes have
more connections outside of the cluster than inside. Limitations of graph SOMs compared to our proposal
will be confirmed in the next Section. 

\section{Results on larger graphs}\label{section:large:graphs}
\subsection{Datasets}
\begin{figure}[htbp]
	\begin{tabular}{cc}
  	\includegraphics[width=0.55\textwidth,height=0.6\textwidth]{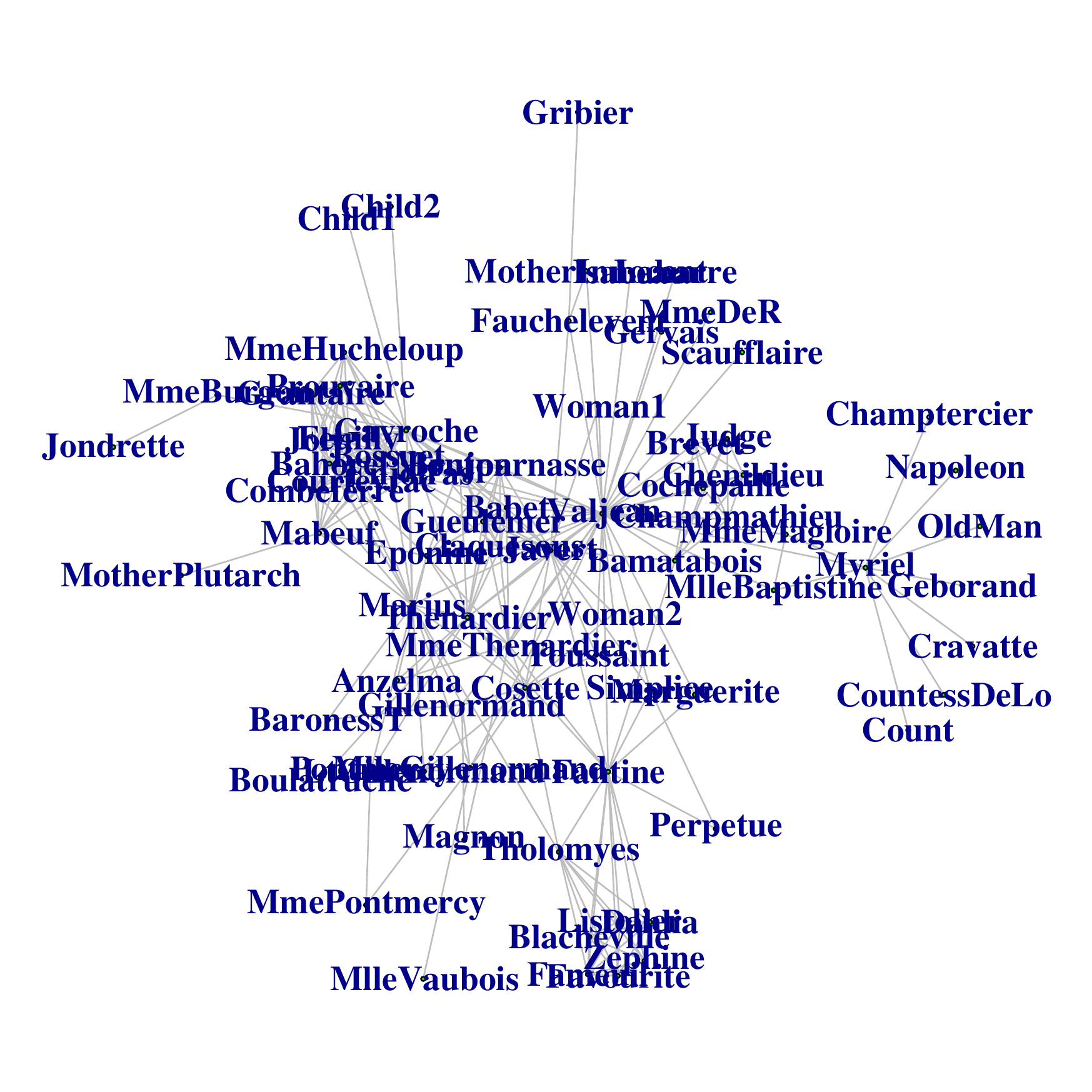} & \includegraphics[width=0.4\textwidth,height=0.45\textwidth]{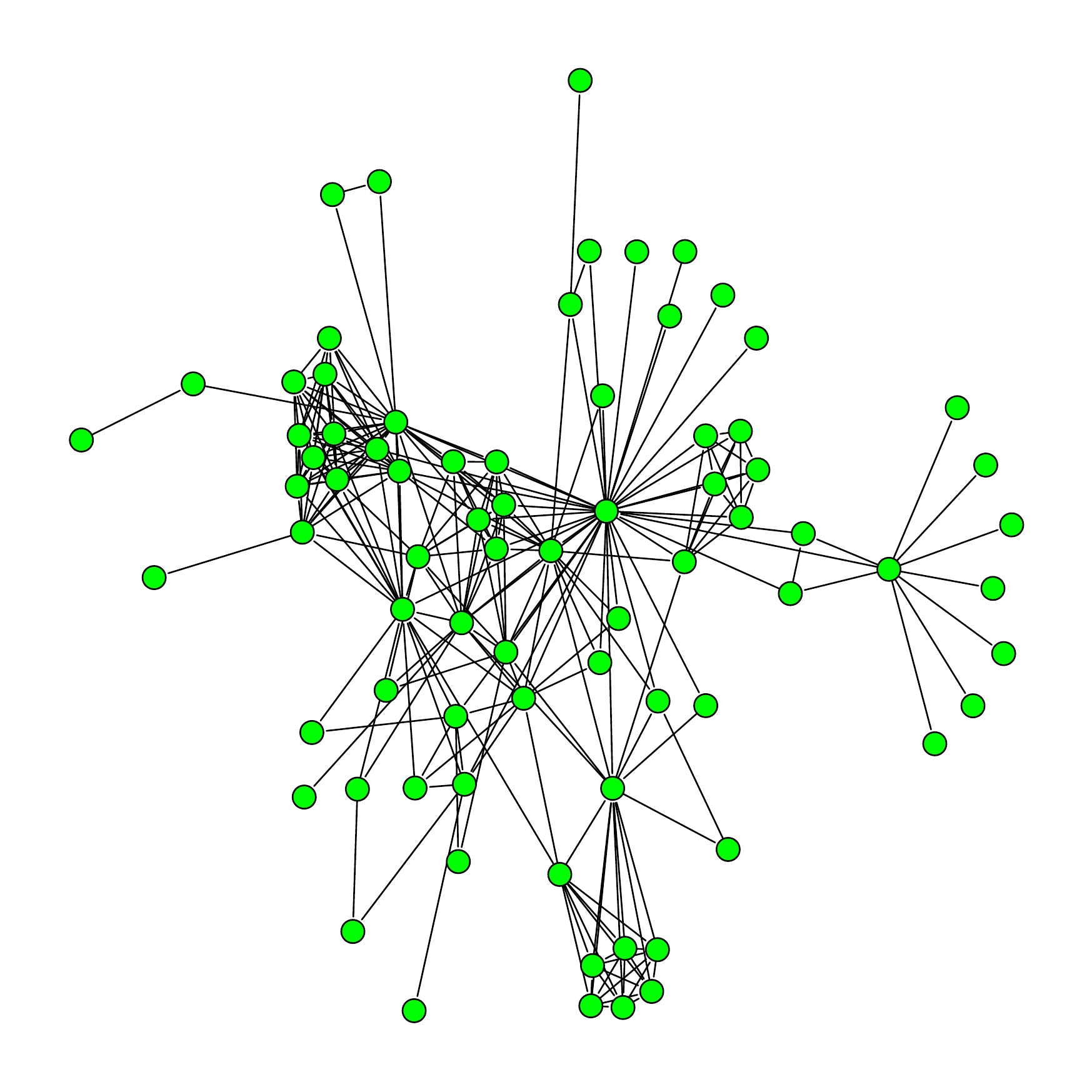}
	\end{tabular}
	\caption{Coappearance network from ``Les Misérables'' (left: with characters' names; right: without label)}
	\label{fig::miserables}
\end{figure}

\begin{figure}[htbp]
	\centering
	\begin{tabular}{cc}
		\includegraphics[width=0.5\textwidth]{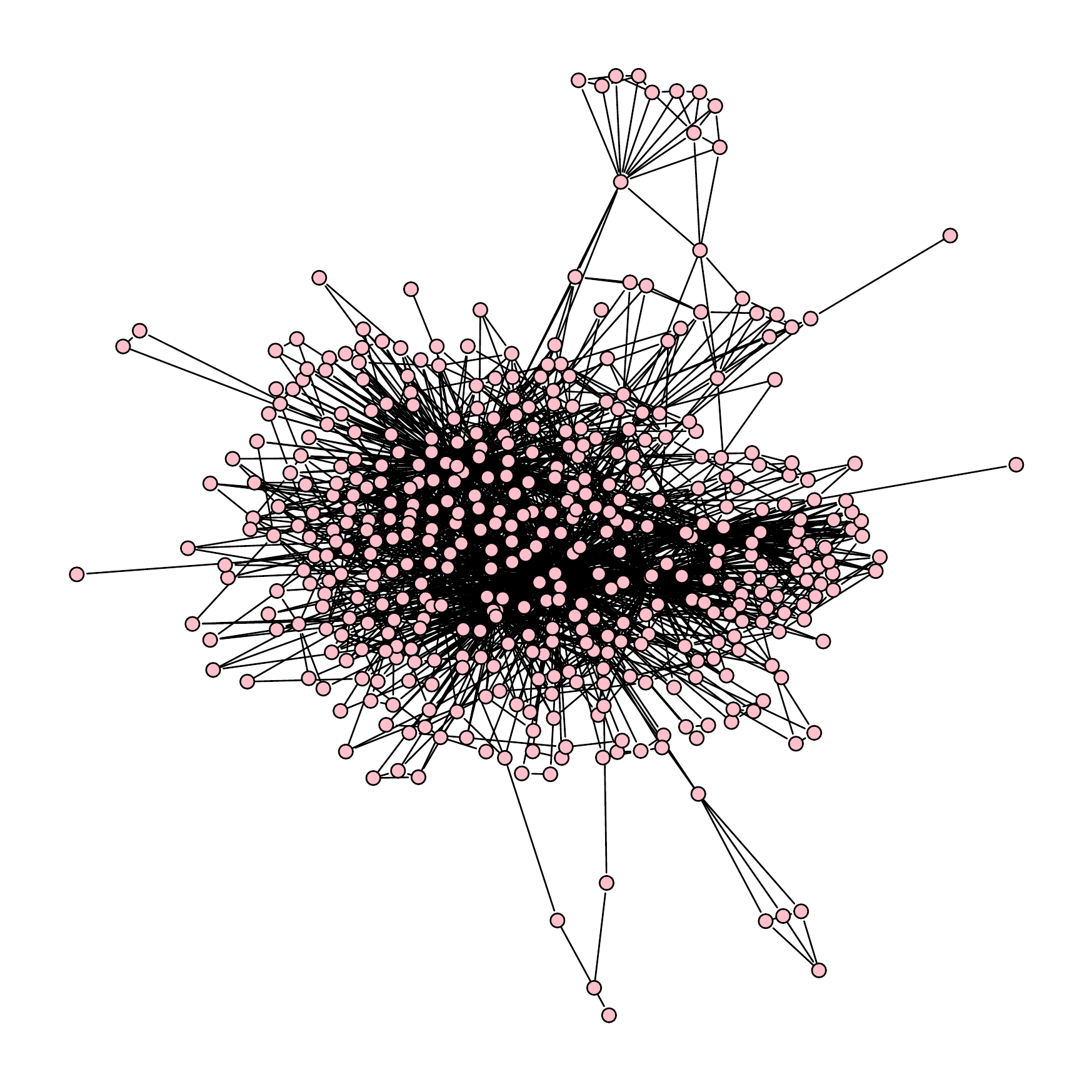} & \includegraphics[width=0.5\textwidth]{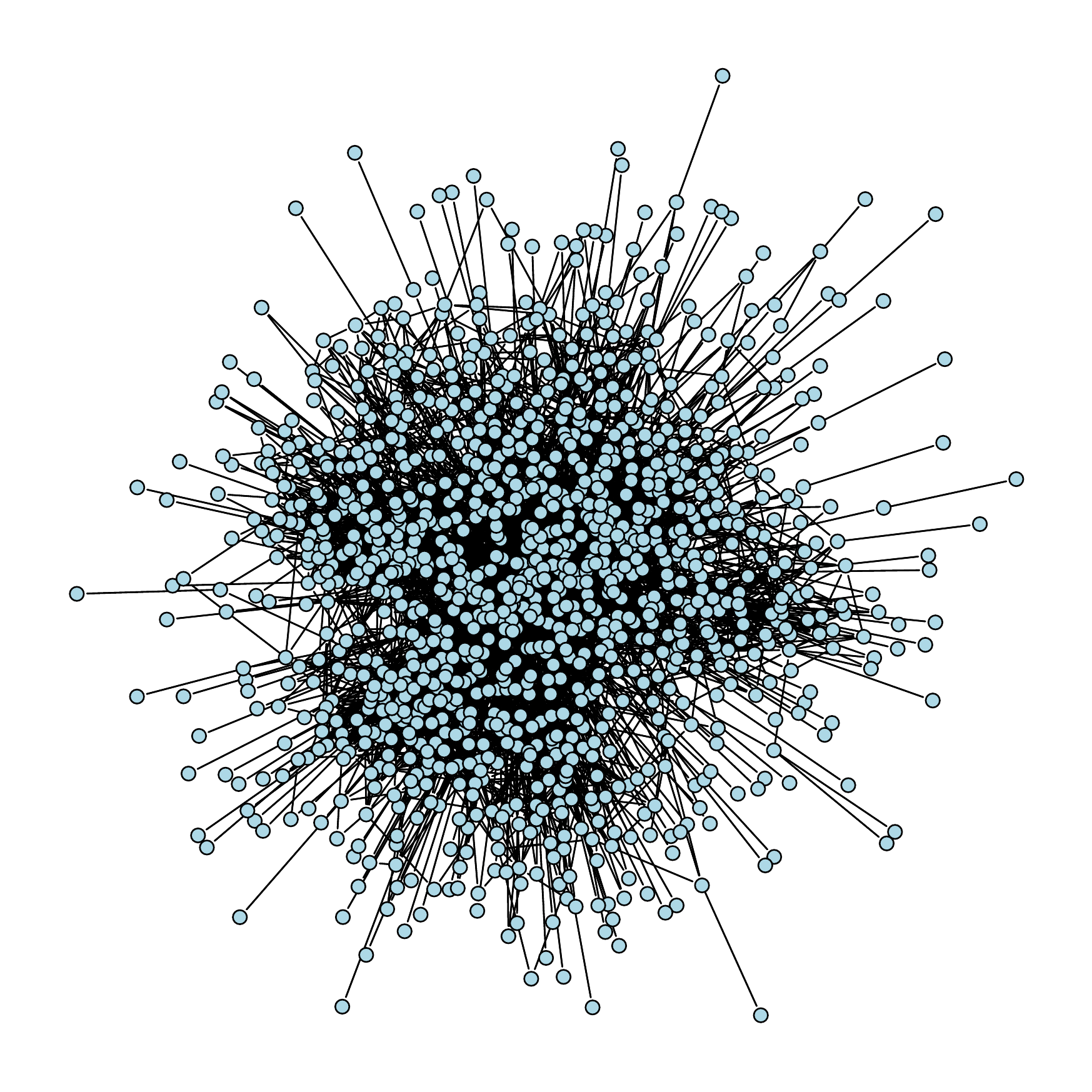}
	\end{tabular}
	\caption{Two real-world networks: Neural network of the worm C. Elegans (left) and E-mail network between the members of the University Rovira i Virgili (right)}
	\label{fig::celegans-email}
\end{figure}

In order to confirm the conclusions reached in the previous section, the
following experiments provide a comparison between the proposed method, the
classical two phases approach and SOM variants on three larger graphs. More
precisely, the studied graphs are:
\begin{itemize}
\item a coappearance network of characters in chapters of the novel
  Les Misérables (Victor Hugo), introduced by \cite{knuth_SGPCC1993}
  and available at
  \url{http://www-personal.umich.edu/~mejn/netdata/lesmis.zip}. This
  graph is undirected and weighted by the number of chapters in which
  each pair of characters appear together. It is larger than the
  graph described in Section~\ref{karate} with 77 nodes representing the
  characters of the novel. The global density of this graph is equal to 8.7\% and its transitivity is equal to
  49.9\%. This 
  indicates that the local connectivity of the graph is very high compared to
  its global one and thus that this graph is likely to be clustered into dense
  subgroups. This graph is represented in Figure~\ref{fig::miserables} (this
  layout has been obtained via the Fruchterman-Reingold force directed algorithm
\cite{FruchtermanReingoldGraph1991} as implemented in igraph);

\item a directed, weighted network representing the neural network of
  the worm ``C. Elegans'', introduced by \cite{watts_strogatz_N1998}
  and available at \url{http://cdg.columbia.edu/cdg/datasets}. The
  graph has been used as an undirected weighted graph: the weights,
  $(W_{ij})$, of the undirected graph are simply defined from the
  weights $(\overrightarrow{W}_{ij})$ of the directed graph by
  $W_{ij}=W_{ji}=\overrightarrow{W}_{ij}+\overrightarrow{W}_{ji}$. This
  graph is connected and contains 453 nodes. Its density is equal to
  2.0\% and its transitivity to 12.4\%. Compared to the graph from
  ``Les Misérables'', this network has a much smaller local
  connectivity indicating that the clustering task should be
  harder. Hence, a smaller modularity is expected for the resulting
  clustering. This graph is represented in
  Figure~\ref{fig::celegans-email} (left). While the graph from ``Les
  Misérables'' was readable, the ``C. Elegans'' graph is impossible to decipher
  when rendered by the standard Fruchterman-Reingold algorithm;
  
\item a network representing the e-mail exchanges between members of the
  University Rovira i Virgili (Tarragona), introduced by
  \cite{guimera_etal_PRE2003} and available at
  \url{http://deim.urv.cat/~aarenas/data/xarxes/email.zip}. This graph is also
  connected and contains 1~133 nodes. Its global density is equal to 0.9\% and
  its transitivity to 16.6\%. As in the previous case, the transitivity of the
  graph is not very high but the gap between the global density and the
  transitivity is much larger and a larger modularity than for ``C. Elegans'' can then be
  expected. This graph is represented in Figure~\ref{fig::celegans-email}
  (right) and is as undecipherable as the ``C. Elegans'' one when rendered
  with the Fruchterman-Reingold algorithm. 
\end{itemize}

\subsection{Experimental setting}
\label{DescExperiments}
\subsubsection{Reference methods}
The experiments compare the proposed method with two other approaches:
\begin{enumerate}
\item as explained in Section \ref{topographicClustering}, the proposed
  approach tries to improve the standard two phases approach by building
  topographically ordered clusters. The reference approach consists therefore
  in a two phases method already used in Section~\ref{karate-modul}: we build
  a good clustering via deterministic annealing maximization of the modularity
  and the graph induced by the clustering is rendered via a force directed
  placement algorithm (the Fruchterman-Reingold algorithm as implemented in
  igraph);
\item an alternative organized approach is the one described in
  \cite{VillaRossiKSOMWSOM2007}: Kohonen's Self Organizing Map algorithm
  is used to build a topographically ordered clustering of the graph (on a prior
  grid) via a well chosen graph kernel. Several popular kernels are defined
  for graphs, including regularized versions of the Laplacian\footnote{We
    recall that the Laplacian of a graph is the $N\times N$ matrix, $L$,
    defined by $L_{ij}=\left\{\begin{array}{ll} -W_{ij} & \textrm{ if }i\neq
        j\\ k_i & \textrm{ if } i=j\end{array}\right.$} (see, e.g.,
  \cite{smola_kondor_COLT2003}), the generalized inverse of the Laplacian
  \cite{foussetaltkde2007} or the modularity kernel
  \cite{zhang_mao_IWMLG2008}. The latter one is defined from the positive part
  of the matrix $B$ involved in the definition of the modularity (see equation
  \eqref{eq:BMatrix}). 
  
  A slightly different way to use the Self Organizing Map for clustering
  the vertices of a graph is what is called ``spectral SOM'' in
  \cite{VillaEtAlMashs2008}. This approach is inspired by spectral
  clustering (see e.g., \cite{vonluxburg_SC2007}) but using a SOM
  instead of the usual $k$-means algorithm. More precisely, the vertices
  of the graph are represented by vectors of \R{C} (where $C$ is the
  initial number of clusters of the prior grid) that are the coordinates
  of the eigenvectors associated to the $C$ smallest non zero
  eigenvalues of the Laplacian of the graph. Then, a usual vector SOM is
  applied to these vectors. 

  Those four SOM variants were used in Section \ref{subsectionKarateSOM} for
  the Karate graph.
\end{enumerate}

\subsubsection{Parameters optimization}
\label{parameters}

The parameters optimization procedure outlined in Section
\ref{section:parameter:tuning} is applied to tunable parameters of the
reference methods (e.g., the number of clusters, the kernel and its
parameters, etc.). Random initialization is included in the procedure for
methods that are not deterministic. 

In the specific case of the two phases approach, we rely on a two phases
optimization: the best clustering is selected by maximization of the
modularity over the number of clusters and then the layout with the lowest
number of crossing edges is kept among 10 different random initializations for
the Fruchterman-Reingold algorithm.

The other solutions are ``all-in-one'' and provide at the same time a
clustering and a layout. We select therefore on Pareto optimal parameter sets.
The following parameters were optimized:
\begin{itemize}
\item the number of clusters for the clustering algorithm was optimally chosen
  in $\{2,\ldots,25\}$ (please note that all methods can produce empty clusters);
\item the prior structure was encoded via a square grid selected among 3
  sizes: $3\times 3$, $4\times 4$ and $5\times 5$ (i.e., 9 clusters, 16
  clusters and 25 clusters). The vertices of the grid have integer coordinates;
\item the entries of the matrix $S$ were calculated via equation
  \eqref{eq:grid:dist}. We compared exponential decrease and linear decrease,
  and used at least four different scaling values, chosen according to the
  induced radius of influence in the prior structure;
\item in the case of the SOM (kernel version or ``spectral SOM''), the optimal
  configuration (size of the prior structure and neighborhood function) was
  selected in the same set of configurations as for the organized modularity;
\item as the batch SOM used in this paper exhibits some dependence to the
  initial configuration, we used 5 different initializations for each run of
  the algorithm. Among them 4 were random ones and the last one was a PCA
  based initialization in which the prior structure is positioned on the plane
  spanned by the first two principal components (this is done with kernel PCA
\cite{scholkopf_etal_NC1998} in the case of kernel SOM and standard PCA in
\R{C} for the spectral SOM);

\item as explained above, we used three different kernels. Among them, only
  the heat kernel has a parameter: it is defined as $K=e^{\beta L}$ where $L$
  is the Laplacian of the graph and $\beta$ a temperature parameter. Four
  values of $\beta$ were used;
 
  We used the three kernels only for the analysis of ``Karate'' and ``Les
  Misérables''. For the two other graphs (C. Elegans and E-mail), we
  restricted the analysis to the generalized inverse that has achieved almost
  the best results in the first experiment (comparable with the best ones
  obtained for the heat kernel) and that does not require the tuning of an
  additional parameter.
\end{itemize}

\subsection{Results and comments}

\subsubsection{Numerical results}
\begin{figure}[htbp]
	\centering
	\includegraphics[width=\textwidth]{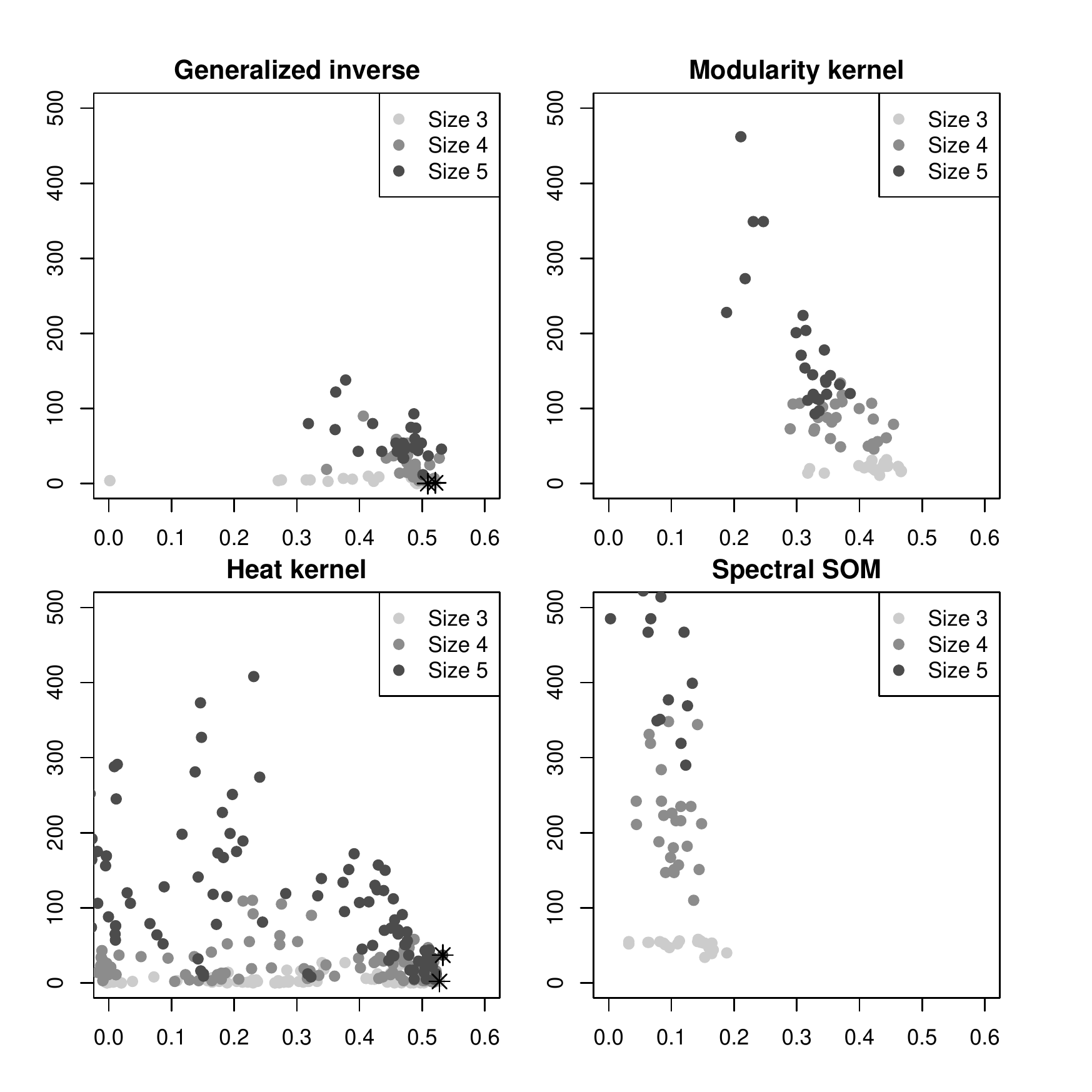}
	\caption{Quality criteria for the kernel and spectral SOM for ``Les
          Misérables'': $x$ axis is the modularity and $y$ axis is the number
          of edge crossings. The gray level of the points gives the initial size of the square grid and Pareto points are indicated by a black star. Solutions with a modularity lower than 0 or a number of edge crossings greater than 500 are omitted on this figure.}\label{fig::ParetoSOM}
\end{figure}

\begin{figure}[htbp]
	\centering
	\includegraphics[width=\textwidth]{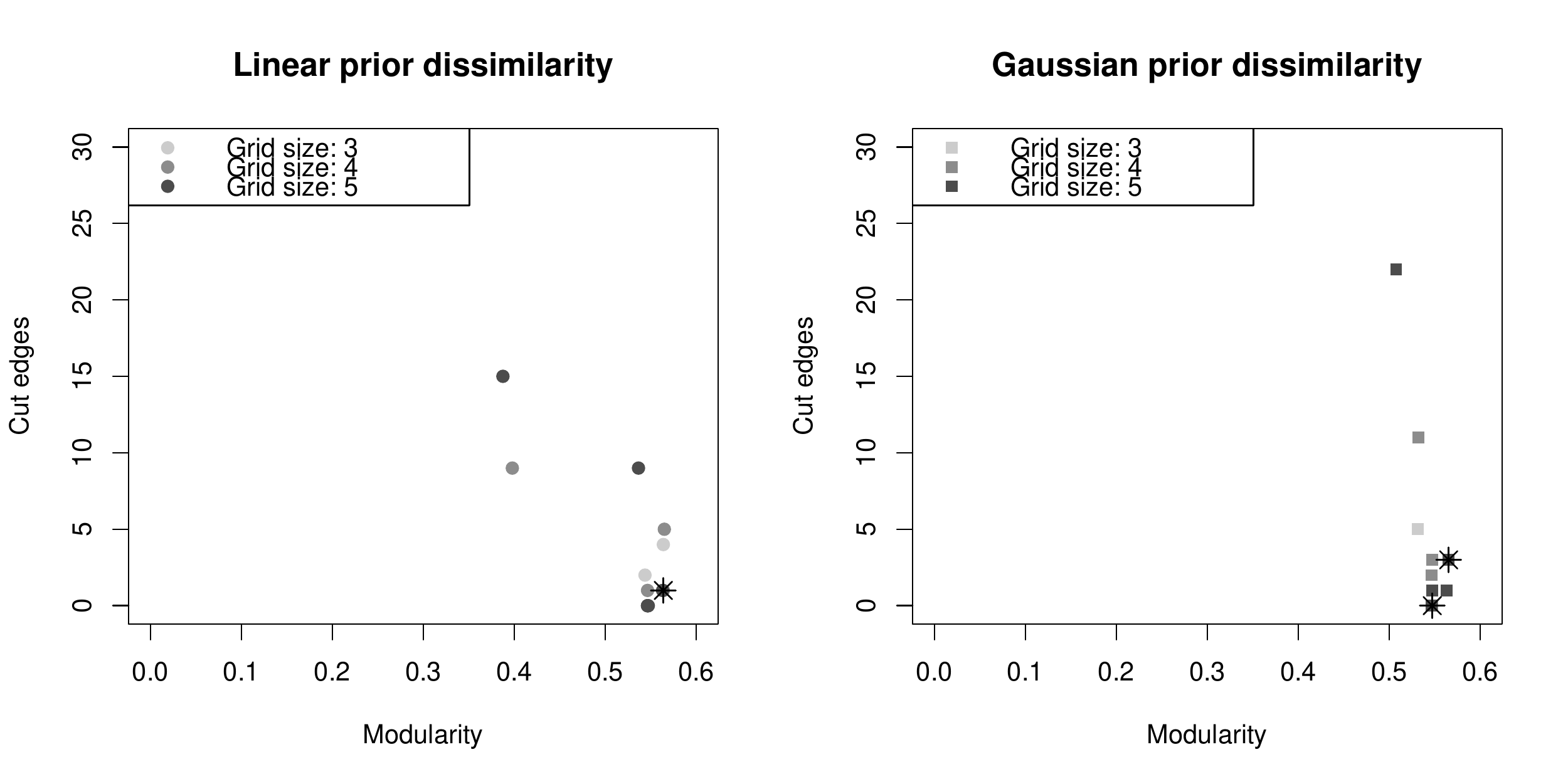}
	\caption{Quality criteria for the optimization of the organized modularity
          by deterministic annealing for ``Les Misérables''. The gray level of the points gives the initial size of the square grid and Pareto points are indicated by a black star.}
\label{fig::ParetoSModularite}
\end{figure}

We first compare the proposed method to the SOM variants with respect to the
chosen quality criteria: the modularity of the clustering and number of edge
crossings in the associated representation. As explained in Section
\ref{topographicClustering} and \ref{section:parameter:tuning}, those quality
criteria have been chosen in order to obtain a good compromise between a
readable clustering based representation of the graph (with small number of
edges crossing) and the fairness of this representation (dense clusters, i.e.,
high modularity). Figures~\ref{fig::ParetoSOM} and
\ref{fig::ParetoSModularite} give the values of the quality criteria for all
the experiments made on the dataset ``Les Misérables'', while Tables
\ref{table::ParetoMiserables}, \ref{table::ParetoCElegans} and
\ref{table::ParetoEMail} give the pairs of values associated to Pareto points
obtained in for each graph. We first focus on ``Les Misérables'' and then give
more general comments. Please note that comments and conclusions about ``Les
Misérables'' apply to the Karate graph with the only exception that the
modularity kernel SOM performs in an acceptable way on the Karate graph. 
Detailed
results are not included to avoid lengthening the article. 
 
As shown in Figures \ref{fig::ParetoSOM} and \ref{fig::ParetoSModularite}, the
clusterings obtained on ``Les Misérables'' by optimizing the soft modularity have
generally larger modularity values than those produced by SOM variants. In
addition, only a limited subset of the parameter space leads to high
modularity with SOM variants. Both outcomes were expected for kernel SOMs
using the heat kernel and the generalized inverse of the Laplacian. Indeed
those kernels induce feature spaces and associated clustering objectives that
have no simple relation with the modularity criterion. As shown in
e.g. \cite{BouletEtAl2008Neurocomputing,YenEtAl2009} both kernels lead to
interesting clustering results, but it is clear from the present results that
one cannot expect to get high modularity clustering from them in general.

The poor results for the spectral SOM are slightly less expected as this
method is related to some form of graph cut measure optimization, using the
relaxation proposed in spectral clustering \cite{vonluxburg_SC2007}. However,
as recalled in Section \ref{subsection:quality}, graph cut measures are quite
different from the modularity, which probably explains the low modularity
clusterings obtained by this method. 

Finally, the results obtained by the modularity kernel are quite
disappointing. Indeed, this kernel is strongly related to the modularity
measure \cite{zhang_mao_IWMLG2008} and should lead to some form of modularity
maximization. In practice, it seems that taking only the positive part of the
$B$ matrix does not capture all the complexity of the modularity, at least in
the SOM context. 

Of course, as our method aims at maximizing an organized version of the
modularity, the higher quality of the results in terms of a closely related
measure (the classical modularity) is rather natural. However, the Figures
show that it also leads to a low number of edge crossings and therefore to
more readable layouts. Figure \ref{fig::ParetoSOM} shows in particular that
SOM variants generally fail to balance modularity and edge crossings: high
modularity clusterings have frequently a significant number of edge crossings
(see also Table \ref{table::ParetoMiserables}).

\begin{table}[htbp]
	\centering
	\begin{tabular}{|l|c|c|c|c|}
\multicolumn{5}{c}{Les Misérables}\\
		\hline
		Method & Number & {\small Modularity} & Nb of pairs & Id\\
		& of clusters && of cut edges&\\
		\hline
		\hline
		SOM (heat kernel) & $5^2$ (22) & 0.5327 & 37 & M1\\
		& $3^2$ (8) & 0.5276 & 2 & M2\\
		\hline
		SOM (Laplacian's inverse) & $3^2$ (9) & 0.5212 & 1 & M3\\
		& $3^2$ (8) & 0.5089 & 0 & M4\\
		\hline
		\hline
		Organized mod. (linear) & $4^2$ (7) & 0.5638 & 1 & {\bf M5}\\
		\hline
		Organized mod. (Gaussian) & $5^2$ (7) & 0.5652 & 3 & {\bf M6}\\
		& $3^2$ (6) & 0.5472 & 0 & {\bf M7}\\
		\hline
		\hline
		Modularity optimization & 8 (5) & 0.5472 & 0 & {\bf M8}\\
		\hline
	\end{tabular}
	
	\caption{Description of the Pareto points for each of the three
          methods (SOM, optimization of organized modularity and optimization of
          ordinary modularity) applied to the dataset ``Les Misérables''. The
          number of clusters gives the number of clusters $C$ used by the
          algorithm and, in parenthesis, the number of non empty clusters
          obtained finally. The column ``Id'' is used to identify the corresponding solution in the text; Pareto points within all solutions have bold Id.}
	\label{table::ParetoMiserables}
\end{table}

Combining both criteria lead to the search of Pareto points. The case of ``Les
Misérables'' is summarized by Table \ref{table::ParetoMiserables}: when
considered all together, the SOM approaches (kernels and spectral) lead to 4
Pareto points which are of lesser quality than the one obtained by the
proposed approach: none of the Pareto points for kernel SOM is a global Pareto
point for the dataset mainly because of low modularity values. In addition,
and for similar reasons, the spectral SOM and the modularity kernel SOM do not
produce any Pareto point. In fact, the only realistic competitive method in
the SOM family is the kernel SOM based on Laplacian's inverse. Indeed, it
gives comparable results as the ones obtained with the heat kernel but without
the need for a time consuming kernel parameter tuning. The two larger datasets,
``C. Elegans'' and ``E-mail'' were therefore investigated only with this
kernel SOM, excluding all other variants. Pareto points are described in
Tables \ref{table::ParetoCElegans} and \ref{table::ParetoEMail}.

\begin{table}[htbp]
	\centering
	\begin{tabular}{|l|c|c|c|c|}
\multicolumn{5}{c}{C. Elegans}\\
		\hline
		Method & Number & {\small Modularity} & Nb of pairs & Id\\
		& of clusters && of cut edges&\\
		\hline
		\hline
		SOM (Laplacian's inverse) & $3^2$ (9) & 0.3228 & 14 & {\bf CE1}\\
		& $3^2$ (9) & 0.3000 & 7 & {\bf CE2}\\
		& $3^2$ (8) & 0.2936 & 1 & {\bf CE3}\\
		\hline
		\hline
		Organized mod. (linear) & $4^2$ (8) & 0.4373 & 41 & CE4\\
		& $3^2$ (7) & 0.4348 & 27 & CE5\\
		& $3^2$ (7) & 0.4321 & 19 & {\bf CE6}\\
		\hline
		Organized mod. (Gaussian) & $3^2$ (8) & 0.4063 & 15 & {\bf CE7}\\
		\hline
		\hline
		Modularity optimization & 18 (8) & 0.4383 & 27 & {\bf CE8}\\
		\hline
	\end{tabular}
	
	\caption{Description of the Pareto points for each of the three methods (SOM, optimization of organized modularity and optimization of ordinary modularity) applied to the dataset ``C. Elegans''. The
          number of clusters gives the number of clusters $C$ used by the
          algorithm and, in parenthesis, the number of non empty clusters
          obtained finally. The column ``Id'' is used to identify the corresponding solution in the text; Pareto points within all solutions have bold Id.\label{table::ParetoCElegans}}
\end{table}

Results from ``C. Elegans'' and ``E-mail'' generally agree with those obtained
on ``Les Misérables'': the kernel SOM does not produce clusterings with high
modularity, but it manages sometimes to reach a low number of edge
intersections in the induced layout. That said, the kernel SOM does not
produce global Pareto points for ``Les Misérables'' and ``E-mail'': on those
datasets, it is therefore strictly less efficient than the proposed method on
the dual objectives point of view (this was also the case for the Karate
graph). The best situation for the kernel SOM is ``C. Elegans'' where the very
low numbers of edge crossings of configurations CE1, CE2 and CE3 make them
Pareto points within the full set of solutions. However, those points have
very low modularities compared to other solutions.

\begin{table}[htbp]
	\centering
	\begin{tabular}{|l|c|c|c|c|}
\multicolumn{5}{c}{E-mail}\\
		\hline
		Method & Number & {\small Modularity} & Nb of pairs & Id\\
		& of clusters && of cut edges&\\
		\hline
		\hline
		SOM (Laplacian's inverse) & $4^2$ (16) & 0.4652 & 218 & E1\\
		& $3^2$ (9) & 0.4566 & 49 & E2\\
		& $3^2$ (9) & 0.4540 & 24 & E3\\
		& $3^2$ (9) & 0.4420 & 22 & E4\\
		& $3^2$ (9) & 0.4242 & 19 & E5\\
		\hline
		\hline
		Organized mod. (Gaussian) & $3^2$ (8) & 0.5694 & 47 & {\bf E6}\\
		& $4^2$ (8) & 0.5693 & 44 & {\bf E7}\\
		& $4^2$ (7) & 0.5554 & 25 & {\bf E8}\\
		& $3^2$ (7) & 0.5456 & 23 & {\bf E9}\\
		& $5^2$ (6) & 0.5401 & 11 & {\bf E10}\\
		\hline
		\hline
		Modularity optimization & 11 (8) & 0.5736 & 56 & {\bf E11}\\
		\hline
	\end{tabular}
	
	\caption{Description of the Pareto points for each of the three
          methods (SOM, optimization of organized modularity and optimization of
          ordinary modularity) applied to the dataset ``E-mail''. 11 Pareto
          points have been found for the kernel SOM but 6 of them are omitted
          in the table because of a very low modularity (less than 0.2). The 
          number of clusters gives the number of clusters $C$ used by the
          algorithm and, in parenthesis, the number of non empty clusters
          obtained finally. The column ``Id'' is used to identify the corresponding solution in the text; Pareto points within solutions have bold Id.\label{table::ParetoEMail}}
\end{table}

In summary, those experiments show that the spectral SOM and the modularity
kernel SOM should be avoided as they cannot produce satisfactory results in
terms of the chosen quality criteria. In addition, while the heat kernel can
produce interesting results, its additional kernel parameter induces a large
computational cost: as shown on Figure \ref{fig::ParetoSOM} most of the
computing efforts are wasted as they lead to poor solutions on both
criteria. The only competitive method is the Laplacian's inverse kernel
SOM. But even if it seems to be the best SOM based method, it still produces
sub-optimal solutions on both criteria for three out of four datasets.

The comparison between the classical two phases approach and our method does
not lead to a clear winner as far as the chosen quality criteria are concerned. As
expected, the modularities of the clusterings obtained in the two phases
approach are among the highest. Interestingly, maximizing the organized
modularity can lead to a higher modularity than a more direct approach (in a
similar way as the SOM which can overcome the $k$-means in term of within
cluster variance): this is the case for ``Les Misérables'' (see Table
\ref{table::ParetoMiserables}). However, in general, the two phases approach
gives the clusterings with the highest modularity. In term of edge crossings,
the two phases approach gives also satisfactory results: this confirms the
analysis from Section \ref{subsection:quality}, where we argued that looking
for dense clusters should reduce the number of edges between clusters. On
complex graphs however, our method leads to lower numbers of edges cuts:
again, this confirms the somewhat contradictory nature of the two quality
criteria explained in Section \ref{subsection:quality}.

By construction, the two phases approach should give a Pareto optimal result,
up to sub-optimal optimization results caused by the combinatorial nature of
both optimization problems. In practice, we obtained Pareto points on all
graphs. In general, those points have higher modularity and higher number of
edge crossings than the ones produced by our method, as expected: our clusters
are somewhat adapted to the clustering induced layout. There is therefore no
obvious superiority of one method over the other. In fact, we are more
interested in the final layout of the graphs than in the exact numerical
results. It is therefore interesting to study the visual representations
obtained different approaches, as long as they correspond to Pareto optimal
points. This is done in the next subsection.

\subsubsection{Drawing the graph from its clustering}
\begin{figure}[htbp]
	\centering
	\includegraphics[width=\textwidth]{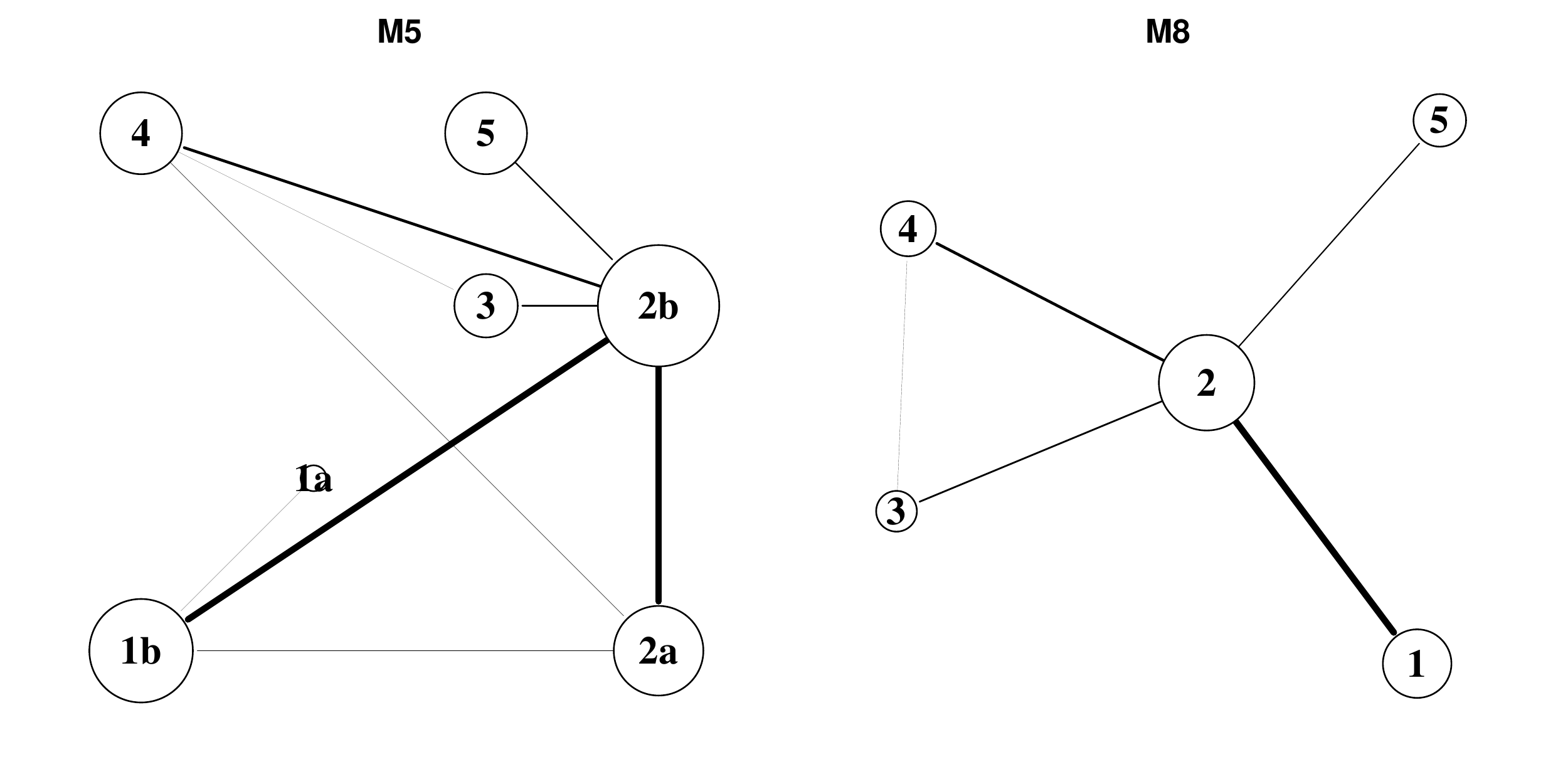}
	
	\caption{Clustering induced graphs. Left: displayed on the prior structure (for M5 obtained by optimizing the organized modularity) and Right: displayed by Fruchterman and Reingold algorithm (for M8 obtained by optimizing the ordinary modularity).\label{fig::miserables-classifLayout}}
\end{figure}

The aim of the proposed method is to provide a simplified but relevant
representation of a large graph through the graph induced by the
clustering. The present section analyses the visual representation obtained
via the proposed method and reference methods. We first start with a detailed
analysis of ``Les Misérables'': Figure~\ref{fig::miserables-classifLayout}
gives the representation of the graph of clusters for clusterings M5
and M8 (See Table \ref{table::ParetoMiserables}).

\begin{figure}[htbp]
	\centering
	\includegraphics[width=\textwidth]{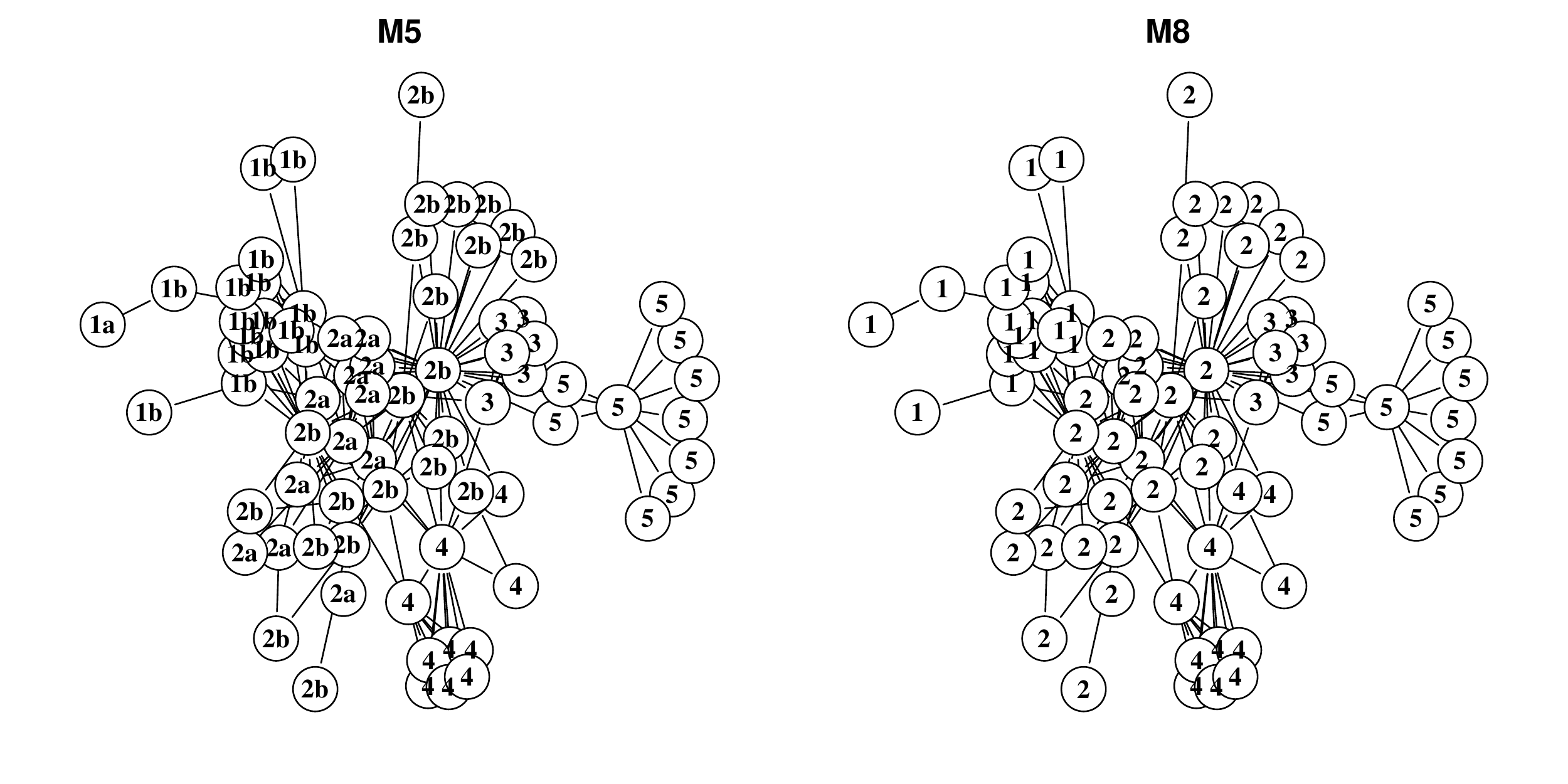}
	
	\caption{Original layout of ``Les Misérables'' labeled according to the clustering; left: M5 (optimization of the organized modularity) and right: M8 (optimization of the ordinary modularity). The number on the vertices is the cluster's number.\label{fig::lesmis.graphesClassif}}
\end{figure}

As the graph is small enough, those layouts can be compared to the original
layout numbered according to the clustering: this is provided by
Figure~\ref{fig::lesmis.graphesClassif}. 
The density of the graph limits the possibilities of analysis, but it appears
clearly that cluster 1a from M5 corresponds to a quite isolated character
(Mother Plutarch who is only connected to Mabeuf), while up to a single
character (Sister Simplice), the union of clusters 2a and 2b in M5 gives
cluster 2 in M8. In both cases, the summary of the graph given by the
clustering seems therefore reasonable and a more detailed analysis is needed. 

It turns out that both representations give a clear understanding of the story
of the novel. It is based on a central group of characters (clusters 2a and 2b
for M5 and 2 for M8) which includes Valjean, Cosette and Marius (among
others). Several sub-stories are narrated in the novel: the story of the
Bishop Myriel who is spiritual guide of Valjean (clusters 5); the story of the
street children Gavroche (clusters 1 in M5 and 1b in M8); and finally the
story of Fantine, poor Cosette's mother (clusters 4). But representation M5
gives another information by separating the main characters into the ones
related to Valjean (Cosette, Marius, Javert, etc. in cluster 2b) and the ones
related to the Thénardier family (mister and misses Thénardier, Éponine, their
daughter, etc. in cluster 2a). In addition, cluster 2b of M5 (the main
characters) has more connections to cluster 1b than cluster 2a (Thenardier
family) to cluster 1b. Therefore, the central position of cluster 2b remains
clearly emphasized in M5, even if the representation is arguably slightly less
readable than representation M8 (it should be noted that the graph induced by
the clustering M5 is planar, but because the organized layout is not optimized
by a graph drawing algorithm, the actual representation has one edge
crossing). Hence, with a small (and avoidable) increase in the number of edge
crossings but with a higher quality of clustering, the information given by
clustering M5 provides a more complete picture of the original network.

\begin{figure}[htbp]
	\centering
	\includegraphics[width=0.6\textwidth]{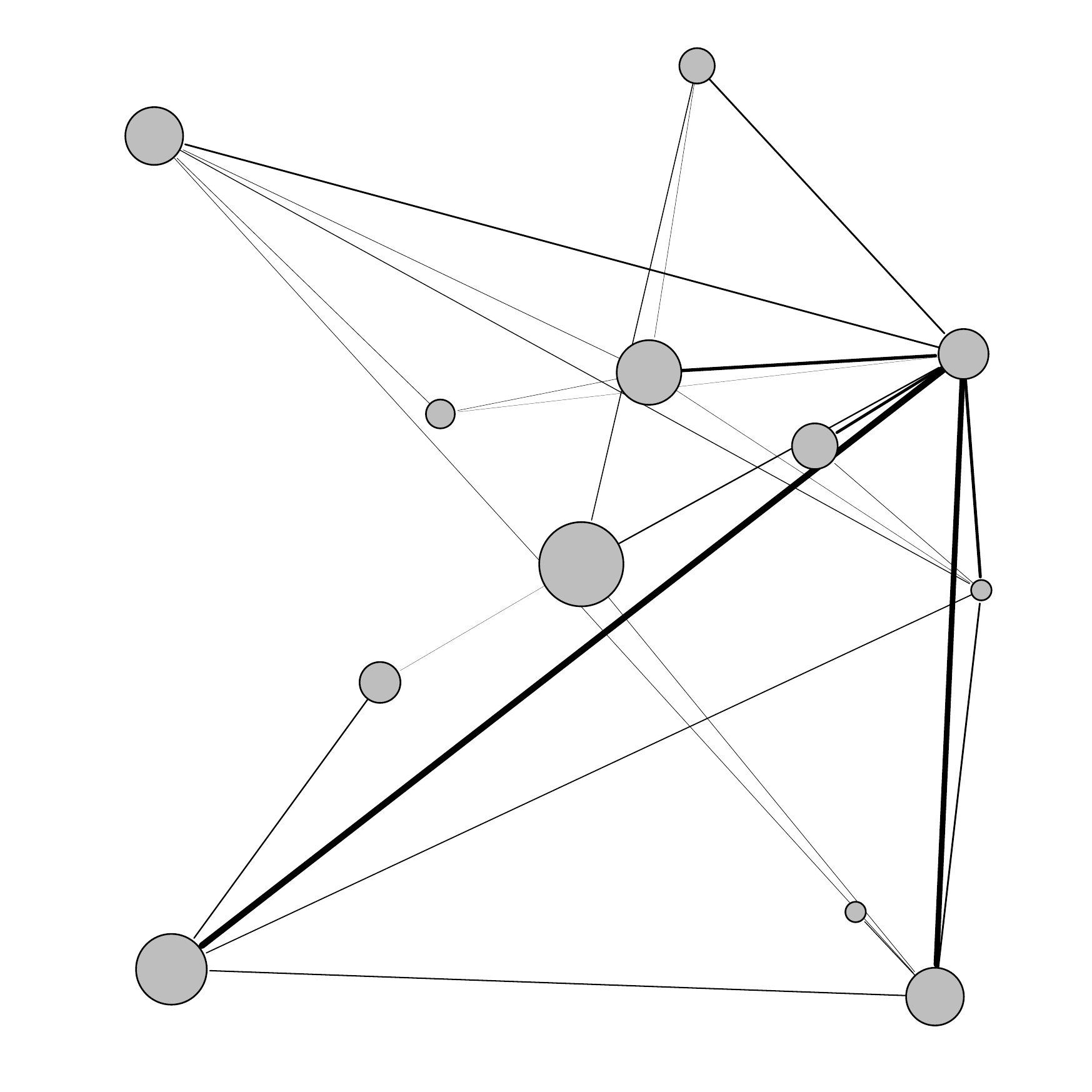}
	\caption{Fuzzy layout of the final configuration of ``Les Misérables''
        (the
    Fruchterman-Reingold algorithm was used to reduce edge overlapping)}
	\label{fig::improvedFuzzyMiserables:final}
\end{figure}

Additional insights can be gained with the help of the fuzzy layout
methodology described in Section \ref{subsection:fuzzylayout}. Figure
\ref{fig::improvedFuzzyMiserables:final} gives an example of such layout for
the final configuration of the deterministic annealing. At first, this
representation might seem very different from M5 on Figure
\ref{fig::miserables-classifLayout}. In fact most of the differences can be
explained by two phenomenons. Firstly, as on Figure
\ref{fig:FuzzyEvolutionKarate} for the Karate graph, some nodes are difficult
to assign to a given cluster and appear therefore in intermediate
position. This explains the presence of some small clusters in between larger
ones : this is the case, for example, of the small cluster at the extreme
right of the Figure who is Javert, the policeman who is pursuing Valjean (one
of the character of the larger cluster just above the small isolated one). But
this character is also very related to the Thénardiers' family (larger cluster
below the small isolated one) because they capture and imprison him at the end
of the novel. The small cluster situated between the large central cluster and
the cluster at the bottom left part of the map is also interesting. It
contains 3 characters that act as connections between several secondary
characters and the rest of the graph: these characters belong to cluster
number 5 in clusterings M5 and M8. 

The second source of differences between M5 and the fuzzy layout is that
several characters are extremely difficult to assign to a cluster and have
therefore almost flat assignment probabilities $\espb{R}{M_{ik}}$ even at low
temperature. Then, the corresponding nodes assume an almost barycentric
position in the fuzzy layout: this explains the presence of a central node in
Figure \ref{fig::improvedFuzzyMiserables:final}. The characters assigned to
this central node are not connected and this cluster should not be seen as a
meaningful one. On the contrary, it contains very isolated characters who
belong to clusters 5, 2 or even 1 in clustering M8. All these characters
have in common to share only a (small weighted) connection with another
character of the network. They are very secondary characters in the
novel. None of those ambiguities could have been detected on a standard two
phases approach or on the SOM variants.

\begin{figure}[htbp]
  \centering
\includegraphics[width=0.45\textwidth]{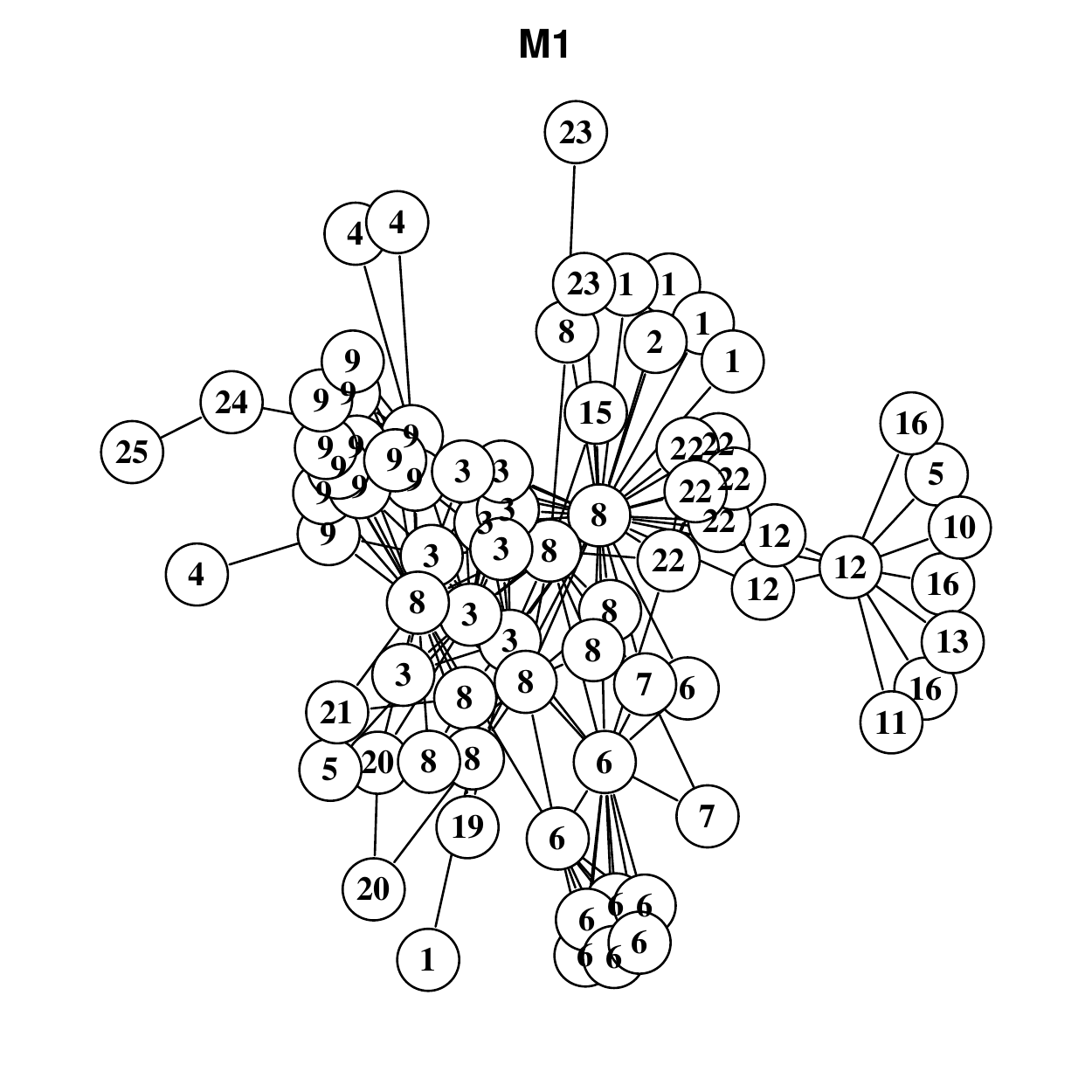}
\includegraphics[width=0.45\textwidth]{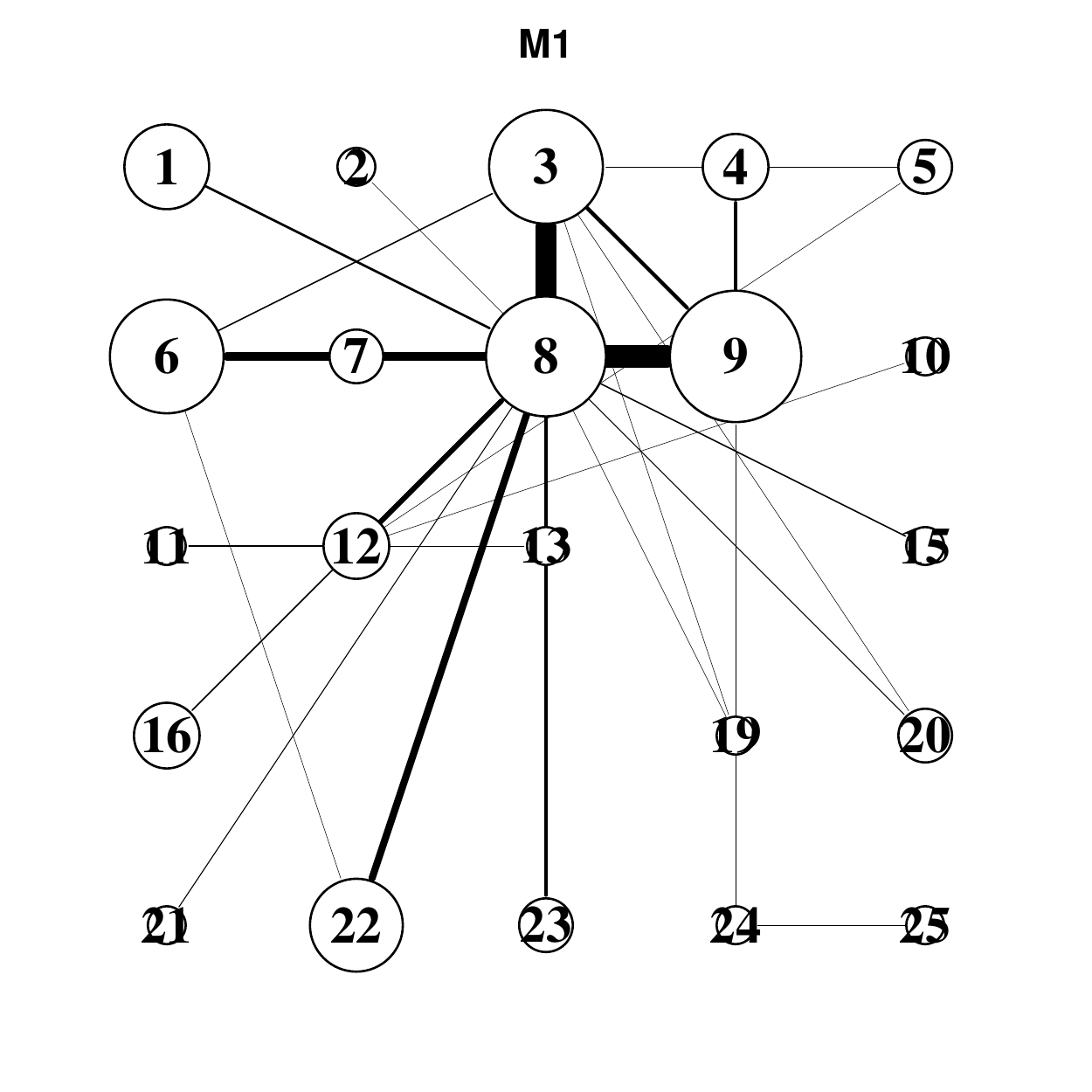}  
  \caption{Representation of the M1 clustering obtained with the heat kernel
    SOM on the original layout (left) and via the prior structure (right)}
  \label{fig:MiserableHeatKernel}
\end{figure}

In addition, the SOM variants obtain poor results on this graph. As shown in
Table \ref{table::ParetoMiserables} they do not provide any global Pareto
optimal points. Therefore it is not very surprising to obtain unsatisfactory
visual representations from those methods. Figure
\ref{fig:MiserableHeatKernel} represents the results associated to clustering
M1 in Table \ref{table::ParetoMiserables} (obtained with the heat kernel
SOM). The resulting clustering has numerous non empty clusters and leads to a
quite cluttered visual representation. In addition, some clusters seems
completely arbitrary: for instance clusters number 5, 10, 11, 13 and 16 are a
seemingly random clustering of the characters related only to the Myriel
characters. The global impression is that the heat kernel SOM suffers from a
tendency to build too fine clusterings on a somewhat arbitrary basis. 

\begin{figure}[htbp]
  \centering
\includegraphics[width=0.45\textwidth]{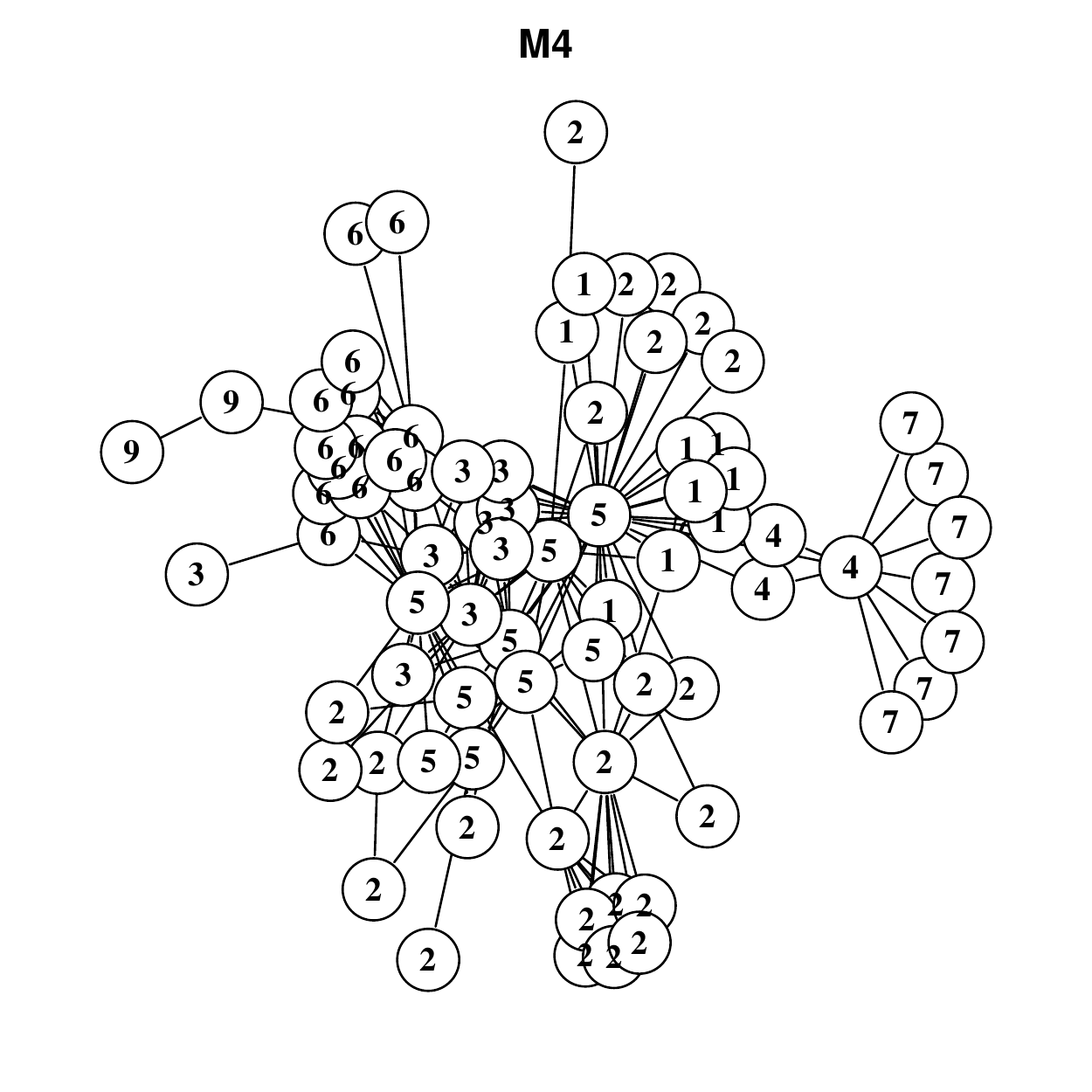}
\includegraphics[width=0.45\textwidth] {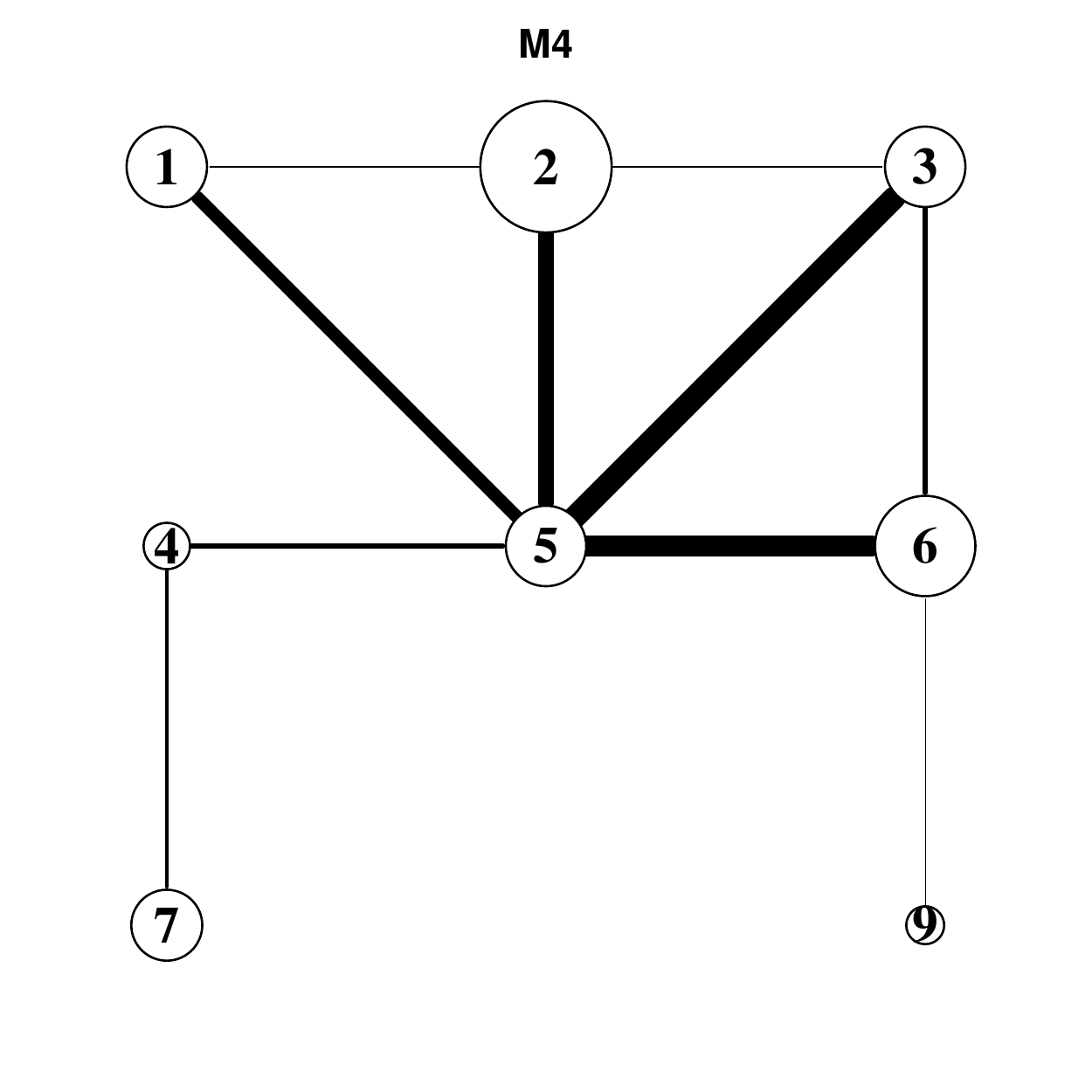}  
\caption{Representation of the M4 clustering obtained with the Laplacian's
  inverse kernel SOM on the original layout (left) and via the prior structure
  (right)}
  \label{fig:MiserableGInv}
\end{figure}

The cluster and layout obtained by the Laplacian's inverse kernel SOM are
given on Figure \ref{fig:MiserableGInv}. Results are quite different from
those of the heat kernel SOM. Indeed, the clustered layout is very easy to
read as there are no edge crossing. However, the rather low modularity of M1
(see Table \ref{table::ParetoMiserables}) is an indication that clusters are
not as dense as for clusterings M5 and M8. In fact, the representation of M1
clustering on the original layout of the graph shows that the associated
visualization is quite misleading. For instance, cluster 3 contains Mother
Plutarch while its unique connection in the graph, Mabeuf, is in cluster
6. Therefore cluster 3 is not even connected. A similar problem can be seen in
cluster 2. Then the main underlying assumption of the clustering induced layout
is not fulfilled: some clusters are not dense at all and are then meaningless. 

This thorough analysis of ``Les Misérables'' confirms results on the Karate
graph (Section \ref{karate}) and the numerical analysis from the previous
section. The analysis leads to two important conclusions. Firstly, the
relevance of modularity and of edge crossings is confirmed: the rather poor results
obtained by SOM variants in terms of numerical results correspond either to
layouts that are difficult to read (see Figure \ref{fig:MiserableHeatKernel})
and/or to misleading clustering (see Figures \ref{fig:MiserableHeatKernel} and
\ref{fig:MiserableGInv}). Secondly, the proposed methodology leads to a better
understanding of the graph than the two phases approach by means of different
clusterings (induced by the organized modularity maximization) and via the
fuzzy layout methodology. 

The sizes of ``C. Elegans'' and ``E-mail'' graphs make them difficult to
analyze in details. We provide therefore only some qualitative comparisons of
the layouts obtained by the investigated methods. 

\begin{figure}[htbp]
	\centering
	\includegraphics[width=\textwidth]{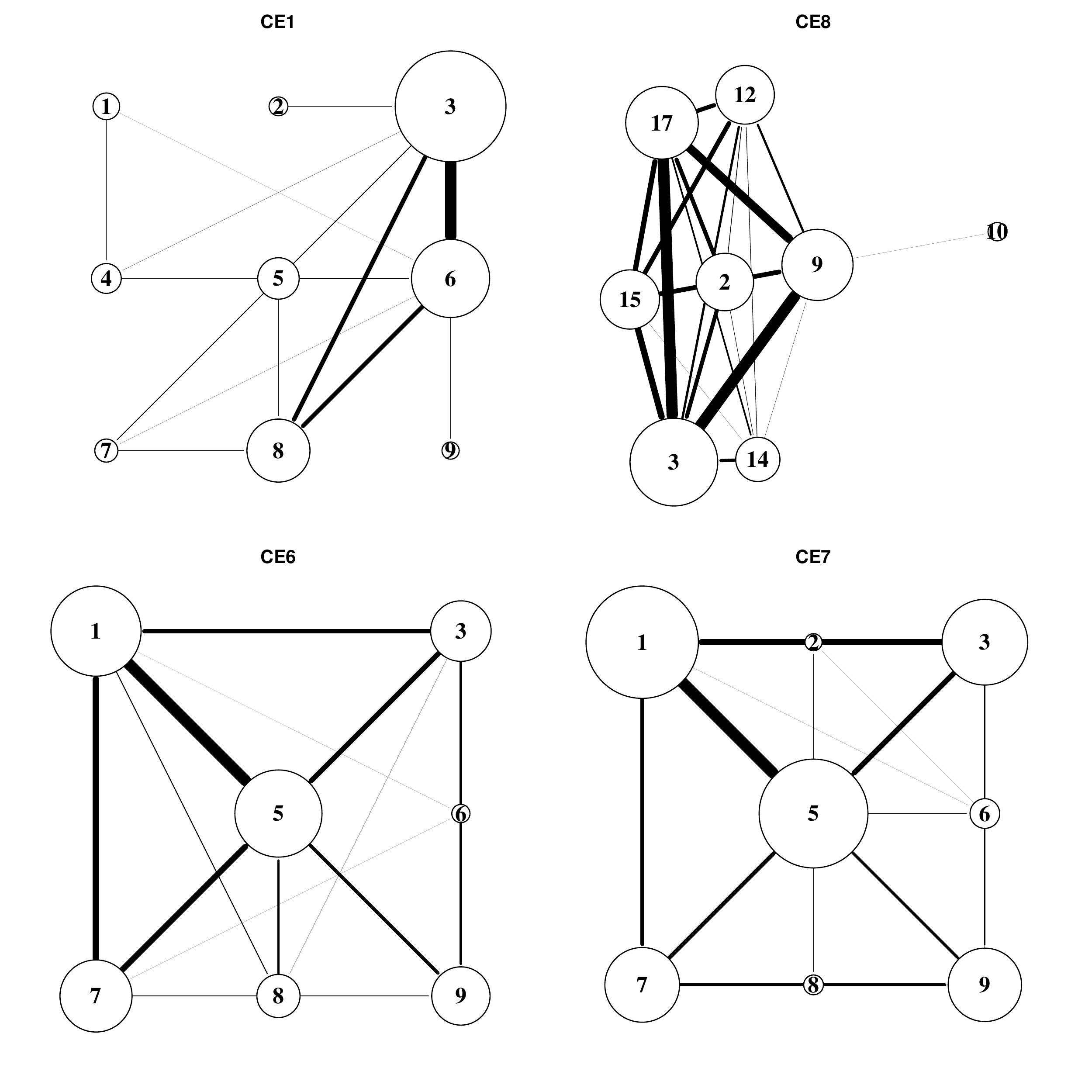}
	
	\caption{Clustering induced graphs for the ``C. Elegans'' dataset. Top left: displayed on the prior structure (for CE1 obtained by kernel SOM); Top right: displayed by Fruchterman and Reingold algorithm (for CE8 obtained by maximization of the modularity); Bottom: displayed on the prior structure (for CE6 and CE7 obtained by maximization of the organized modularity).}
	\label{fig::celegans-classifLayout}
\end{figure}

Figure \ref{fig::celegans-classifLayout} gives four layouts of the graph
``C. Elegans'' obtained from a subset of Pareto points of Table
\ref{table::ParetoCElegans}:
\begin{itemize}
\item CE1 is the layout obtained by kernel SOM with the highest
  modularity (chosen to avoid misleading clusterings);
\item the CE8 is the layout obtained by the two phases approach and is induced
  by the clustering with the highest modularity;
\item CE6 and CE7 are the two Pareto points obtained by our method: CE7 has
  the smallest modularity but also a smaller number of edge crossings within
  this two solutions. These two solutions have a high modularity (larger than
  0.4) and the modularity of CE6 is very close to those of CE8 but with a smaller number of edge crossings. 
\end{itemize}

Despite a very small number of edge crossings, CE1 is poorly informative:
three clusters (3, 6 and 8) contain more than 80~\% of the vertices of the
graph and the other clusters are thus very small compared to the three largest
ones. The small clusters are then not very informative and their existence
probably explains the poor modularity of the clustering. In fact, all solutions
provided by kernel SOM for this dataset share the same problem with some large
clusters and some very small ones. This type of behavior is neither specific
to the chosen graph, nor to the Laplacian's inverse kernel as shown on Figure
\ref{fig:MiserableGInv} and in \cite{BouletEtAl2008Neurocomputing} for the
heat kernel. 

CE8 shows another type of layout problems: the graph induced by the clustering
is almost a complete graph (up to cluster 10, the graph is complete). The only
information conveyed by this layout is that nodes in cluster 10 are only
connected to nodes in cluster 9. The width of the edges can be used to infer
that some clusters are only loosely connected (e.g., 12 and 14), but grasping
the general organization of the graph is quite difficult with this layout.  

CE6 and CE7 both have a slightly smaller modularity than CE7 but they are
easier to read because of a smaller number of edge crossings. Their clustering
qualities are higher than the one of CE1 (with a modularity larger than 0.4
and less imbalance between the clusters' sizes), comparable for CE6 to the one
of the clustering obtained in the two phases method CE8. Therefore, CE6 (and to
a lesser extent CE7) is almost as faithful as CE8 but gives some understanding
of the structure of the graph because the resulting simplified graph is not
complete. As expected, a small reduction in clustering quality can lead, in
some situations, to clusters that are more adapted to visual exploration. This
validates again the principle of integrating the clustering process and the
layout process.

\begin{figure}[ht]
	\centering
		\includegraphics[width=0.6\textwidth]{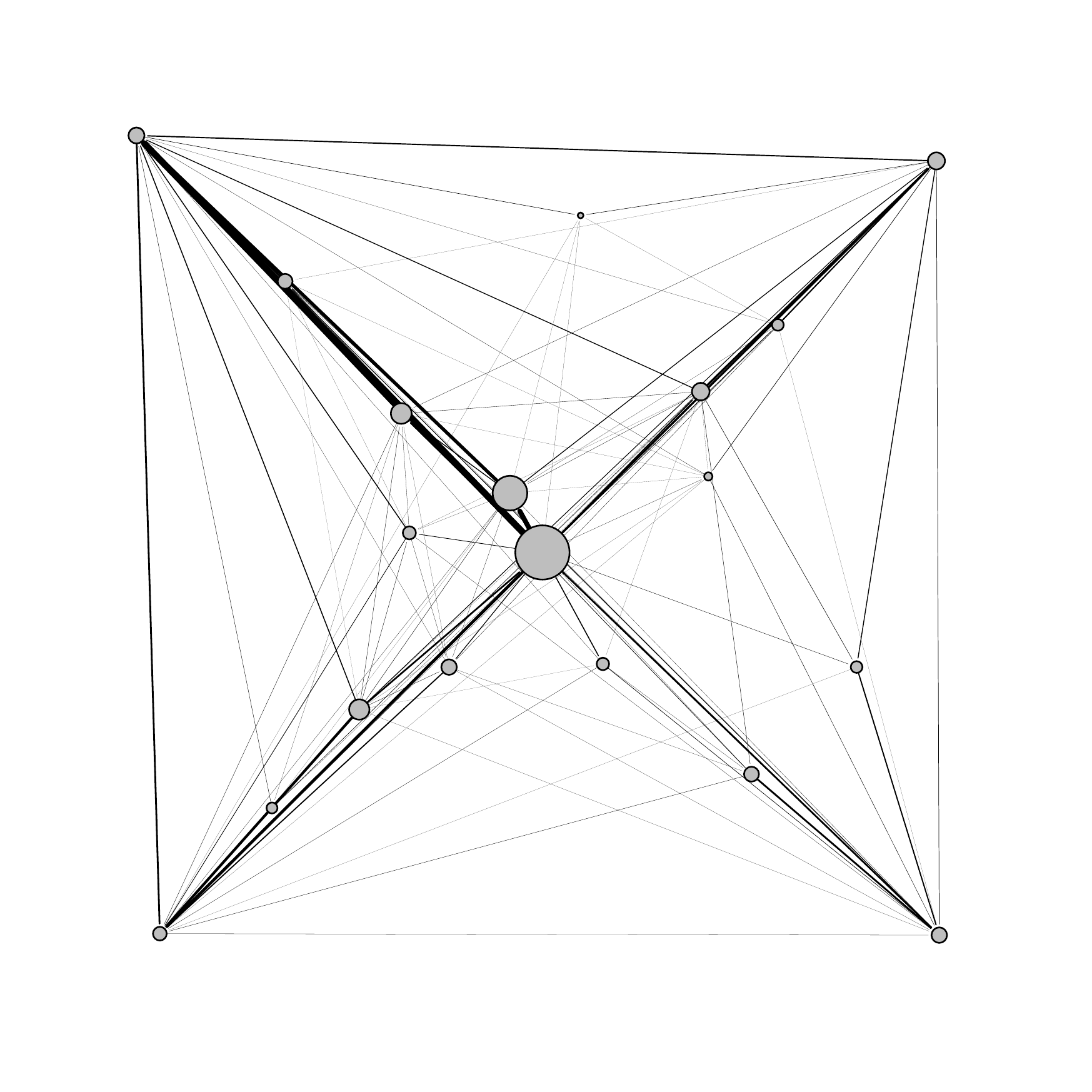}
	\caption{Fuzzy layout during the final cooling phase for the solution CE6 of ``C. Elegans''}
	\label{fig::fuzzyCelegans}
\end{figure}

In addition, fuzzy layouts such as the one provided in Figure
\ref{fig::fuzzyCelegans} can be used to analyze how clusters are built during
annealing. In the particular case of solution CE6, the layout shows that, e.g.,
cluster 8 is formed at the end of the annealing. The analyst can investigate
first the associated nodes and their connection pattern to outside clusters,
or on the contrary, focus his/her attention on well established clusters. 

\begin{figure}[htbp]
  \centering
  \includegraphics[width=\textwidth]{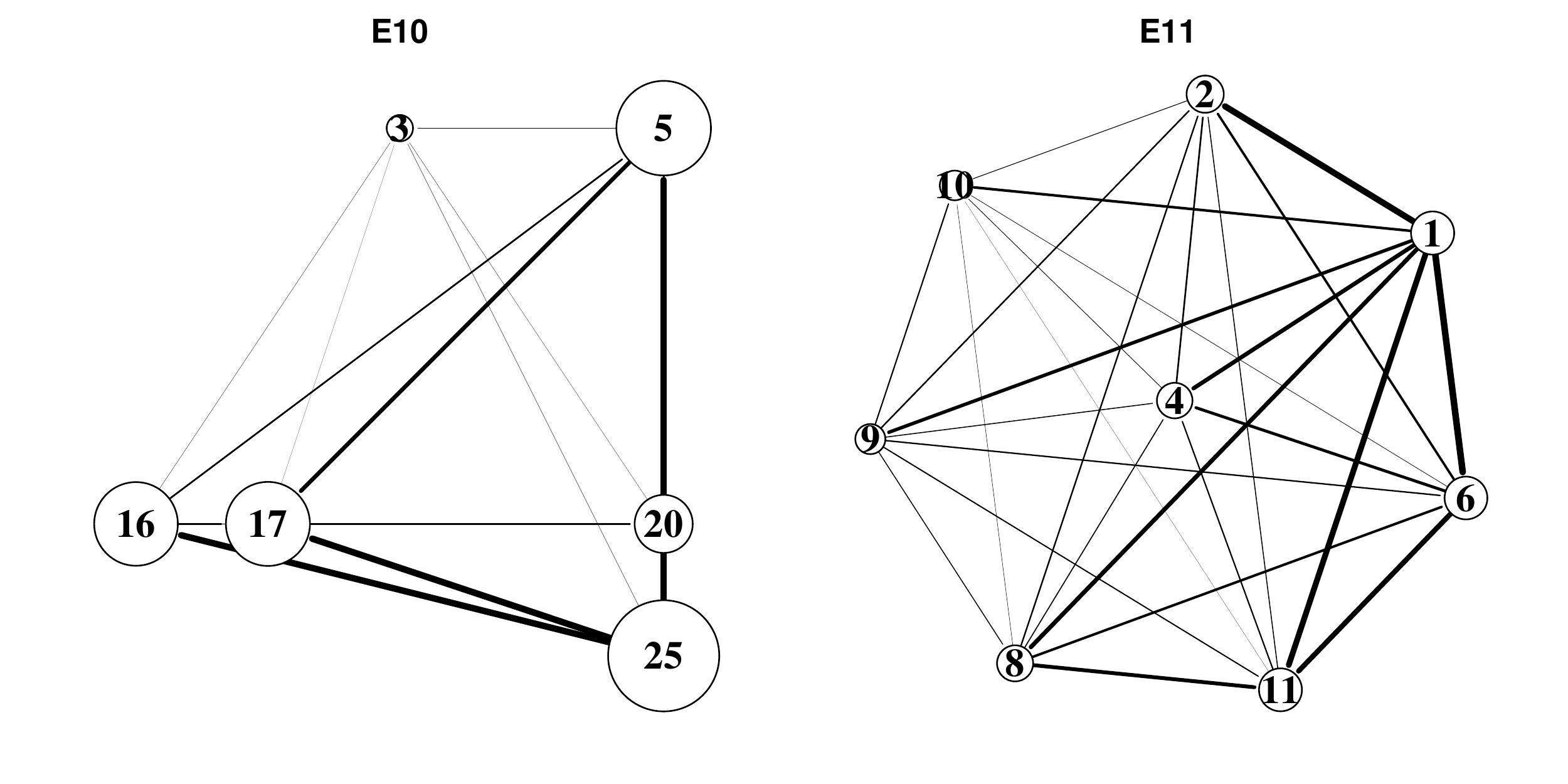}
  
  \caption{Clustering induced graphs for the ``E-mail'' dataset.  Left:
    displayed on the prior structure (E10 obtained by  maximization of the organized modularity);
    Right: displayed by Fruchterman and Reingold algorithm (E11 obtained by
     maximization of the modularity).}
  \label{fig::email-classifLayout}
\end{figure}

Finally, Figure \ref{fig::email-classifLayout} provides two layouts of the graph
``E-mail''. We chose to represent the solution obtained by the two phases
approach (E11, see Table \ref{table::ParetoEMail}) and the Pareto point
obtained by our method with the smallest number of edge crossings (E10).

\begin{figure}[ht]
	\centering
        \includegraphics[width=0.6\textwidth]{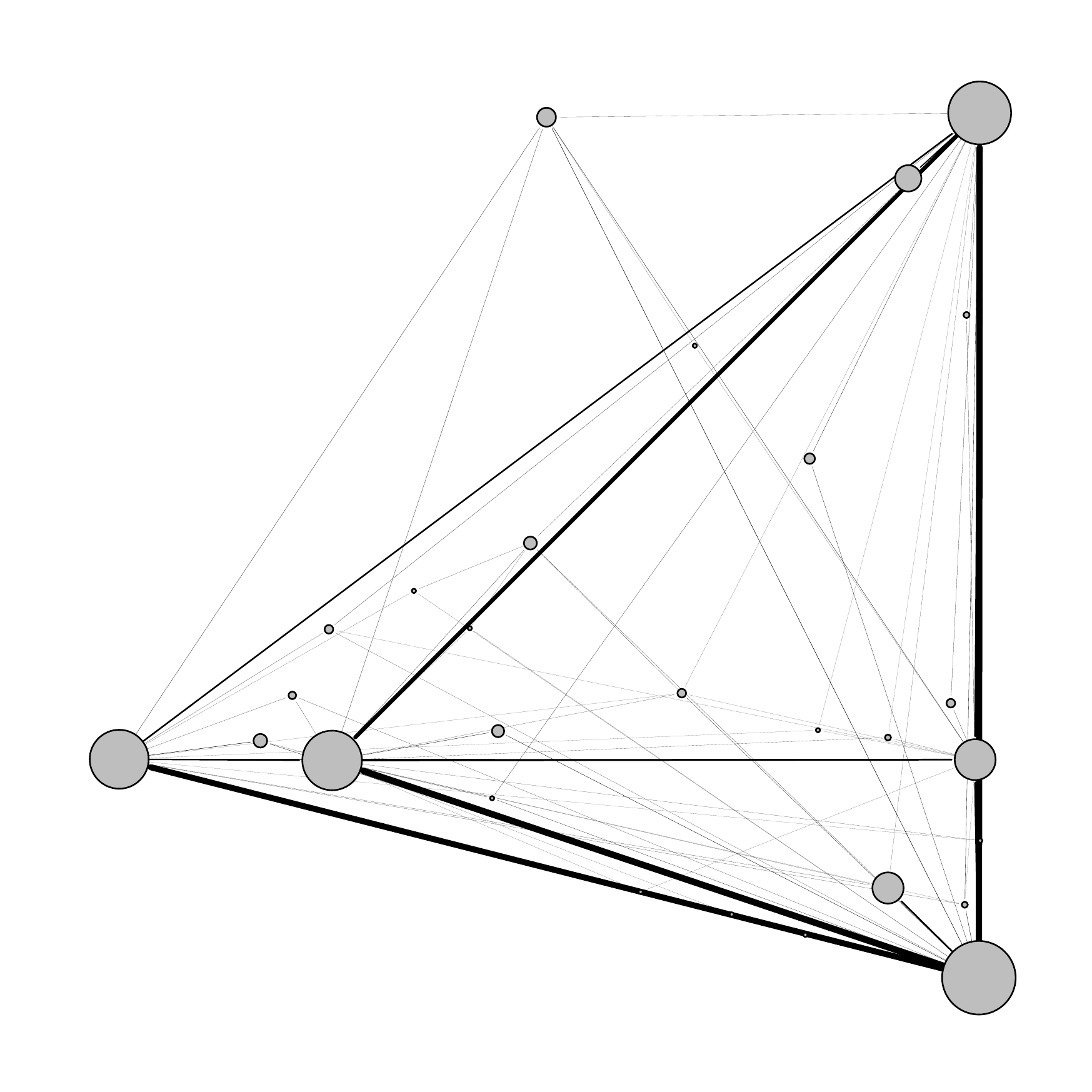} 
	\caption{Fuzzy layout during the final cooling phase for the solution E10 of ``E-mail''}
	\label{fig::fuzzyEmail}
\end{figure}

As in the previous example, the solution provided by the optimization of the
organized modularity, despite a clustering quality that is slightly worse,
gives a more understandable simplification of the graph than the dual phases
approach. In fact, the graph associated to E11 is complete and one has to
rely on the width of the edges to try to infer the importance of the relations
between clusters. The layout associated to E10 seems therefore easier to
grasp. In addition, the associated fuzzy layout (see Figure
\ref{fig::fuzzyEmail}) shows that most of the clusters are well defined and
identifies a few nodes that are difficult to assign to clusters. Further
investigations could target those nodes. 

\subsubsection{A brief summary of the experiments}

Several conclusions can be drawn from the experiments conducted on the four
real world graphs. Firstly, maximizing the organized modularity leads to good
solutions with a high modularity and a acceptable number of edge crossings. In
some situations, maximizing the organized modularity actually gives a higher
modularity clustering, in a similar way to what can be observed for the SOM
versus the $k$-means. However, the main gain is a better trade off between
modularity and readability of the clustering induced graph: we trade a small
reduction in modularity for either a larger number of clusters (limiting the
general tendency of the modularity to pick up a small number of clusters) or a
reduced number of edge crossings (or both). All in one, organized modularity
maximization brings therefore simplified representations of graphs that seem
more informative than the standard two steps approach. In addition, as the
modularity is almost optimal, the representations provided by maximization of
the organized modularity are reasonably faithful, especially compared to the
low quality results generally obtained by graph adapted SOM. While the final
layout obtained by the organized modularity are not perfect (it is quite clear
for instance that the left graph on Figure \ref{fig::miserables-classifLayout}
is planar and could be rendered with no edge crossing), their ordered and
regular nature makes them very readable. In conclusion, we consider that a good
practice consists in combining our method to the two phases approach to get
different and complementary views on the same graph. A future implementation
challenge is to provide linked multi-views \cite{BeckerCleveland1987Brushing}
of a graph that would give the analyst a visual comparison method between
those views.

In addition, relying on deterministic annealing provides a path in the
clustering space that is worth exploring via intermediate fuzzy layouts. The
main advantage of those layouts is to increase the number of clusters and to
providing hints on the final assignment of vertices. Atypical vertices are
clearly pinpointed with this approach and can be further studied by the
analyst. Our use of the fuzzy layout was quite limited in this paper as we
believe they should be provided in an interactive environment in which the
analyst can navigate in the algorithm results. Further studies are needed to
validate the interest of the concept. In addition, as those layouts are closer
to clustered layouts than the rest of our work, comparisons with those latter
methods (e.g., \cite{Noack2007JGAA}) would be interesting.

Those results show that the combination of modularity maximization and edge
crossings minimization is very well adapted to graph visualization. Then, it
is quite natural to observe that SOM variants are not adapted to the targeted
application. The analysis of the Karate graph has shown that they are not able
to recover the ground truth clustering. The graph ``Les Misérables'' showed
quite strong limitations of the clustering obtained by SOM variants (with
meaningless clusters) without any particular gain in term of readability of
the clustering induced graph. This was confirmed on ``C. Elegans'' in terms of
layout and on ``E-mail'' via quality criteria.

\section{Conclusion}
We have proposed in this paper a new organized modularity quality measure for graph
clustering inspired by the topographic mapping paradigm initiated by
Prof. Kohonen's SOM. The organized modularity aims at producing a clustering of a
graph that respects constraints coming from a prior structure. This prior
structure is then used to display the clusters in an ordered way. A
deterministic annealing scheme has been derived to maximize efficiently the
organized modularity. Its iterative nature can be leveraged to provide
intermediate layouts of the graph that emphasize the progressive construction
of the final clustering, pinpointing sub clusters, atypical vertices, etc. 

An experimental study conducted on four real world graphs ranging from
34 to 1 133 vertices has shown that the proposed method outperforms similar
approaches based on adaptation of the SOM to graph data both in term of
clustering quality and in term of readability of the clustering induced
graphs. The proposed method gives also better or complementary results
compared to a two steps approach in which one first build a graph clustering
that maximizes the standard modularity and then uses a graph visualization
algorithm to display the clustering induced graph. Finally, the computational cost
of the whole approach (which includes optimization of an influence parameter
in the prior structure) remains acceptable and is compatible with graphs with
a few thousands of nodes. 

\section*{Acknowledgment}
The authors thank the anonymous referees for their valuable comments that
helped improving this paper. 

\appendix

\section{Derivations of the deterministic annealing algorithm}
\subsection{Equivalence between $O$ and $F$}\label{section:OandF}
We have
\[
O(M)=F(M)+\frac{1}{2m}\sum_{i}\sum_{k,l}M_{ik}S_{kl}M_{il}\left(W_{ii}-P_{ii}\right).
\]
Then, as $M$ is an assignment matrix, $M_{ik}$ is non zero (and equal to one)
only when $k=c(i)$ and therefore
\begin{eqnarray*}
O(M)&=&F(M)+\frac{1}{2m}\sum_{i}S_{c(i)c(i)}\left(W_{ii}-P_{ii}\right)\\
&=&F(M)+\frac{1}{2m}\sum_{i}\left(W_{ii}-P_{ii}\right),
\end{eqnarray*}
as $S_{kk}=1$ for all $k$. Therefore $O(M)-F(M)$ is independent of $M$ and
maximizing $F$ is equivalent to maximizing $O$.

\subsection{Mean field equations}\label{section:MeanFieldEquations}
Let us denote $KL(E)=KL(R|P)=\sum_{M}R(M,E)\ln\frac{R(M,E)}{P(M)}$. At a
minimum of $KL$, the partial derivatives $\frac{\partial KL}{\partial E_{jl}}$
must be equal to zero. Those derivatives can be computed easily from the
definition of $R(M,E)$. We fist note that
\begin{eqnarray*}
KL(E)&=&\sum_{M}R(M,E)\ln R(M,E)-\beta\sum_{M}R(M,E)(F(M)-\log Z_F)\\
&=&\sum_{M}R(M,E)\ln R(M,E)-\beta\espb{R}{F(M)}+\beta\log Z_F.
\end{eqnarray*}
We therefore need to compute $\frac{\partial }{\partial
  E_{jl}}\left(\sum_{M}R(M,E)\ln R(M,E)\right)$. We have 
\[
\frac{\partial U(M,E)}{\partial E_{jl}}=M_{jl},
\]
and
\[
\frac{\partial \exp(\beta U(M,E))}{\partial E_{jl}}=\beta M_{jl}\exp(\beta U(M,E))
\]
As $Z_R(E)=\sum_{M}\exp(\beta U(M,E))$, we have
\begin{eqnarray*}
\frac{\partial Z_R(E)}{\partial E_{jl}}&=&\beta\sum_{M}M_{jl}\exp(\beta
U(M,E))\\
&=&\beta Z_R(E)\espb{R}{M_{jl}}.\\
\end{eqnarray*}
Therefore, recalling $R(M,E)=\frac{\exp(\beta U(M,E))}{Z_R(E)}$, we have
\begin{eqnarray*}
\frac{\partial R(M,E)}{\partial E_{jl}}&=&
-\frac{\exp(\beta U(M,E))}{Z^2_R(E)}\frac{\partial Z_R(E)}{\partial E_{jl}}+\frac{1}{Z_R(E)}\beta M_{jl}\exp(\beta U(M,E))\\
&=&\beta\left(M_{jl}-\espb{R}{M_{jl}}\right)R(M,E).\\
\end{eqnarray*}
Then
\begin{eqnarray*}
\sum_{M}\ln Z_R(E)\frac{\partial R(M,E)}{\partial E_{jl}}&=&\ln Z_R(E)\beta\sum_{M}\left(M_{jl}-\espb{R}{M_{jl}}\right)R(M,E)\\
&=&0,
\end{eqnarray*}
because $Z_R(E)$ does not depend on $M$ and by definition of
$\espb{R}{M_{jl}}$. Moreover,
\begin{eqnarray*}
\sum_{M}\frac{\partial \ln Z_R(E)}{\partial
  E_{jl}}R(M,E)&=&\sum_{M}\frac{\beta Z_R(E)\espb{R}{M_{jl}}}{Z_R(E)}R(M,E)\\ 
&=&\beta\espb{R}{M_{jl}},
\end{eqnarray*}
and therefore
\[
\frac{\partial }{\partial
  E_{jl}}\left(\sum_{M}\ln Z_R(E)R(M,E)\right)=\beta\espb{R}{M_{jl}},
\]
which leads to 
\[
\frac{\partial }{\partial
  E_{jl}}\left(\sum_{M}R(M,E)\ln R(M,E)\right)=\frac{\partial }{\partial
  E_{jl}}\left(\beta\sum_{M}U(M,E)R(M,E)\right)-\beta\espb{R}{M_{jl}}.
\]
By definition of $U(M,E)$, we have
\begin{eqnarray*}
\sum_{M}U(M,E)R(M,E)&=&\sum_{M}\sum_{ik}E_{ik}M_{ik}R(M,E)\\
&=&\sum_{ik}E_{ik}\espb{R}{M_{ik}}. 
\end{eqnarray*}
Then, by independence, $\frac{\partial \espb{R}{M_{ik}}}{\partial E_{jl}}=0$
when $j\neq i$. Indeed we have :
\begin{eqnarray*}
\frac{\partial \espb{R}{M_{ik}}}{\partial
  E_{jl}}&=&\sum_{M}M_{ik}\frac{\partial R(M,E)}{\partial E_{jl}}\\
&=&\beta\sum_{M}M_{ik}\left(M_{jl}-\espb{R}{M_{jl}}\right)R(M,E)\\
&=&\beta\left(\espb{R}{M_{ik}M_{jl}}-\espb{R}{M_{ik}}\espb{R}{M_{jl}}\right)\\
&=&0.
\end{eqnarray*}
The final equality comes from independence of $M_{ik}$ and $M_{jl}$ under $R$
when $i\neq j$ (this simplification is the motivation for using a bi-linear
cost function $U$ and therefore a distribution $R$ that factorizes). 

Finally, we have
\[
\frac{\partial }{\partial E_{jl}}\left(\sum_{M}U(M,E)R(M,E)\right)=
\espb{R}{M_{jl}}+\sum_{k}\frac{\partial \espb{R}{M_{jk}}}{\partial E_{jl}}E_{jk},
\]
and therefore
\[
\frac{\partial KL(E)}{\partial E_{jl}}=\beta\left(\sum_{k}\frac{\partial
    \espb{R}{M_{jk}}}{\partial E_{jl}}E_{jk}-\frac{\partial \espb{R}{F(M)}}{\partial E_{jl}}\right),
\]
from which we obtain the mean field equations \eqref{eqMeanField}.

\subsection{Expectation minimization scheme}\label{section:EMequations}
We need to compute $\frac{\partial \espb{R}{F(M)}}{\partial E_{jl}}$ starting
from 
\[
\espb{R}{F(M)}=\sum_{u\neq v}\sum_{k,t}\espb{R}{M_{uk}}S_{kt}\espb{R}{M_{vt}}B_{uv}.
\]
As shown above, when $j\neq i$, $\frac{\partial \espb{R}{M_{ik}}}{\partial
  E_{jl}}=0$ and therefore when $u\neq j$ and $v\neq j$, 
\[
\frac{\partial \left(\espb{R}{M_{uk}}\espb{R}{M_{vt}}\right)}{\partial E_{jl}}=0.
\]
Then
\begin{eqnarray*}
\frac{\partial \espb{R}{F(M)}}{\partial E_{jl}}&=&\sum_{u\neq
  v}B_{uv}\sum_{k,t}S_{kt}\frac{\partial
  \left(\espb{R}{M_{uk}}\espb{R}{M_{vt}}\right)}{\partial E_{jl}}\\
&=&\sum_{u\neq j}B_{uj}\sum_{k,t}S_{kt}\espb{R}{M_{uk}}\frac{\partial
  \espb{R}{M_{jt}}}{\partial E_{jl}}\\
&&\ +\sum_{v\neq j}B_{jv}\sum_{k,t}S_{kt}\espb{R}{M_{vt}}\frac{\partial
  \espb{R}{M_{jk}}}{\partial E_{jl}}\\
&=&2\sum_{u\neq j}B_{uj}\sum_{k,t}S_{kt}\espb{R}{M_{uk}}\frac{\partial \espb{R}{M_{jt}}}{\partial E_{jl}},
\end{eqnarray*}
using the symmetry of $B$ and $S$ for the last equation. Then if for all $j$
and $k$, we set the values of $E_{jk}$ to
\[
E_{jk}=2\sum_{i\neq j}\sum_{l}\espb{R}{M_{il}}S_{kl}B_{ij},
\]
we have obviously
\begin{eqnarray*}
\sum_{k}\frac{\partial \espb{R}{M_{jk}}}{\partial
  E_{jl}}E_{jk}&=&2\sum_{k}\frac{\partial \espb{R}{M_{jk}}}{\partial
  E_{jl}}\sum_{i\neq j}\sum_{t}\espb{R}{M_{it}}S_{kt}B_{ij}\\
&=&2\sum_{i\neq j}B_{ij}B_{ij}\sum_{k,t}S_{kt}\espb{R}{M_{it}}\frac{\partial
  \espb{R}{M_{jk}}}{\partial  E_{jl}}\\
&=&\frac{\partial \espb{R}{F(M)}}{\partial E_{jl}},
\end{eqnarray*}
and the mean field equations are fulfilled. 

\subsection{Fixed point analysis}\label{subsec:fixedpoint}
Finding the mean field $E$ via the EM-like scheme given by equations
\eqref{eqEstep} and \eqref{eqMstep} corresponds to looking for a fixed point
of the following matrix valued function: 
\begin{equation}
  \label{eqFixedPointDefinition}
G_{jk}(E)=2\sum_{i\neq j}B_{ij}\sum_{l}S_{kl}\frac{\exp(\beta E_{il})}
{\sum_{p}\exp(\beta E_{ip})}.
\end{equation}
The stability of a fixed point $E^0$ can be analyzed with a first order Taylor
expansion (where $\|.\|_F$ is the Frobenius norm) 
\[
G_{jk}(E^1)=G_{jk}(E^0)+\sum_{u}\sum_{t}\frac{\partial G_{jk}}{\partial E_{ut}}(E^0)\left(E^1_{ut}-E^0_{ut}\right)+o(\|E^1-E^0\|_F),
\]
which recalls that the stability is governed by the eigenvalues of the
$NC\times NC$ Jacobian matrix $\left(\frac{\partial G_{jk}}{\partial
 E_{ut}}\right)_{(j,k),(u,t)}$.

Obviously, $\frac{\partial G_{jk}}{\partial E_{jt}}=0$. When $u\neq j$, we have  
\begin{eqnarray*}
\frac{\partial G_{jk}}{\partial E_{ut}}&=&2B_{uj}\sum_{l}S_{kl}\frac{\partial
}{\partial E_{ut}}\left(\frac{\exp(\beta E_{ul})}
{\sum_{p}\exp(\beta E_{up})}\right)\\
&=&\frac{2B_{uj}\beta\exp(\beta E_{ut})}{\left(\sum_{p}\exp(\beta
    E_{up})\right)^2}\left(S_{kt}\sum_{p}\exp(\beta
  E_{up})-\sum_{l}S_{kl}\exp(\beta E_{ul})\right)\\
&=&2\beta B_{uj}\espb{R}{M_{ut}}\left(S_{kt}-\sum_{l}S_{kl}\espb{R}{M_{ul}}\right)\\
\end{eqnarray*}
At the limit of infinite temperature (when $\beta=0$), equation
\eqref{eqEstep} leads to $\espb{R}{M_{jk}}^0=\frac{1}{C}$, and therefore for
the high temperature fixed point $E^0$,
\[
\frac{\partial G_{jk}}{\partial E_{ut}}(E^0)=\frac{2B_{uj}\beta}{C}\left(S_{kt}-\frac{1}{C}\sum_{l}S_{kl}\right),
\]
while equation \eqref{eqMstep} gives in addition
\[
E^0_{jk}=\frac{2}{C}\sum_{i\neq j}B_{ij}\sum_{l}S_{kl}. 
\]
Let us denote $H=S(I-\frac{1}{C}\mathbf{1})$ the centering matrix ($I$ is the $C\times C$
identity matrix and $\mathbf{1}$ is the $C\times C$ matrix with all terms
equal to 1). Using the symmetry of $S$, we have
\[
\frac{\partial G_{jk}}{\partial E_{ut}}(E^0)=\frac{2B_{uj}\beta}{C}H_{tk}.
\]
Then, using the symmetry of $B$, this leads to 
\[
\sum_{u}\sum_{t}\frac{\partial G_{jk}}{\partial
  E_{ut}}(E^0)\Delta_{ut}=\frac{2\beta}{C}(B\Delta H)_{jk},
\]
for any $N\times C$  matrix $(\Delta_{ut})_{ut}$. Let $\lambda_S$ denote $S$'s
largest eigenvalue (in absolute value if $S$ is not 
positive) and let $\lambda_B$ denote $B$'s largest eigenvalue (also in absolute value). Let $\delta$ be
a vector of \R{C}, then $\|\delta^TS\|_2\leq \lambda_S\|\delta\|_2$ (where
$\|.\|_2$ is the Euclidean norm). As $I-\frac{1}{C}\mathbf{1}$ has
eigenvalues $1$ and $0$, $\|\delta^TH\|_2\leq \lambda_S\|\delta\|_2$. In
addition, if $\mu$ is a vector in \R{N}, $\|B\mu\|\leq \lambda_B\|\mu\|$. Then
$\|B\Delta H\|_F\leq \lambda_S\lambda_B\|\Delta\|_F$. Then 
\[
\|G(E^1)-G(E^0)\|_F\leq
\frac{2\beta\lambda_S\lambda_B}{C}\|E^1-E^0\|_F+o(\|E^1-E^0\|_F). 
\]
Therefore, when the temperature $\frac{1}{\beta}$ is higher than
$\frac{2\lambda_B\lambda_S}{C}$, $\|G(E^1)-G(E^0\|_F<\|E^1-E^0\|_F$ and
therefore the fixed point $E^0$ is stable: small perturbations vanish.

\bibliographystyle{abbrvnat}
\bibliography{som-like-graph}

\end{document}